\documentclass[11pt]{article}

\usepackage{bbm}
\usepackage{eqnarray}
\usepackage{color}
\usepackage{bm}
\usepackage{amssymb}
\usepackage[dvips]{graphicx}
\usepackage{epsfig}
\usepackage{epsf}
\usepackage{float}
\usepackage{subfigure}
\usepackage{amsfonts,amsmath,amsthm,fancyhdr,multirow,hyperref}
\usepackage{booktabs,longtable,authblk}
\usepackage{mathrsfs,hhline}

\usepackage[top=2.5cm, bottom=2.5cm, left=3cm, right=3cm]{geometry}
\newtheorem{theorem}{Theorem}[section]
\newtheorem{assumption}{Assumption}

\newtheorem{definition}[theorem]{Definition}

\newtheorem{lemma}[theorem]{Lemma}
\newtheorem{proposition}[theorem]{Proposition}

\allowdisplaybreaks[4]

\renewcommand{\H}{\mathcal{H}}
\renewcommand{\P}{\mathcal{P}}
\newcommand{\A}{\mathcal{A}}
\newcommand{\T}{\mathcal{T}}
\newcommand{\B}{\mathcal{B}}

\newcommand{\J}{J}

\newcommand{\N}{\mathcal{N}}
\newcommand{\M}{\mathcal{M}}

\newcommand{\be}{\mathbb{E}}
\newcommand{\bn}{\mathbb{N}}
\numberwithin{equation}{section}

\newcommand{\lei}[1]{\textcolor{red}{#1}}

\title{Stochastic Gradient Descent for Operator Learning in Hilbert Spaces: Convergence Rates and Minimax Lower Bounds$^\dag$\footnotetext{\dag~The work described in this paper is supported by the National Natural Science Foundation of China [Grants Nos.12171093 and 12061160462] and Shanghai Science and Technology Program [Project No. 21JC1400600]. Email addresses: jqyang24@m.fudan.edu.cn (J.-Q. Yang), leishi@fudan.edu.cn (L. Shi). The corresponding author is Lei Shi.}}

\author{Jia-Qi Yang}
\author{Lei Shi}
\affil{School of Mathematical Sciences and Shanghai Key Laboratory for
	Contemporary Applied Mathematics, Fudan University, Shanghai 200433, China.}

\date{}

\begin{document}
	\maketitle
\begin{abstract}
		This study investigates the use of stochastic gradient descent (SGD) to learn operators between general Hilbert spaces. We study weak and strong regularity conditions for the target operator that characterize its structure  and complexity. Under these conditions, we establish upper bounds for convergence rates of the SGD algorithm and derive a minimax lower bound analysis, further illustrating that our convergence analysis and regularity conditions quantitatively characterize the statistical difficulty of operator estimation under these regularity conditions. 
        The analysis extends to nonlinear regression targets under model misspecification, in which case SGD converges to the best linear approximation.
        Moreover, applying our analysis to operator learning problems based on vector-valued and scalar-valued reproducing kernel Hilbert spaces 
        yields new convergence results, thereby refining the conclusions of existing literature.
\end{abstract}


 
\section{Introduction}\label{section: introduction}
	
Operator learning has become an active area of research in machine learning research over recent years, asserting significant influence across disciplines such as economics, physics, computational biology, and engineering. In practical terms, functional data are often represented as vectors within an infinite-dimensional Hilbert space—a paradigm widely employed within physical and engineering contexts, notably within elastodynamics, thermodynamics, molecular dynamics, and turbulence \cite{kovachki2021neural,wang2023learning}. These domains often naturally represent data as functions. The supervised operator learning framework becomes relevant when a mapping from functional input to functional output is sought. For instance, one may aim to learn a mapping from parameters, such as boundary and initial conditions, to solutions of partial differential equations. Additionally, operator learning finds applications in surrogate approaches for structured output prediction, where the goal is to predict structured outputs \cite{weston2002kernel,kadri2013generalized,brouard2016input,ciliberto2020general}. These surrogate methods recast structured output outcomes into the Hilbert space, reframing the task as a regression of operators within infinite-dimensional output spaces. Structured prediction problems manifest in various applications, including image completion, label ranking, and graph prediction. These problems can be reformulated as supervised operator learning tasks.

One strategy for addressing supervised operator learning problems is to employ parameterized operator architectures, of which the most popular one is learning operators with neural networks. One of the first operator network architectures, proposed in a seminal study \cite{chen1995universal}, was motivated and  theoretically supported by a universal approximation theorem. This architecture was later developed into DeepONet \cite{lu2021learning}, engineered using multilayer feedforward neural networks. Following this development, variants of DeepONet \cite{jin2022mionet,lu2022comprehensive,wang2022improved} and practical applications \cite{cai2021deepm,di2021deeponet,lin2021operator} were subsequently proposed.  Another prominent architecture is the neural operator framework introduced by \cite{li2020neural}, primarily inspired by the solution form of partial differential equations and their associated Green's functions. Leveraging the Fourier convolution theorem to perform the integral transform in neural operators gave rise to the Fourier neural operator (FNO) \cite{li2020fourier}. Recently, \cite{boulle2023elliptic} utilized randomized numerical linear algebra theory to develop an algorithm that approximates the solution operators of three-dimensional uniformly elliptic partial differential equations (PDEs) at a near-optimal exponential rate. Additionally, \cite{schafer2024sparse} proposed an algorithm for learning elliptic boundary value problems, achieving the best known trade-off between accuracy 
$\epsilon$ and the number of required matrix-vector products.
Additional contemporary architectures encompass PCA-based representations \cite{bhattacharya2021model}, random feature approaches \cite{nelsen2021random}, wavelet approximations to integral transforms \cite{gupta2021multiwavelet}, attention-based architectures \cite{kissas2022learning} and structure-preserving operator networks \cite{bouziani2024structure}.

Non-parametric approaches are another category of architectures, among which the operator-valued kernel method is widely employed. This specific approach extends the scalar-valued kernel from the function learning of Reproducing Kernel Hilbert Space (RKHS) \cite{smale2006online,ying2008online} to operator learning within the RKHS of operators, utilizing the operator-valued reproducing kernel (to be introduced in Section \ref{3.2}). Previous research has extensively investigated this vector-valued kernel-based method, positing that it facilitates the learning of operators between general vector spaces under the assumption that the operator lies in a vector-valued RKHS \cite{micchelli2005learning,caponnetto2008universal,kadri2010nonlinear,kadri2016operator,owhadi2020ideas}. This theoretical framework is compelling due to its flexibility; it can accommodate both continuous and discrete inputs while the associated vector spaces are typically only required to be normed and separable. Recently, a rank-reduced method for solving least-squares problems, predicated on the operator-valued kernel, has been proposed and analyzed \cite{brogat2022vector}. This method has demonstrated superior performance in numerical simulations of structured prediction tasks compared to existing state-of-the-art approaches. 
In \cite{batlle2024kernel}, the authors have introduced a comprehensive kernel-based framework to facilitate operator learning. This framework employs operator-valued RKHSs and Gaussian processes to approximate mappings between infinite-dimensional spaces. It was then evaluated extensively with popular neural network approaches like DeepONet and FNO. Their findings indicate that even with the application of basic kernels, such as linear or Matérn, the kernel-based methodology demonstrates a competitive edge in balancing cost and accuracy. In the majority of benchmark tests, this method has achieved or exceeded the performance of neural network architectures.

This paper explores learning mappings between general Hilbert spaces using the stochastic gradient descent (SGD) algorithm. We have developed a theoretical framework that facilitates a convergence analysis for kernel-based operator learning. Our theoretical analysis, in turn, further elucidates the mathematical foundation of operator learning utilizing Hilbert-valued random variables. To this end, we first introduce the basic model as follows,
\begin{equation}\label{linear}
y=S^\dagger x+\epsilon, 
\end{equation}
where the input $x\in\H_{1}$ is a random element taking values in a real separable Hilbert space $(\H_{1},\langle,\rangle_{\H_{1}},\|\cdot\|_{\H_{1}})$, the output $y\in\H_{2}$ is also a random element taking values in another real separable Hilbert space $(\H_{2},\langle,\rangle_{\H_{2}},\|\cdot\|_{\H_{2}})$, 
and $\epsilon\in \H_2$ is a centered random noise independent of $x$ with finite variance, i.e., $\sigma^2:=\mathbb{E}[\|\epsilon\|_{\H_{2}}^2]<\infty$. We denote by $\B(\H_{1},\H_{2})$ the Banach space of bounded linear operators from $\H_{1}$ to $\H_{2}$ equipped with the operator norm and assume $S^\dagger\in\B(\H_{1},\H_{2})$. Throughout, we use the normalization $\mathbb E[\|x\|_{\H_1}^2]\le 1$, since any finite bound only affects constants.
One typical example of model \eqref{linear} is the random Fredholm equation where $\H_1=\H_2$ taken to be the space $\mathcal{L}^2[0,1]$ of square-integrable functions on $[0,1]$, and $S^\dagger$ is an integral operator defined as
\begin{equation*}
\begin{split}
S^\dagger : \H_1 &\to \H_2 \\
x &\to \int_0^1 S^\dagger(u,\cdot) x(u)du,
\end{split}
\end{equation*}
which is induced by a square-integrable function $S^\dagger(\cdot,\cdot)$ on $[0,1]^2$. The study of random Fredholm equations dates back to the early 1960s \cite{bharucha1960random} and encompasses a range of important models, including the functional historical model. The study of such models, initially unveiled in a seminal work \cite{malfait2003historical}, furnishes practical applications for the analysis of data from physical, biological, and economic domains. We would like to point out that model \eqref{linear} is more general and embraces a wide range of applications. Though $S^\dagger$ is a linear operator operating within separable Hilbert spaces, it can also be remodeled by employing operator-valued kernel methods, enabling it to act as an operator-valued regression model capable of approximating nonlinear operators. Our theoretical analysis remains applicable in this context (refer to Section \ref{section: related work}). The adaptability of model \eqref{linear} allows it to extend into a functional linear regression model based on scalar-valued kernels \cite{ying2008online,chen2022online,guo2022capacity,guo2022rates}, wherein $y$ is a scalar response (refer to Subsection \ref{3.3}). These are effectively incorporated within our theoretical framework.

This study investigates the application of SGD in solving model (\ref{linear}). We say an operator $A \in \mathcal{B}(\H_1,\H_2)$ is Hilbert-Schmidt if $\sum_{k \geq 1}\left\|A e_k\right\|_{\H_2}^2<\infty$ for some (any) orthonormal basis $\left\{e_k\right\}_{k \geq 1}$ of $\H_1$. The space of Hilbert-Schmidt operators, denoted by $\B_{\mathrm{HS}}(\H_1,\H_2)$, constitutes a Hilbert space equipped with the inner product $\langle A, B\rangle_{\mathrm{HS}}:=\sum_{k\geq 1}\left\langle A e_k, B e_k\right\rangle_{\H_2}$, $\forall A, B \in \B_{\mathrm{HS}}(\H_1,\H_2)$ and $\|\cdot\|_{\mathrm{HS}}$ is the induced norm.  In particular, $\B_{\mathrm{HS}}(\H_1,\H_2)$ is a subspace of $\B (\H_1,\H_2)$ as the operator norm is bounded by the Hilbert-Schmidt norm. To derive an estimator for $S^\dagger$, we  consider a least-squares problem (convex in $S$) that minimizes $\mathcal{E}(S):=\mathbb{E}\|y-S x\|_{\H_2}^2$ across all $S \in\B_{\mathrm{HS}}(\H_{1},\H_{2})$. 
The risk functional is Fréchet differentiable \cite{dunford1988linear} on \(\mathcal B_{\rm HS}(H_1,H_2)\), with Hilbert–Schmidt gradient
$\nabla\mathcal{E}(S)= 2\mathbb{E}[(S x-y)\otimes x]$, wherein $\otimes$ represents the tensor product for any $e \in \H_1$ and $f\in \H_2$, defined by $f \otimes e := \langle e, \cdot\rangle_{\H_1} f$.
Let \(\{(x_t,y_t)\}_{t\ge1}\) be an i.i.d. sequence with the same distribution as \((x,y)\) 
in model \eqref{linear}, an unbiased estimator of $\nabla\mathcal{E}(S)$ is $2(S x_t-y_t)\otimes x_t$. Consequently, the SGD algorithm for estimating $S^\dagger$ is defined by $S_1=\mathbf{0}$, and 
\begin{equation} \label{sgd}
	S_{t+1}=S_{t}-\eta_t(S_tx_t-y_t)\otimes x_t,\quad t\geq 1,
\end{equation}
where $\eta_t>0$ is the step size, with the factor $2$ in
the stochastic gradient absorbed into $\eta_t$. 
Throughout the paper, $\mathbf{0}$ denotes the zero element of the relevant vector or operator space, whereas $0$ denotes the scalar zero in $\mathbb{R}$.
To evaluate the performance of the estimator $S_{t+1}$, we introduce the prediction error and the estimation error. Prediction error is represented by $\mathcal{E}(S_{t+1})-\mathcal{E}(S^{\dagger})$, equivalent to the square of a semi-norm of $S_{t+1}-S^\dagger$ expressed as $\left\|(S_{t+1}-S^\dagger)L_C^{1/2}\right\|^2_{\mathrm{HS}}$ (refer to \eqref{form2}), where $L_C:=\mathbb{E}[x\otimes x]$ is the uncentered covariance operator. Additionally, $S_{t+1}\in \B_{\mathrm{HS}}(\H_1,\H_2)$ for $t\geq 1$, the estimation error is then defined as $\|S_{t+1}-S^\dagger\|^2_{\mathrm{HS}}$, provided $S^\dagger \in \B_{\mathrm{HS}}(\H_1,\H_2)$.

In this paper, we provide a rigorous convergence analysis of the SGD algorithm \eqref{sgd}, utilizing appropriate step sizes $\{\eta_t\}_{t=1}^T$, with a focus on prediction and estimation errors. Two general types of step sizes are considered: a decreasing step size, for instance, a polynomially decaying step size represented as $\eta_t=\eta_1 t^{-\theta}$ with $\theta \in(0,1)$ and some $\eta_1>0$, and a horizon-dependent constant step size, depicted as $\eta_t=\eta(T)$, which relies on the total number of iterations $T$, namely, the sample size processed by the SGD algorithm after $T$ iterations. These two types of step sizes are preferred in many practical application scenarios. Additionally, there are some adaptive step size selection methods, but these are outside the scope of this paper. Previous theoretical research on SGD has primarily focused on these two step sizes, demonstrating that these step sizes play an implicit regularization role in the algorithm, thereby enhancing the algorithm's generalization and robustness \cite{smale2006online}. It is important to note that a decreasing step-size schedule can be specified without knowing a terminal horizon and can therefore be used for an indefinitely continuing data stream. Therefore, it is not necessary to know the total sample size $T$ in advance when choosing the step size. This type of step size is more applicable in scenarios requiring real-time iterative updates. Our study is especially driven by scenarios involving infinite-dimensional $\mathcal{H}_2$, as observed in various practical applications such as functional linear regression with functional response, non-parametric operator learning with vector-valued kernel, inference for Hilbertian time series, and distribution regression \cite{bosq2000linear,crambes2013asymptotics,szabo2016learning,brogat2022vector}. The study of linear operator learning on infinite-dimensional spaces has garnered much interest due to the rise of generative learning models and machine learning-based PDE solvers. However, existing literature predominantly deals with rather specific problem settings and assumptions. Recently, from the standpoint of inverse problems, \cite{mollenhauer2022learning} established a comprehensive mathematical framework for model \eqref{linear} regarded as a fundamental model of infinite-dimensional regression, thereby providing solid theoretical foundations for its applications in the scenarios mentioned above. The work \cite{persson2024randomized} studied the infinite-dimensional analog of the Nyström approximation to compute low-rank approximations to non-negative self-adjoint trace-class operators. 


Motivated by these developments, this paper studies single-pass, last-iterate SGD for direct linear operator regression between separable Hilbert spaces. 
We establish prediction-error bounds under both weak and strong regularity conditions, and Hilbert--Schmidt estimation bounds under strong regularity. 
In addition, we establish a more refined convergence analysis by introducing the spectral decay condition of the covariance operator, a condition employed to further reflect the regularity of the input vector. We also establish minimax lower bounds under a trace-type spectral condition, without prescribing a polynomial eigenvalue decay rate. The bounds show the minimax optimality of constant-step SGD for Hilbert–Schmidt estimation under strong regularity and reveal a distinct difficulty under weak regularity with infinite-dimensional outputs. Our framework further provides unified prediction and estimation guarantees for vector- and scalar-valued kernel learning and functional linear models.


The rest of this paper is structured as follows: Section \ref{section: main results} delineates our main results, along with essential assumptions and notations. It also extends our analysis to nonlinear targets. Section \ref{section: related work} considers a more general model with a bias term. Additionally, we apply our analysis to two other significant settings, namely learning with vector-valued and scalar-valued reproducing kernel Hilbert spaces, including functional linear regression models. Section \ref{section: Error decomposition} introduces a comprehensive decomposition of the prediction and estimation errors associated with the algorithm (\ref{sgd}), accompanied by basic estimates. The derivations of upper bounds are detailed in Sections \ref{proof1} and \ref{proof2}, while minimax lower bounds are established in Section \ref{Section:lower bounds}. Section \ref{Conclusion and Discussion} discusses our future work. Some proofs are deferred to the Appendix.

\section{Main Results}\label{section: main results}	
	
This section presents our main results on the convergence rates of prediction and estimation errors for the SGD algorithm (\ref{sgd}) under various regularity conditions. We also establish minimax lower bounds, demonstrating that some of our convergence upper bounds are optimal. 

We begin with some notations. Recall that 
$\mathcal{B}(\H_{1},\H_{2})$ denotes the collection of all bounded linear operators from $\H_{1}$ to $\H_{2}$ which is a Banach space with respect to the operator norm $\|\cdot\|$, and $\B_{\mathrm{HS}}(\H_1,\H_2)$ represents the space of Hilbert-Schmidt operators from $\H_{1}$ to $\H_{2}$, equipped with the Hilbert-Schmidt norm $\|\cdot\|_{\mathrm{HS}}$. We denote $\mathrm{Tr}(L)$ as the trace of $L \in \mathcal{B}(\H_{1},\H_{1})$, which is the sum of all associated eigenvalues of $L$, assuming $L$ is a positive self-adjoint compact operator. The uncentered covariance operator of $x$ is $L_C = \mathbb{E}[x \otimes x]$, where $x \otimes x$ is a tensor product operator belonging to $\B_{\mathrm{HS}}(\H_1,\H_1)$. The condition $\mathbb{E}[\|x\|_{\H_{1}}^2]\leq1$, which we assume throughout this paper, ensures that $\mathbb{E}[\|x \otimes x\|_{\mathrm{HS}}] = \mathbb{E}[\|x\|^2_{\H_{1}}] \leq 1$. Employing Bochner's integral \cite{yosida2012functional}, we confirm that $L_C$ is well-defined as a Hilbert-Schmidt operator with 
$$\|L_C\| \leq \|L_C\|_{\mathrm{HS}} \leq \mathbb{E}\left[\|x \otimes x\|_{\mathrm{HS}}\right] \leq 1.$$ Additionally, $L_C$ is self-adjoint, nonnegative, and compact. Consequently, for any $\alpha > 0$, the $\alpha$-th power of $L_C$, denoted as $L_C^\alpha$, is well-defined. Moreover, it is evident that 
$$\left\|L_C^{1/2}\right\|^2_{\mathrm{HS}} = \mathrm{Tr}(L_C) = \sum_{k\geq 1} \langle L_C e_k, e_k \rangle_{\H_1} = \sum_{k\geq 1}\mathbb{E} \left[ \langle x, e_k \rangle^2_{\H_1}\right] = \mathbb{E}[\|x\|_{\H_{1}}^2] \leq 1, $$ where $\{e_k\}_{k \geq1}$ is the orthonormal basis of $\H_1$. Therefore, $L_C^{1/2}$ is a Hilbert-Schmidt operator and $L_C$ is of trace class.

\subsection{Regularity Assumptions} \label{subsection: assumption}

In this subsection, we present the regularity assumptions for our main
theorems. We first clarify that
$S^{\dagger}\in\B_{\mathrm{HS}}(\H_1,\H_2)$ is not required to derive
\eqref{sgd}. Under the model and moment assumptions in
Section~\ref{section: introduction}, $x$ and $y$ have finite second
moments. For any $S\in\B_{\mathrm{HS}}(\H_1,\H_2)$, Cauchy--Schwarz
gives $\mathbb{E}[\|(Sx-y)\otimes x\|_{\mathrm{HS}}]<\infty$, so
$\mathbb{E}[(Sx-y)\otimes x]$ is well defined as a Bochner integral
in $\B_{\mathrm{HS}}(\H_1,\H_2)$. For any
$\Delta S\in\B_{\mathrm{HS}}(\H_1,\H_2)$,
\begin{equation*}
\begin{aligned}
\mathcal{E}(S+\Delta S)-\mathcal{E}(S)
&=2\left\langle
\mathbb{E}[(Sx-y)\otimes x],\Delta S
\right\rangle_{\mathrm{HS}}\\
&\quad+\mathbb{E}\|\Delta Sx\|_{\H_2}^2,
\end{aligned}
\end{equation*}
where $\mathbb{E}\|\Delta Sx\|_{\H_2}^2
\le\mathbb{E}\|x\|_{\H_1}^2\|\Delta S\|_{\mathrm{HS}}^2$.
Thus, $\mathcal{E}$ is Fr\'echet differentiable on
$\B_{\mathrm{HS}}(\H_1,\H_2)$, with Hilbert--Schmidt gradient
$\nabla\mathcal{E}(S)=2\mathbb{E}[(Sx-y)\otimes x]$.
Since $S_1=0$ and each increment $S_{t+1}-S_t$ has rank at most one,
all iterates are finite-rank operators almost surely. Consequently,
\eqref{sgd} remains an SGD iteration in
$\B_{\mathrm{HS}}(\H_1,\H_2)$ even when the target is not Hilbert--Schmidt.

Prediction error remains well defined for bounded targets: under the
linear model, every $S\in\B(\H_1,\H_2)$ satisfies
\begin{equation*}
\begin{aligned}
\mathcal{E}(S)-\mathcal{E}(S^{\dagger})
&=\|(S-S^{\dagger})L_C^{1/2}\|_{\mathrm{HS}}^2\\
&\le\|S-S^{\dagger}\|^2\operatorname{Tr}(L_C)<\infty.
\end{aligned}
\end{equation*}
Moreover, let $(P_n)_{n\geq1}$ be finite-rank orthogonal projections
on $\H_1$ converging strongly to the identity. Then
$S^{\dagger}P_n\in\B_{\mathrm{HS}}(\H_1,\H_2)$ and, by dominated
convergence, $\mathcal{E}(S^{\dagger}P_n)\to\mathcal{E}(S^{\dagger})$.
Hence,
$\inf_{S\in\B_{\mathrm{HS}}(\H_1,\H_2)}\mathcal{E}(S)
=\mathcal{E}(S^{\dagger})$, although the infimum need not be attained
in this space. In contrast, since each $S_t$ is Hilbert--Schmidt,
$S_t-S^{\dagger}$ is Hilbert--Schmidt if and only if $S^{\dagger}$
is Hilbert--Schmidt. This distinction motivates the weak and strong
regularity conditions introduced below.

\begin{assumption}[Weak regularity condition of $S^\dagger$] \label{a1}
There exists a bounded operator $\J$ in $\B(\H_1,\H_2)$ and a positive parameter $r>0$ such that
\begin{equation*}
S^{\dagger}=\J L_C^r.
\end{equation*}
\end{assumption}

Assumption \ref{a1} is instrumental in estimating prediction error. In this context, a larger value of $r>0$ strengthens the assumption, meaning that $S^\dagger$ has higher regularity.  Given that $L_C$ is compact, $S^\dagger$ is thus a compact operator. Furthermore, if $r\geq1/2$, $S^\dagger$ qualifies as a Hilbert-Schmidt operator, given that $L_C^{1/2}\in\B_{\mathrm{HS}}(\H_1,\H_1)$. Assumption \ref{a1} is more general, considering that $S^\dagger$ might not be a Hilbert-Schmidt operator when $r<1/2$. 
This assumption is particularly useful for deriving prediction-error bounds when the target need not be Hilbert--Schmidt.
However, under Assumption~\ref{a1}, if $\H_2$ is finite-dimensional, or more generally if $\operatorname{Ran}(S^\dagger)$ is finite-dimensional, then one can find a Hilbert--Schmidt operator $\widetilde{\J}\in\B_{\mathrm{HS}}(\H_1,\H_2)$ and some $\widetilde r>0$ such that
$S^\dagger=\widetilde{\J}L_C^{\widetilde r}.$ This leads to Assumption~\ref{a4} and allows us to derive a sharper error bound.

\begin{assumption}[Strong regularity condition of $S^\dagger$]\label{a4}
		There exists a Hilbert-Schmidt operator $\widetilde{\J}\in \B_{\mathrm{HS}}(\H_1,\H_2)$ and a positive parameter $\widetilde r>0$ such that
		\begin{equation*}
			S^{\dagger}=\widetilde{\J} L_C^{\widetilde r}.
		\end{equation*}
	\end{assumption}

If $S^\dagger$ is a finite-rank operator, then Assumptions \ref{a1} and \ref{a4} are equivalent. A prevalent example occurs when the output space $\H_2$ is finite-dimensional, such as $\H_2=\mathbb{R}$ in the functional linear regression model (see Section \ref{3.3} for details). Given Assumption \ref{a4}, $S^\dagger$ qualifies as a Hilbert-Schmidt operator, enabling us to derive tighter estimation and prediction bounds relative to those attainable under Assumption \ref{a1}.
 
\begin{assumption}[Spectral decay condition of $L_C$]\label{a2}
		\begin{equation*}
			\mbox{$\mathrm{Tr}(L_C^s) < \infty$ for some $0<s\leq1$}.
		\end{equation*}
	\end{assumption}

Assumption \ref{a2} implies that the decreasing sequence of the eigenvalues of the covariance operator $L_C$, denoted by $\{\lambda_k\}_{k\geq1}$, decays at least as rapidly as $\frac{ \left[\mathrm{Tr}(L^s_C)\right]^{1/s}}{k^{1/s}}$. Hence, the smaller the value of $s$, the stronger the assumption, signifying a faster decay rate of the eigenvalues of $L_C$. Moreover, when the eigenvalues of $L_C$ decay exponentially, Assumption \ref{a2} holds for any $s>0$. 
It should be noted that this assumption is always satisfied for any $s\geq1$ since $\mathrm{Tr}(L_C^s)\leq\mathrm{Tr}(L_C)\left\|L_C^{s-1}\right\|\leq 1$. This eigenvalue decay assumption was initially introduced in \cite{dieuleveut2017harder} and subsequently adopted in \cite{guo2019fast,pillaud2018statistical,guo2022rates,guo2022capacity} to better reflect the regularity of high-dimensional or even infinite-dimensional Hilbert-valued random variables, thereby aiding in the establishment of dimension-free and often tight convergence analysis.

\begin{assumption}[Moment condition]\label{a3}
    There exists a constant $c>0$ such that for every
bounded linear operator $A \in \mathcal{B}(\H_{1},\H_{i})$,
    \begin{equation*}
        \mathbb{E}\left[\left\|A x\right\|^4_{\H_i}\right] \leq c\left(\mathbb{E}\left[\|A x\|^2_{\H_i}\right]\right)^2, \text{ where } i=1,2.
    \end{equation*}    
\end{assumption}

We now present a simpler equivalent statement of Assumption \ref{a3} alongside a novel sufficient condition.
\begin{proposition} \label{1}
For any fixed $c>0$, Assumption~\ref{a3} holds with constant $c$ if and
only if, for every $f\in\H_1$,
    \begin{equation*}
        \mathbb{E}\left[\left\langle x,f\right\rangle_{\H_{1}}^4\right]
        \leq c\left(\mathbb{E}\left[\left\langle x,f\right\rangle_{\H_{1}}^2\right]\right)^2.
    \end{equation*}
    Furthermore, Assumption \ref{a3} is satisfied if $x$ is strictly sub-Gaussian in $\H_1$.
\end{proposition}

The proof of Proposition \ref{1} and the definition of strictly sub-Gaussian random variables are provided in Appendix \ref{Appendix A}. Proposition \ref{1} implies that all linear functionals of $x$ have bounded kurtosis. A similar assumption has been adopted in several studies \cite{yuan2010reproducing,cai2012minimax,chen2022online,guo2022capacity}, particularly where $\H_1=L^2[0,1]$ and $\H_{2}=\mathbb{R}$. Notably, Assumption \ref{a3} is satisfied when $x$ represents a Gaussian random element in a general Hilbert space \cite{da2014introduction} or a Gaussian process in $L^2[0,1]$. 

It is also worth noting that the sub-Gaussian condition adopted in \cite[Assumption~2.1]{reiss2020nonasymptotic} implies the moment condition characterized in Proposition~\ref{1}. Indeed, taking $k=4$ in their assumption yields
\[
\mathbb{E}\big[\langle x,f\rangle_{\H_1}^4\big]
\leq
16C_1^4
\left(
\mathbb{E}\big[\langle x,f\rangle_{\H_1}^2\big]
\right)^2,
\]
so the condition in Proposition~\ref{1} holds with $c=16C_1^4$.

\subsection{Upper Bounds on Convergence Rates}\label{subsection: upper rates}

In this subsection, we present the upper bounds for the convergence of errors when solving model \eqref{linear} using the SGD algorithm \eqref{sgd}. Our established convergence results are non-asymptotic, and these outcomes describe the convergence rate for prediction error and estimation error after $T$ iterations through the selection of appropriate step sizes (including decaying and constant step sizes). In the proofs of these results, we also provide estimates for the constants in the upper bounds and further clarify the dependence of these constants on the regularity parameters  (namely, $r$, $\widetilde r$, and $s$). Recall that $S^\dagger$ is the target operator in model \eqref{linear}. The prediction error and estimation error are respectively denoted as $\mathcal{E}(S_{t+1})-\mathcal{E}(S^\dagger)$ and $\left\|S_{t+1}-S^{\dagger}\right\|_{\mathrm{HS}}^2$, where $\mathcal{E}(S)=\mathbb{E}\|y-S x\|_{\H_2}^2$ for any $S\in\B(\H_1,\H_2)$.  For $k \in \mathbb{N}$, let $\mathbb{E}_{z_1, \cdots, z_k}$ denote taking expectation with respect to $\{z_i:=(x_i,y_i)\}_{i=1}^k$, which is written as $\mathbb{E}_{z^k}$ for short.
Denote $\mathbb{N}_T$ as the set $\{1,2,\cdots, T\}$.

\noindent\textbf{Input rescaling.}
The normalization $\mathbb{E}\|x\|_{\H_1}^2\le1$ can be removed
from the upper-bound results by rescaling the input.
Suppose instead that
$\mathbb{E}\|x\|_{\H_1}^2\le\kappa_x^2$ for some $\kappa_x>0$,
and define
\begin{equation*}
\bar x=\kappa_x^{-1}x,\qquad
\bar L_C=\kappa_x^{-2}L_C,\qquad
\bar S^\dagger=\kappa_x S^\dagger.
\end{equation*}
Then $y=\bar S^\dagger\bar x+\epsilon$ and
$\mathbb{E}\|\bar x\|_{\H_1}^2\le1$.
If $\bar\eta_t$ is an admissible step-size schedule for the
corresponding normalized theorem, use
$\eta_t=\kappa_x^{-2}\bar\eta_t$ in the original iteration.
Starting from zero, the normalized and original iterates satisfy
$\bar S_t=\kappa_x S_t$, and hence
$\bar S_t\bar x=S_tx$. Prediction errors are therefore unchanged,
whereas, whenever $S^\dagger$ is Hilbert--Schmidt, $\|S_t-S^\dagger\|_{\mathrm{HS}}^2
=\kappa_x^{-2}
\|\bar S_t-\bar S^\dagger\|_{\mathrm{HS}}^2.$

The regularity exponents and the moment constant are unchanged.
More precisely, the operators $J$ and $\widetilde J$
in Assumptions \ref{a1} and \ref{a4} become
$\bar J=\kappa_x^{2r+1}J$ and
$\overline{\widetilde J}
=\kappa_x^{2\widetilde r+1}\widetilde J$,
respectively, so that 
\begin{equation*}
\bar S^\dagger=\bar J\bar L_C^r
\quad\text{or}\quad
\bar S^\dagger=
\overline{\widetilde J}\bar L_C^{\widetilde r},
\qquad
\operatorname{Tr}(\bar L_C^s)
=\kappa_x^{-2s}\operatorname{Tr}(L_C^s).
\end{equation*}
Consequently, the powers of $T+1$ and the logarithmic factors in
the upper bounds remain unchanged. 

The first main result establishes the convergence rates of prediction error when using algorithm \eqref{sgd} equipped with decaying step sizes. 

\begin{theorem} \label{t1}
Define $\{S_t\}_{t \in \mathbb{N}_{T+1}}$ by (\ref{sgd}) with \(T\ge2\). Under Assumption \ref{a1} with $r>0$, Assumption \ref{a2} with $0<s\leq1$, and Assumption \ref{a3} with $c>0$, let $\left\{\eta_t=\eta_1 t^{-\theta}\right\}_{t\in  \mathbb{N}_T}$ with $0<\theta<1$ and $\eta_1$ satisfying 
\begin{equation} \label{condition}
			c\left[3c_5(\eta_1,1,\theta)\max\left\{\frac{1}{e\theta},1\right\}+\eta_1^2\right]<1
			\text{\  and\ } \eta_1\|L_C\|<1,
		\end{equation} where $c_5(\eta_1,1,\theta)$ will be specified by Proposition \ref{eta1}.
\begin{itemize}
			\item[(1)]	If $s=1$, choose $\theta=\min\left\{\frac{2r}{2r+1},\frac{1}{2}\right\}$. Then
			\begin{equation*}
				\mathbb{E}_{z^T}[\mathcal{E}(S_{T+1})-\mathcal{E}(S^{\dagger})]
				\leq c_r(T+1)^{-\theta}\log(T+1).
			\end{equation*}
			\item[(2)]
			If $0<s<1$, choose 
			\[
			\theta=\min\left\{\frac{2r+1-s}{2r+2-s},\frac{2-s}{3-s}\right\}=
			\begin{cases}
				\frac{2r+1-s}{2r+2-s}, & \text{when } r<\frac{1}{2}, \\
				\frac{2-s}{3-s}, & \text{when } r\geq\frac{1}{2}.
			\end{cases}
			\] Then
			\[
			\be_{z^{T}}\left[\mathcal{E}(S_{T+1})-\mathcal{E}(S^{\dagger})\right]
			\leq c_{r,s}(T+1)^{-\theta}.
			\]
\end{itemize}
		Here the constants $c_r$ and $c_{r,s}$ are independent of $T$ and will be given in the proof.
	\end{theorem}

 It should be pointed out that, as previously discussed, the condition $\mathbb{E}\left[\|x\|^2_{H_1}\right]\leq 1$ ensures that Assumption \ref{a2} is always satisfied for $s=1$. Therefore, in this scenario, we do not impose additional restrictions on the spectral decay rate of the covariance operator $L_C$. 
 The upper bounds in Theorem \ref{t1} exhibit saturation at $r=1/2$:
their decay exponents no longer increase with $r$ beyond this
threshold. Taking $0<s<1$ improves the decay exponent but does
not change this threshold.

	
The second main result establishes the convergence rates of prediction error when using algorithm \eqref{sgd} equipped with constant step sizes.

\begin{theorem} \label{t2}
Define $\{S_t\}_{t \in \mathbb{N}_{T+1}}$ by (\ref{sgd}) with $T\geq2$. Under Assumption \ref{a1} with $r>0$, Assumption \ref{a2} with $0<s\leq1$, and Assumption \ref{a3} with $c>0$, choose $\left\{\eta_t=\eta_*(T+1)^{-\frac{2r+1-s}{2r+2-s}}\right\}_{t\in \bn_T}$ with $0<\eta_*\leq\frac{e(2r+1-s)}{(1+14c)(2r+2-s)}$. It holds that
		\begin{equation*}
			\mathbb{E}_{z^{T}}[\mathcal{E}(S_{T+1})-\mathcal{E}(S^{\dagger})]
			\leq \widetilde{c}_{r,s}
			\begin{cases}
				(T+1)^{-\frac{2r}{2r+1}}\log(T+1), &\text{if }s=1,\\
				(T+1)^{-\frac{2r+1-s}{2r+2-s}}, &\text{if }0<s<1.
			\end{cases}
		\end{equation*}
		Here the constant $\widetilde{c}_{r,s}$ is independent of $T$ and will be given in the proof.
	\end{theorem}	

    In contrast, the upper bound in Theorem \ref{t2}, obtained with
horizon-dependent constant step sizes, exhibits no saturation
in $r$: its decay exponent increases with $r$ for fixed $s$.
Taking $0<s<1$ further improves the decay exponent, consistently with the phenomenon revealed by Theorem \ref{t1}.


In the following two theorems, we establish the convergence analysis of the SGD algorithm \eqref{sgd} under the stronger regularity assumption of the target operator (namely, Assumption \ref{a4}). Theorem \ref{t5} pertains to decaying step sizes, while Theorem \ref{t6} addresses constant step sizes.
	
\begin{theorem} \label{t5}
Define $\{S_t\}_{t \in \mathbb{N}_{T+1}}$ by (\ref{sgd}) with \(T\ge2\). Under Assumption \ref{a4} with $\widetilde r>0$, Assumption \ref{a2} with $0<s\leq1$, and Assumption \ref{a3} with $c>0$, let $\left\{\eta_t=\eta_1 t^{-\theta}\right\}_{t\in  \mathbb{N}_T}$ with $\eta_1$ satisfying \eqref{condition} and  
		\[
		\theta=\min\left\{\frac{2-s}{3-s},\frac{2\widetilde{r}+1}{2\widetilde{r}+2}\right\}=
		\begin{cases}
			\frac{2-s}{3-s}, & \text{when }1-2\widetilde{r}<s\leq1,\\
			\frac{2\widetilde{r}+1}{2\widetilde{r}+2}, &\text{when } 0<s\leq1-2\widetilde{r}.
		\end{cases}
		\]
		Then
		\begin{equation*}
			\mathbb{E}_{z^{T}}[\mathcal{E}(S_{T+1})-\mathcal{E}(S^{\dagger})]
			\leq c_{\widetilde{r},s}'
			\begin{cases}
				(T+1)^{-\frac12}\log(T+1), &\text{if } s=1, \\
				(T+1)^{-\frac{2-s}{3-s}}, &\text{if } 1-2\widetilde{r}<s<1, \\
				(T+1)^{-\frac{2\widetilde{r}+1}{2\widetilde{r}+2}}, &\text{if } 0<s\leq1-2\widetilde{r}.
			\end{cases}
		\end{equation*}
		Here the constant $c_{\widetilde{r},s}'$ is independent of $T$ and will be given in the proof.
	\end{theorem}
	
	\begin{theorem} \label{t6}
        Define $\{S_t\}_{t \in \mathbb{N}_{T+1}}$ by (\ref{sgd}) with $T\geq2$. Under Assumption \ref{a4} with $\widetilde r>0$, Assumption \ref{a2} with $0<s\leq1$, and Assumption \ref{a3} with $c>0$, choose step-sizes $\left\{\eta_t=\eta_{*}(T+1)^{-\frac{2\widetilde{r}+1}{2\widetilde{r}+2}}\right\}_{t\in \bn_T}$ with $0<\eta_*\leq\frac{e(2\widetilde{r}+1)}{(1+14c)(2\widetilde{r}+2)}$. We have
		\begin{equation*}
			\mathbb{E}_{z^{T}}[\mathcal{E}(S_{T+1})-\mathcal{E}(S^{\dagger})]
			\leq c_{\widetilde{r},s}''
			\begin{cases}
				(T+1)^{-\frac{2\widetilde{r}+1}{2\widetilde{r}+2}}\log(T+1), &\text{if }s=1,\\
				(T+1)^{-\frac{2\widetilde{r}+1}{2\widetilde{r}+2}}, &\text{if }0<s<1.
			\end{cases}
		\end{equation*}
		Here the constant $c_{\widetilde{r},s}''$ is independent of $T$ and will be given in the proof.
	\end{theorem}
	
The two theorems above reveal that the convergence behavior of the prediction error under the strong regularity condition (i.e., Assumption \ref{a4}) is similar to its performance under the weak regularity condition (i.e., Assumption \ref{a1}). 
The prediction upper bound in Theorem \ref{t5} exhibits saturation
in $\widetilde r$, whereas that in Theorem \ref{t6} does not.
In fact, the same issue has been observed in other studies on the SGD algorithm in various scenarios \cite{ying2008online,MR3519927,guo2019fast,guo2022capacity}. Additionally, in Theorem \ref{t6}, taking $0<s<1$ removes the logarithmic factor
from the upper bound without changing its polynomial exponent.


Finally, under Assumptions \ref{a1} and \ref{a2}, if $r>s/2$, then
Assumption \ref{a4} holds with $\widetilde r=r-s/2$ and
$\widetilde J=JL_C^{s/2}$. Indeed,
\[
S^\dagger=\widetilde J L_C^{\widetilde r},
\qquad
\|\widetilde J\|_{\mathrm{HS}}
\le \|J\|\bigl[\operatorname{Tr}(L_C^s)\bigr]^{1/2}
<\infty.
\]
With this substitution, the prediction bounds in
Theorems \ref{t5} and \ref{t6} have the same powers of $T+1$ and
logarithmic factors as those in Theorems \ref{t1} and \ref{t2},
respectively, although the multiplicative constants may differ.


At the end of this subsection, we derive the convergence of the estimation error of the SGD algorithm \eqref{sgd} based on Assumption \ref{a4}. The convergence of the estimation error is in the sense of the Hilbert-Schmidt norm, which is a strong form of convergence compared to the convergence of the prediction error. 
	
We first formulate the convergence result on the estimation error with decaying step sizes.

\begin{theorem} \label{t3}
 Define $\{S_t\}_{t \in \mathbb{N}_{T+1}}$ by (\ref{sgd}) with \(T\ge2\). Under Assumption \ref{a4} with $\widetilde r>0$, Assumption \ref{a2} with $0<s<1$, and Assumption \ref{a3} with $c>0$, let $\left\{\eta_t=\eta_1 t^{-\theta}\right\}_{t\in  \mathbb{N}_T}$ with $\eta_1$ satisfying \eqref{condition} and 
		\[\theta=\min \left\{\frac{2\widetilde r +s}{1+2 \widetilde r +s},\frac{1}{2}\right\}=
		\begin{cases}
			\frac{2\widetilde{r}+s}{1+2\widetilde{r}+s}, &\text{when } \widetilde{r}<\frac{1-s}{2}, \\
			\frac{1}{2}, & \text{when }\widetilde{r}\geq\frac{1-s}{2}.
		\end{cases}
		\] Then 
		\begin{equation*}
			\be_{z^{T}}\left[\left\|S_{T+1}-S^{\dagger}\right\|_{\mathrm{HS}}^2\right]
			\leq c_{\widetilde{r},s}
			\begin{cases}
				(T+1)^{-\frac{2\widetilde{r}}{1+2\widetilde{r}+s}}, &\text{if }\widetilde{r}<\frac{1-s}{2}, \\
				(T+1)^{-\frac{1-s}{2}}\log(T+1), &\text{if }\widetilde{r}\geq\frac{1-s}{2}.
			\end{cases}
		\end{equation*}
		Here the constant $c_{\widetilde{r},s}$ is independent of $T$ and will be given in the proof.		
\end{theorem}
	
	It should be pointed out that our analysis cannot yield the convergence rate of SGD algorithm \eqref{sgd} with decaying step sizes for the case of $s=1$.  
    The present proof establishes the result only for \(0<s<1\). Whether the case \(s=1\) is unattainable for decaying-step SGD or merely requires a sharper argument is left open.
    
 
 The last theorem of this subsection focuses on the convergence of estimation error with constant step sizes.

\begin{theorem} \label{t4}
 Define $\{S_t\}_{t \in \mathbb{N}_{T+1}}$ by (\ref{sgd}) with $T\geq2$. Under Assumption \ref{a4} with $\widetilde r>0$, Assumption \ref{a2} with $0<s\leq1$, and Assumption \ref{a3} with $c>0$, choose step-sizes $\left\{\eta_{t}=\eta_{*}(T+1)^{-\frac{2\widetilde{r}+s}{1+2\widetilde{r}+s}}\right\}_{t\in \bn_T}$ with $0<\eta_{*}\leq\frac{e(2\widetilde{r}+s)}{(1+14c)(1+2\widetilde{r}+s)}$.
 There holds
		\begin{equation*}
			\be_{z^{T}}\left[\left\|S_{T+1}-S^{\dagger}\right\|_{\mathrm{HS}}^2\right]\leq
			\widetilde{c}_{\widetilde{r},s}(T+1)^{-\frac{2\widetilde{r}}{1+2\widetilde{r}+s}}.
		\end{equation*}
Here the constant $\widetilde{c}_{\widetilde{r},s}$ is independent of $T$ and will be given in the proof.
	\end{theorem}

    For fixed $0<s<1$, the estimation upper bound in Theorem \ref{t3}
exhibits saturation at $\widetilde r=(1-s)/2$, whereas that
in Theorem \ref{t4} does not.
In the following subsection, combining our established minimax lower bound estimates, we will further discuss the optimality of the above convergence rates.

\subsection{Minimax Lower Bounds}\label{subsection: minimax}
	
In this subsection, we present the established minimax lower bounds. We first further describe the data generation mechanism. As introduced in Section \ref{section: introduction}, the sample $\{x_t,y_t\}_{t\geq 1}$ is a sequence of independent and identically distributed (i.i.d.) copies of $(x,y)$, where $y\in \H_2$ is related to $x \in \H_1$ through model \eqref{linear}. To further characterize the randomness of $(x,y)$, we assume that $(x,y)$ follows a joint probability distribution $\rho$ on $\H_1 \times \H_2$, where $x$ follows the marginal distribution $\rho_x$ of $\rho$ on $\H_1$, and the conditional distribution of $\rho$ (for a given $x$) characterizes the randomness of $y$ and satisfies $\mathbb{E}[y|x]=S^{\dagger}x$. Let $\mathbf{z}=\left\{(x_i,  y_i)\right\}_{i=1}^T$ be i.i.d. samples of $\rho^{\otimes T}$ and $S_{\mathbf{z}}$ be an estimator of the target operator $S^\dagger$, i.e., a measurable mapping from $\mathbf{z}$ to $\B(\H_1,\H_2)$ or $\B_{\mathrm{HS}}(\H_1,\H_2)$. The minimax lower rate is derived over a family of distributions $\mathcal{P}$ and all possible estimators $S_{\mathbf{z}}$. Below, we present the prior assumptions for $\mathcal{P}$, which describe the joint probability $\rho$ of the observation $(x,y)$:

\begin{itemize}
\item[1.] For $\rho_x-$almost all $x\in \H_1$, there holds 
$$\mathbb{E}(y|x)= S^{\dagger}x, \text{ for some } S^{\dagger} \in \B(\H_1,\H_2).$$
\item[2.] $\epsilon=y- S^{\dagger}x \in \H_2$ is independent of $x$ satisfying
$$\mathbb{E} \epsilon =\mathbf{0} \text{ and } \sigma^2=\mathbb{E} \|\epsilon\|^2_{\H_2}<\infty.$$
\item[3.] $S^{\dagger}$ satisfies Assumption \ref{a1} with $r>0$ and $\|J\|\leq R <\infty$.
\item[4.] The input satisfies Assumption \ref{a3} with constant $c$, and
\[
\operatorname{Tr}(L_C)=\mathbb E\|x\|_{H_1}^2\le1,
\qquad
\operatorname{Tr}(L_C^s)\le Q,
\]
where $0<s\le1$ and $Q>0$.
\end{itemize}

Let
\[
\Omega=
\{(\sigma,r,R,s,Q,c):
\sigma,r,R,Q>0,\ 0<s\le1,\ c\ge3\},
\]
and let $\mathcal P_\omega$ denote the distributions satisfying
Conditions 1--4 with the fixed parameter tuple $\omega$.
The quantities $L_C$, $S^\dagger$, and $\mathcal E$ are understood
to depend on $\rho$. The regularity condition selects the target
that vanishes on $\ker L_C$.
We first state the prediction lower bound under weak regularity.


\begin{theorem}\label{thm-minimax1}
Suppose that \(\H_1\) and \(\H_2\) are infinite-dimensional real separable Hilbert spaces.
Then 
\begin{equation*}	
\inf_{\omega\in\Omega}\liminf_{T\to \infty}\inf_{S_{\bf z}}\sup_{\rho \in\P_\omega}\mathbb{P}_{{\bf z}\sim \rho^{\otimes T}}\left(\mathcal{E}(S_{\bf z})-\mathcal{E}(S^{\dagger})\geq\gamma_\omega T^{-\frac{1+2r-s}{2r+1}}\right)>0,
\end{equation*} where the second infimum is taken over all estimators based on $\bf{z}$, i.e., measurable mappings $S_{\bf z}: \bf{z} \rightarrow \B(\H_1,\H_2)$, and $\gamma_\omega>0$ is independent of $T$ and depends on $\omega$.
\end{theorem}

In the theorem below, we establish the minimax lower bounds for prediction error and estimation error under the strong regularity condition of Assumption \ref{a4}. Correspondingly, we replace Condition 3 in the prior assumptions about the family of distributions $\mathcal{P}$ with:
\begin{itemize}
\item[$3^\prime$.] $S^{\dagger}$ satisfies Assumption \ref{a4} with
$\widetilde r>0$ and $\|\widetilde J\|_{\mathrm{HS}}\le R$.
\end{itemize} 
Let
\[
\widetilde\Omega=
\{(\sigma,\widetilde r,R,s,Q,c):
\sigma,\widetilde r,R,Q>0,\ 0<s\le1,\ c\ge3\}.
\]
For $\omega\in\widetilde\Omega$, let
$\widetilde{\mathcal P}_\omega$ denote the class obtained by
replacing Condition 3 with Condition $3'$.

	
\begin{theorem}\label{thm-minimax2}
Suppose that \(\H_1\) is infinite-dimensional and \(\H_2\) is a nontrivial real separable Hilbert space.
Then 
\begin{equation*}	
\inf_{\omega\in \widetilde \Omega}\liminf_{T\to \infty}\inf_{S_{\bf z}}\sup_{\rho \in\widetilde{\mathcal P}_\omega}\mathbb{P}_{{\bf z}\sim \rho^{\otimes T}}\left(\mathcal{E}(S_{\bf z})-\mathcal{E}(S^{\dagger})\geq \gamma_\omega' T^{-\frac{2\widetilde{r}+1}{1+2\widetilde{r}+s}}\right)>0
\end{equation*} 
and 
\begin{equation*}	
\inf_{\omega\in \widetilde \Omega}\liminf_{T\to \infty}\inf_{S_{\bf z}}\sup_{\rho \in\widetilde{\mathcal P}_\omega}\mathbb{P}_{{\bf z}\sim \rho^{\otimes T}}\left(\left\|S_{\bf z}-S^\dagger\right\|_{\mathrm{HS}}^2\geq\gamma_\omega'' T^{-\frac{2\widetilde{r}}{1+2\widetilde{r}+s}}\right)>0.
\end{equation*} 
For the prediction lower bound, the infimum is over measurable
estimators taking values in $\mathcal B(\H_1,\H_2)$; for the
estimation lower bound, it is over measurable estimators taking
values in $\mathcal B_{\mathrm{HS}}(\H_1,\H_2)$.
The constants $\gamma'_\omega,\gamma''_\omega>0$ depend only
on $\omega$ and are independent of $T$.
\end{theorem}



Theorem \ref{thm-minimax1} requires infinite-dimensional $\H_2$ because the
matrix packing uses an increasing number of output directions.
In contrast, Theorem \ref{thm-minimax2} is proved on a one-dimensional output
subspace. In fixed finite output dimension, the operator and
Hilbert--Schmidt norms are equivalent up to
dimension-dependent constants. 
Furthermore, 
since \(Z\ge0\), $\mathbb E Z\ge a\,\mathbb P(Z\ge a)$, 
the probability lower bounds immediately imply corresponding expectation lower bounds.


For each fixed $\omega$, Conditions 3 and $3'$ give
$\|J\|\le R$ and $\|\widetilde J\|_{\mathrm{HS}}\le R$,
respectively, while Condition 4 fixes the bounds $Q$ and $c$.
Together with $\|L_C\|\le1$, $\|S^\dagger\|\le R$, and
$\mathcal E(S^\dagger)=\sigma^2$, these bounds allow the
constants in Theorems \ref{t1}--\ref{t4} to be chosen independently
of $\rho$ in the corresponding class.
The step-size constants are fixed over that class;
for decaying steps, require \eqref{condition} and $0<\eta_1<1$.

Consequently, Theorem \ref{t4} is minimax rate-optimal for
Hilbert--Schmidt estimation over
$\widetilde{\mathcal P}_\omega$ for $0<s\le1$.
Theorem \ref{t3} is also rate-optimal when
$0<\widetilde r<(1-s)/2$, and is optimal up to a logarithmic
factor at $\widetilde r=(1-s)/2$.
For prediction, Theorems \ref{t2} and \ref{t6} match the respective
lower bounds up to a logarithmic factor when $s=1$.


\subsection{Nonlinear Targets and Best Linear Approximation}\label{subsection: inverse problem}

We now consider nonlinear targets, where the regression map 
\(x \mapsto \mathbb E[y|x]\) is not necessarily linear. 
In this setting, SGD no longer recovers the nonlinear target itself; rather, it converges to the best linear approximation \(S^\dagger\). 
Under the same weak or strong regularity conditions imposed on \(S^\dagger\), the prediction and Hilbert-Schmidt estimation errors achieve the same convergence rates as in the corresponding linear results.

Recall the general model of operator learning introduced in Subsection \ref{subsection: minimax}, in which we begin with a joint probability distribution $\rho$ of $(x,y)$ on $\mathcal{H}_1 \times \mathcal{H}_2$. Consider the scenario where $\mathbb{E}[y|x]: \mathcal{H}_1 \to \mathcal{H}_2$ is nonlinear with respect to $x\in\mathcal{H}_1$. In this case, we define 
$$ \mathbb{E}[y|x]=S^\dagger x+ \delta(x),$$
where 
$$S^\dagger \in \arg\min\left\{\mathcal{E}(S)=\be\left\|Sx-y\right\|^2_{\H_2}:S\in\B(\H_1,\H_2)\right\}$$
is a risk-minimizing bounded linear predictor, and
$\delta(x)=\mathbb{E}[y| x]-S^\dagger x$ is the corresponding
approximation residual. We assume that such a bounded minimizer
exists.
By  \cite{mollenhauer2022learning}, a bounded linear operator that minimizes $\mathcal{E}(S)=\be\left\|Sx-y\right\|^2_{\H_2}$ exists if and only if $\mathrm{Ran}\left(\be[x\otimes y]\right)\subseteq\mathrm{Ran}\left(L_C\right)$, with $\mathrm{Ran}(\cdot)$ indicating the range of the operator. 
Under this existence condition, $S^\dagger$ minimizes
$\mathcal{E}(S)$ over $S\in\B(\H_1,\H_2)$ if and only if it satisfies
\begin{equation} \label{non1}
    S^\dagger L_C=\be[y\otimes x].
\end{equation}
This result can be derived by calculating the Gâteaux derivative of $\mathcal{E}(S)$ at $S$, as detailed in \cite{mollenhauer2022learning}. Furthermore, $S^\dagger$ is unique if and only if $L_C$ is injective. For more conditions equivalent to the existence of $S^\dagger$ and a description of the set of $S^\dagger$ minimizing $\mathcal{E}(S)$, one can refer to \cite{mollenhauer2022learning}.

We analyze the SGD iteration \eqref{sgd} for estimating this best
linear approximation. Specifically, we consider the nonlinear model
\begin{equation} \label{nonlinear}
    y=S^\dagger x+ \delta(x) + \epsilon.
\end{equation}
Here $\epsilon$ is centered, independent of $x$, and satisfies
$\mathbb{E}\|\epsilon\|_{\H_2}^2=\sigma^2<\infty$.
We assume that the approximation residual has a finite fourth moment,
and write
\begin{equation*}
\mu^2:=\mathbb{E}\|\delta(x)\|_{\H_2}^2,
\qquad
\nu_\delta^2:=
\left(\mathbb{E}\|\delta(x)\|_{\H_2}^4\right)^{1/2}<\infty.
\end{equation*}
The normal equation \eqref{non1} implies
$\mathbb{E}[\delta(x)\otimes x]=0$.
Moreover, $\mathcal{E}(S^\dagger)=\mu^2+\sigma^2$ is the minimum
risk over bounded linear predictors. We continue to use
$L_C=\mathbb{E}[x\otimes x]$ and the SGD iteration \eqref{sgd}.

The weak or strong regularity condition is imposed on the selected
minimizer $S^\dagger$. Either regularity condition implies
$S^\dagger h=0$ for every $h\in\ker L_C$ and therefore selects a
unique representative among the bounded minimizers. The
Hilbert--Schmidt estimation error refers to this representative, and injectivity of $L_C$ is not required. The following theorem transfers
the corresponding linear convergence rates to model \eqref{nonlinear}, with
explicit restrictions on the step-size prefactors.

\begin{theorem} \label{Thm:nonlinear}
    Consider model \eqref{nonlinear}, where
$\mathbb{E}\|x\|_{\H_1}^2\le1$ and $\epsilon$ is centered,
independent of $x$, and satisfies
$\mathbb{E}\|\epsilon\|_{\H_2}^2=\sigma^2<\infty$.
Suppose that a bounded minimizer
\begin{equation*}
S^\dagger\in
\operatorname*{arg\,min}_{S\in\B(\H_1,\H_2)}\mathcal{E}(S)
\end{equation*}
exists, and that
$\delta(x)=\mathbb{E}[y|x]-S^\dagger x$ satisfies
$\nu_\delta^2=
(\mathbb{E}\|\delta(x)\|_{\H_2}^4)^{1/2}<\infty$. 
Let $z_1,\ldots,z_T$ be independent copies of $(x,y)$, with
$T\ge2$, and let $S_1=0$ and $S_{t+1}$ be generated by
\eqref{sgd} for $t=1,\ldots,T$.

Assume that Assumption~\ref{a3} holds with constant $c>0$.
For comparison with Theorems~\ref{t1} and~\ref{t2}, impose Assumption~\ref{a1}
on the selected $S^\dagger$; for comparison with
Theorems~\ref{t5}--\ref{t4}, impose Assumption~\ref{a4}.
In each case, impose Assumption~\ref{a2} with the same range of $s$
as in the corresponding theorem. The following conclusions hold.

\begin{enumerate}
\item[(1)]
For the decaying-step results corresponding to
Theorems~\ref{t1},~\ref{t5}, and~\ref{t3}, set $\eta_t=\eta_1t^{-\theta}$,
where $\theta$ is chosen as in the corresponding theorem.
Choose $\eta_1>0$ such that
\begin{equation*}
3c\left[
3c_5(\eta_1,1,\theta)
\max\left\{\frac{1}{e\theta},1\right\}
+\eta_1^2
\right]<1,
\qquad
\eta_1\|L_C\|<1.
\end{equation*}
Here $c_5$ is the constant in Proposition~\ref{eta1}.
Then the prediction bounds of Theorems~\ref{t1} and~\ref{t5} and the
Hilbert--Schmidt estimation bound of Theorem~\ref{t3} retain the
same powers of $T+1$ and the same logarithmic factors,
with possibly different multiplicative constants.

\item[(2)]
For the constant-step results, set
$\eta_t=\eta_*(T+1)^{-a}$ for $t=1,\ldots,T$, where
\begin{equation*}
a=
\begin{cases}
\dfrac{2r+1-s}{2r+2-s},
& \text{for the prediction bound of Theorem~\ref{t2}},\\[1ex]
\dfrac{2\widetilde r+1}{2\widetilde r+2},
& \text{for the prediction bound of Theorem~\ref{t6}},\\[1ex]
\dfrac{2\widetilde r+s}{1+2\widetilde r+s},
& \text{for the estimation bound of Theorem~\ref{t4}}.
\end{cases}
\end{equation*}
Choose $0<\eta_*\le ea/(1+42c)$.
Then the corresponding bounds of Theorems~\ref{t2},~\ref{t6}, and~\ref{t4}
retain the same powers of $T+1$ and the same logarithmic factors,
with possibly different multiplicative constants.
\end{enumerate}

All multiplicative constants are independent of $T$ and may
additionally depend on $\nu_\delta^2$.
\end{theorem}

The proof of this theorem is provided in Appendix \ref{Appendix:nonlinear}.

\section{Extensions and Applications}\label{section: related work}

The Hilbert-space formulation provides a common analytical foundation for several important learning models. In this section, we bring affine operator regression, vector-valued kernel regression, and scalar-valued functional learning into the same framework. The resulting representations identify the corresponding SGD iterations and prediction and estimation errors, allowing the convergence guarantees of Section \ref{section: main results} to transfer to these settings under the respective assumptions.



\subsection{An Extension of Model \eqref{linear}} \label{3.1}

We consider the following biased extension of model \eqref{linear}, defined as
\begin{equation}\label{linear2}
	y=S^\dagger x + y_0 + \epsilon \text{ for some } y_0 \in \mathcal{H}_{2} \text{ and }S^\dagger \in \mathcal{B}(\mathcal{H}_{1}, \mathcal{H}_{2}). 
 \end{equation}
This model introduces a bias term $y_0$. Additionally, we assume that the input is centered. Our goal is to extend the conclusions derived from model \eqref{linear} to this case.

Consider $\mathcal{H}_3:= \mathcal{H}_1 \times \mathbb{R}=\{(x,s): x \in \mathcal{H}_{1}, s \in \mathbb{R}\}$ with the inner product 
\[\langle(x_1,s_1),(x_2,s_2)\rangle_{\mathcal{H}_3} = \langle x_1,x_2\rangle_{\mathcal{H}_{1}} + s_1s_2.\]
It is easy to verify that $(\H_3,\langle\cdot,\cdot\rangle_{\H_3},\|\cdot\|_{\H_3})$ is a separable Hilbert space and its complete orthonormal basis is $\{(\mathbf{0},1), (e_k,0)_{k\geq 1}\}$. Let $W^\dagger: \mathcal{H}_3 \to \mathcal{H}_2$ be the mapping $(x,s) \mapsto S^{\dagger}x + sy_0$ induced by $S^\dagger$. Clearly, $W^\dagger$ is a bounded linear operator. Additionally, $W^\dagger$ is a Hilbert-Schmidt operator if and only if $S^\dagger$ is a Hilbert-Schmidt operator. Then model \eqref{linear2} can be expressed as
\begin{equation}\label{linear22}
y = W^\dagger(x,1) + \epsilon,
\end{equation} which is exactly the form of model \eqref{linear}. According to SGD algorithm \eqref{sgd}, one can prove by induction that, at the \(t\)-th iteration, for all $(x,s) \in \H_3$, $W_t$ has the form of $W_t(x,s) = S_t x + s \beta_t$, where $S_t \in \B(\H_1,\H_2)$ and $\beta_t \in \H_2$ given by
\begin{equation*} 
	\begin{cases}
		S_1 = \mathbf{0}, \beta_1 = \mathbf{0},\\
		S_{t+1} = S_{t} - \eta_{t}(S_tx_t + \beta_{t} - y_t) \otimes x_t,\\
		\beta_{t+1} = \beta_{t} - \eta_{t}(S_tx_t + \beta_{t} - y_t).
	\end{cases}
\end{equation*}

We next show that if the assumptions on $S^\dagger$ in Subsection \ref{subsection: assumption} hold for model \eqref{linear}, then the corresponding assumptions also hold for $W^\dagger$ in model \eqref{linear22} with samples $\{(x_t,1), y_t\}_{t\geq 1}$. To demonstrate this conclusion, we introduce some symbols. For any $T_1 \in \B(\H_1, \H_1)$ and $T_2 \in \B(\mathbb{R}, \mathbb{R})$, and for any $a \in \H_1, b \in \mathbb{R}$, define $T_1 \oplus T_2 \in \B(\H_3, \H_3)$ as $T_1 \oplus T_2(a, b) := (T_1a, T_2b)$. Let $\widetilde{L_C} = \be\left[(x, 1) \otimes (x, 1)\right]$. Then, one can prove that 
\[
\widetilde{L_C} = L_C \oplus I,
\]
where $I$ is the identity operator. Moreover, by performing a spectral decomposition on the compact operator $L_C$, it can further be demonstrated that
\[
\widetilde{L_C}^s = L_C^s \oplus I
\]
for any $s > 0$, and $\mathrm{Tr}\left(\widetilde{L_C}^s\right) = \mathrm{Tr}\left(L_C^s\right) + 1$. Therefore, if assumption \ref{a2} holds for $L_C$, the corresponding assumption is also valid for $\widetilde{L_C}$. Combining Proposition \ref{1}, it is easy to prove that if assumption \ref{a3} is true for $\{x_t\}_{t\geq 1}$ with $c > 0$, then the corresponding assumption for $\{(x_t, 1)\}_{t\geq 1}$ is also true with $\widetilde{c} = 8\max(c, 1)$. 
For Assumptions \ref{a1} and \ref{a4}, suppose that $S^\dagger=GL_C^q$
with $q>0$ and $G\in\B(\H_1,\H_2)$.
Define $G_{\mathrm{aug}}\in\B(\H_3,\H_2)$ by
$G_{\mathrm{aug}}(a,b)=Ga+by_0$.
Then
\begin{equation*}
W^\dagger=G_{\mathrm{aug}}\widetilde L_C^q,
\qquad
\|G_{\mathrm{aug}}\|^2
\le\|G\|^2+\|y_0\|_{\H_2}^2.
\end{equation*}
Moreover, if $G$ is Hilbert--Schmidt, then
\[
\|G_{\mathrm{aug}}\|_{\mathrm{HS}}^2
=\|G\|_{\mathrm{HS}}^2+\|y_0\|_{\H_2}^2.
\]
Taking $(q,G)=(r,J)$ or
$(q,G)=(\widetilde r,\widetilde J)$ transfers the weak or strong
regularity condition, respectively. The required normalization and
the resulting error bounds are summarized below.


Since $\mathbb{E}\|(x,1)\|_{\H_3}^2\le2$, apply Theorems~\ref{t1}--\ref{t4} to the normalized
input $(x,1)/\sqrt2$ and target $\sqrt2\,W^\dagger$.
This corresponds to using $\eta_t=\bar\eta_t/2$ in the iteration
above, where $\bar\eta_t$ is chosen according to that theorem for
the normalized model, with moment constant
$\widetilde c=8\max(c,1)$.
Under the respective regularity assumptions, the prediction and
Hilbert--Schmidt estimation bounds for $W_{T+1}$ retain the
corresponding convergence rates, up to multiplicative constants.


Therefore, after the above rescaling, the convergence results in Section \ref{section: main results} apply to the biased model \eqref{linear2}, yielding the corresponding prediction and Hilbert--Schmidt estimation bounds for \(W_{T+1}\).

\subsection{Application to Learning with Vector-valued RKHS}	\label{3.2}
	
In this subsection, we apply the established theoretical results to the framework of learning with vector-valued RKHS, thereby obtaining the corresponding convergence analysis. Vector-valued functions are critical to learning problems in various fields, including computational biology, physics, and economics \cite{lutkepohl2013vector,brouard2016fast,cai2021deepm}, which require simultaneous predictions of multiple variables. In addition to the case of infinite dimensional output \cite{kadri2016operator}, surrogate approaches in structured output prediction are primarily motivated for vector-valued regression \cite{brouard2016input,ciliberto2020general}, which can be utilized in graph prediction \cite{brouard2016fast}, image completion \cite{weston2002kernel} and label ranking \cite{korba2018structured}. To address the vector-valued regression problems, an important and widely used theoretical framework is learning with vector-valued RKHS \cite{brogat2022vector}. The essential properties and related approximation theories of vector-valued RKHS have been studied extensively in the literature \cite{carmeli2006vector,micchelli2005learning,carmeli2010vector}.

Next, we present the model of vector-valued regression and establish its connection with the operator learning model \eqref{linear}. Then, we will provide an explicit form of the corresponding SGD algorithm. We denote $\H_{2}$ as a separable Hilbert space and $\mathcal{X}$ as a Polish space. The vector-valued regression model associates the explanatory variable $x \in \mathcal{X}$ with the response variable $y \in \H_2$ via 
\begin{equation}\label{model2}
	y=h^\dagger(x)+\epsilon,
\end{equation}
where $h^\dagger: \mathcal{X} \rightarrow \H_2$ is a measurable function and $\epsilon \in \H_2$ represents centered random noise independent of $x$, with finite variance. We assume that $h^\dagger$ resides in a vector-valued RKHS $\H_K$, equipped with an operator-valued positive kernel $K: \mathcal{X} \times \mathcal{X} \rightarrow \B(\H_2,\H_2)$. Here, $K$ is called positive if
\begin{enumerate}
	\item $K(x,x') = [K(x',x)]^*$ for any $(x,x')\in \mathcal{X} \times \mathcal{X}$, where $[K(x',x)]^* \in \B(\H_2,\H_2)$ denotes the adjoint operator of $K(x',x)$;
	\item For any $n\in \mathbb{N}$ and any $\{(x_i, y_i)\}_{i=1}^n \in \left(\mathcal{X} \times \H_2\right)^n$,  there holds $\sum_{i,j=1}^{n} \langle K(x_i,x_j)y_i,y_j \rangle_{\H_{2}} \geq 0$.
\end{enumerate}  Specifically, when $\H_2 = \mathbb{R}$, 
$K$ reduces to a scalar-valued positive kernel, which we denote by $\mathcal{K}$ in the subsequent discussions. By the definition of vector-valued RKHS, $\H_K$ is the completion of the linear span of $\{K(x, \cdot)y : (x, y) \in \mathcal{X} \times \H_2\}$ according to the norm induced by the inner product satisfying $\langle K(x, \cdot)y, K(x', \cdot)y' \rangle_{K} = \langle K(x, x')y, y' \rangle_{\H_{2}}$. In our study, we consider a specific form of operator-valued kernel described in the following assumption, which is also utilized in previous works \cite{ciliberto2016consistent,ciliberto2020general,brogat2022vector}.
\begin{assumption}\label{a5}
The vector-valued RKHS $\H_K$ is generated by the operator-valued kernel $K(x,x')=\mathcal{K}(x,x')I$, where $I$ is the identity operator on $\H_2$ and $\mathcal{K}:\mathcal{X}\times\mathcal{X}\rightarrow\mathbb{R}$ is a scalar-valued positive Mercer kernel satisfying $\|\mathcal{K}\|_{\infty}\leq\kappa^2$ for some $0<\kappa<\infty$.
\end{assumption}

Under Assumption \ref{a5}, consider the RKHS $\H_1$, which is generated by the scalar-valued positive kernel $\mathcal{K}$ on $\mathcal{X} \times \mathcal{X}$. 
The continuity of $\mathcal K$ and the separability of $\mathcal X$
imply that $\H_1$ is separable.
Then, we have the following fundamental result, which asserts that the vector-valued RKHS $\H_K$ is isometrically isomorphic to the space of Hilbert-Schmidt operators $\B_{\mathrm{HS}}(\H_1,\H_2)$.

\begin{proposition} \label{2} 
Under Assumption \ref{a5}, the vector-valued RKHS $\H_K$, associated with the operator-valued kernel $K(x,x^\prime)=\mathcal{K}(x,x^\prime)I$, is isometrically isomorphic to the space of Hilbert-Schmidt operators $\B_{\mathrm{HS}}(\H_1,\H_2)$, where $\H_1$ is the RKHS $\H_{\mathcal{K}}$ generated by $\mathcal{K}$. Moreover, for each $h \in \H_K$, there is a unique operator $S_h \in \B_{\mathrm{HS}}(\H_1,\H_2)$ such that 
\begin{equation*}
h(x)=S_h\phi(x), \ \forall x\in\mathcal{X},
\end{equation*}
and $\|h\|_{\H_K}=\|S_h\|_{\mathrm{HS}}$, where $\phi:\mathcal{X}\to \H_1,\phi(x)=\mathcal{K}(x,\cdot)$ is the feature map.
\end{proposition}

We leave the proof in the Appendix \ref{Appendix B}.
Under Assumption \ref{a5}, suppose that the target function
$h^\dagger$ in model \eqref{model2} belongs to $\H_K$. Then
according to Proposition \ref{2}, 
$$h^\dagger=S^\dagger\phi \text{ for some }  S^\dagger\in \B_{\mathrm{HS}}(\H_1,\H_2).$$ Now, we consider the vector-valued regression model \eqref{model2} in a random design setting. Therefore, we assume that $\{(x_t, y_t)\}_{t\in \mathbb{N}_T}$ are $T$ i.i.d. copies of $(x,y)$, where $x$ is a random variable taking values in $\mathcal{X}$ and $y$ is related to $x$ through \eqref{model2}. To predict the output $y$ from the input $x$ and estimate $h^\dagger$, it is essential to estimate the Hilbert-Schmidt operator $S^\dagger$ using $\{(\phi(x_t), y_t)\}_{t\in \mathbb{N}_T}$. By applying the SGD algorithm, as delineated in Equation (\ref{sgd}), we can formulate the iteration process:
\begin{equation*}
		\begin{cases}
			S_1=\mathbf{0}, \\
			S_{t+1}=S_t-\eta_t(S_t\phi(x_t)-y_t)\otimes \phi(x_t), \\
			h_{t+1}=S_{t+1}\phi.
		\end{cases}
\end{equation*}

Recall that, for any $x\in \mathcal{X}$, $\phi(x)=\mathcal{K}(x,\cdot)$, and $\|\mathcal{K}\|_{\infty}\leq \kappa^2$. Then $\mathbb{E}\left[\mathcal{K}(x,x)\right]\leq\kappa^2$ which implies $L_C=\mathbb{E}[\phi(x)\otimes\phi(x)]$ is of trace class. Under Assumptions \ref{a4}, \ref{a2}, \ref{a3}, and \ref{a5}, we obtain convergence rates of $\mathbb{E}_{z^T}[\mathcal{E}(S_{T+1})-\mathcal{E}(S^{\dagger})]$ and $\be_{z^{T}}\left[\left\|S_{T+1}-S^{\dagger}\right\|_{\mathrm{HS}}^2\right]$, where $\mathbb{E}_{z^T}[\mathcal{E}(S_{T+1})-\mathcal{E}(S^{\dagger})]=\mathbb{E}_{z^T}[\mathcal{E}(h_{T+1})-\mathcal{E}(h^{\dagger})]$ and $\be_{z^{T}}\left[\left\|S_{T+1}-S^{\dagger}\right\|_{\mathrm{HS}}^2\right]=\be_{z^{T}}\left[\left\|h_{T+1}-h^{\dagger}\right\|_{\H_K}^2\right]$ by Proposition \ref{2}. We thus establish convergence upper bounds on prediction and estimation errors of $h_{T+1}$.

\subsection{Application to Learning with Scalar-valued RKHS}	\label{3.3}

In this subsection, we apply our theoretical analysis to the learning framework within scalar-valued RKHS. In fact, the problem discussed in this section can be regarded as a specific case of learning with vector-valued RKHS. Under Assumption \ref{a5}, we set $\H_2=\mathbb{R}$ and define the kernel function $K(x,x')=\mathcal{K}(x,x')$, where $\mathcal{K}: \mathcal{X} \times \mathcal{X} \to \mathbb{R}$ is a positive kernel. Consequently, the vector-valued regression model \eqref{model2} reduces to nonlinear regression learning in the scalar-valued RKHS $\H_{\mathcal{K}}$. Leveraging the reproducing property of $\H_{\mathcal{K}}$, model \eqref{model2} can be reformulated as
\begin{equation}\label{model3}
y=h^\dagger(x)+\epsilon=\langle h^\dagger,\mathcal{K}(x,\cdot)\rangle_{\H_{\mathcal{K}}}+\epsilon.
\end{equation}
The SGD algorithm for solving \eqref{model3} is given by
\begin{equation} \label{sgd3}
\begin{cases}
h_1=0, \\
h_{t+1}=h_t-\eta_t(h_t(x_t)-y_t)\mathcal{K}(x_t,\cdot).
\end{cases}
\end{equation}

Systematic research on model (\ref{model3}) and algorithm (\ref{sgd3}) began with seminal works \cite{smale2006online} and \cite{ying2008online}. Since then, there has been extensive research on various variants of algorithm (\ref{sgd3}), see, e.g., \cite{smale2006online,MR3519927,MR3714260,pillaud2018statistical,guo2023optimality}. The study in \cite{ying2008online} derived the prediction and estimation errors of algorithm \eqref{sgd3} under the condition that Assumption \ref{a2} holds with $s=1$. More recently, the research in \cite{guo2022rates} delved deeply into algorithm (\ref{sgd3}) as a special case of randomized Kaczmarz algorithms in Hilbert spaces (strictly speaking, it refers to the normalized version of algorithm (\ref{sgd3})). By incorporating Assumptions \ref{a2} and \ref{a4}, the authors in \cite{guo2022rates} improved the analysis of \cite{ying2008online}. 
Our analysis additionally imposes Assumption \ref{a3} and yields the
prediction and estimation bounds in Theorems \ref{t5}, \ref{t6}, and \ref{t4}.


Some studies, such as \cite{guo2022capacity,chen2022online}, explore functional SGD in RKHSs. The functional linear model is represented as
\begin{equation} \label{model4}
y=\int_{\T}\beta^\dagger(u)x(u)du+\epsilon,
\end{equation}
where $\T$ is a compact set in a Euclidean space, $x\in\mathcal{L}^2(\T)$ denotes a square-integrable random input function, and $\beta^\dagger\in\mathcal{L}^2(\T)$ is an unknown slope function.

    
    Let $\H_\mathcal{K}$ be the RKHS generated by a continuous positive kernel
$\mathcal{K}$ on $\mathcal T\times\mathcal T$, and assume
$\mathbb E\|x\|_{\mathcal{L}^2(\T)}^2<\infty$.
Since $\mathcal T$ is compact, the canonical map
$\tau:\H_\mathcal{K}\to \mathcal{L}^2(\T)$ is bounded, with adjoint
\[
\tau^*g=\int_{\mathcal T}\mathcal{K}(u,\cdot)g(u)\,du.
\]
Let $C_x:=\mathbb E[x\otimes x]$ denote the uncentered
covariance operator on $\mathcal{L}^2(\T)$.
The corresponding operator for the transformed input $\tau^*x$ is
\[
L_C:=\mathbb E[(\tau^*x)\otimes(\tau^*x)]
=\tau^*C_x\tau:\H_\mathcal{K}\to \H_\mathcal{K}.
\]
    If we further assume that $\beta^\dagger$ belongs to $\H_{\mathcal{K}}$, then we can rewrite \eqref{model4} as
	\[
	y=\left\langle \tau\beta^\dagger,x\right\rangle_{\mathcal{L}^2(\T)}+\epsilon=\left\langle\beta^\dagger,\tau^*x\right\rangle_{{\H_{\mathcal{K}}}}+\epsilon.
	\]
	By letting the target operator $S^\dagger(\cdot)=\langle\beta^\dagger,\cdot\rangle_{\H_{\mathcal{K}}}$ in model (\ref{linear})  and implementing the SGD  algorithm (\ref{sgd}) on a sequence of samples $\{(\tau^*x_t,y_t)\}_{t\geq1}$, the iterative process is given by
	\begin{equation*}
		\begin{cases}
			\beta_1=\mathbf{0}, \\
			\beta_{t+1}=\beta_t-\eta_t\left(\int_\T\beta_{t}(u)x_t(u)du-y_t\right)\int_\T {\cal K}(v,\cdot)x_t(v)dv, \\
			S_{t}(\cdot)=\langle\beta_t,\cdot\rangle_{\H_{\cal K}},
		\end{cases}
	\end{equation*}
    which agrees with the iteration in \cite{guo2022capacity,chen2022online}.
In this setting, Assumption \ref{a4} is equivalent to
\[
\beta^\dagger=L_C^{\widetilde r}g^\dagger
=(\tau^*C_x\tau)^{\widetilde r}g^\dagger,
\qquad g^\dagger\in \H_{\cal K},\quad \widetilde r>0.
\]
With Assumption \ref{a2} imposed on $L_C$ and Assumption \ref{a3} on
$\tau^*x$, Theorems \ref{t5}--\ref{t4} yield the corresponding prediction
and $\H_{\cal K}$-norm estimation rates, using the input rescaling
in Subsection \ref{subsection: upper rates}. Here,
$\|S_t-S^\dagger\|_{\mathrm{HS}}
=\|\beta_t-\beta^\dagger\|_{\H_{\cal K}}$.


\section{Error Decomposition and Basic Estimates}	\label{section: Error decomposition}

This section will present the error decomposition and basic estimates utilized in the convergence analysis of upper bounds. To this end, we first establish some useful observations for later use. By the definition of the prediction error, for any $S \in \mathcal{B}(\H_{1},\H_{2})$,
\begin{equation*}
		\begin{split}
			\mathcal{E}(S)-\mathcal{E}(S^{\dagger})
			&= \mathbb{E}\left[\left\|y-Sx\right\|_{\H_{2}}^{2}\right] - \mathbb{E}\left[\|y-S^{\dagger}x\|_{\H_{2}}^{2}\right] \\
			&= \mathbb{E}\left[\|(S^{\dagger}-S)x+\epsilon\|_{\H_{2}}^{2}\right] - \mathbb{E}\|\epsilon\|_{\H_{2}}^{2} \\
			&= \mathbb{E}\left[\|(S^{\dagger}-S)x\|_{\H_{2}}^{2}\right] + 2\mathbb{E}\left[\langle \epsilon, (S^{\dagger}-S)x\rangle_{\H_{2}}\right]. 
		\end{split}
	\end{equation*}
Given that $\epsilon$ is centered and independent of $x$, we deduce that $\mathbb{E}\left[\langle \epsilon, (S^{\dagger}-S)x\rangle_{\H_{2}}\right]  = 0$. It then follows that
	\begin{equation}\label{form1}
		\mathcal{E}(S)-\mathcal{E}(S^{\dagger}) = \mathbb{E}\left[\|(S^{\dagger}-S)x\|_{\H_{2}}^{2}\right].
	\end{equation}
Moreover, let $\{f_k\}_{k\geq1}$ be an orthonormal basis of $\H_2$. The expression $(S^{\dagger}-S)x$ can be expanded using Fourier series, leading to
	\begin{equation}\label{form2}
		\begin{aligned}
			\mathcal{E}(S)-\mathcal{E}(S^{\dagger}) &= \mathbb{E}\left[\sum_{k \geq1}\langle(S^\dagger-S)x, f_k\rangle_{\H_{2}}^2\right] \\
			&= \sum_{k \geq1}\mathbb{E}\left[\left\langle(S^\dagger-S)x \otimes x(S^\dagger-S)^* f_k, f_k\right\rangle_{\H_{2}}\right] \\
			&= \left\|\left(S-S^\dagger\right)L_C^{\frac{1}{2}}\right\|_{\mathrm{HS}}^{2}.
		\end{aligned}	
	\end{equation}
Thus, we have derived two equivalent formulations (\ref{form1}) and (\ref{form2}) for the prediction error.

We can demonstrate that the iterative process in algorithm \eqref{sgd} satisfies the following recursive relationship through simple calculations. The proof of this lemma is provided in Appendix \ref{Lemma 4.1}.
\begin{lemma} \label{lemma4.1}
	Let $\{S_t\}_{t \geq 1}$ be defined as in (\ref{sgd}). Then
	\begin{equation}
		S_{t+1}-S^\dagger=(S_t-S^\dagger)(I-\eta_t L_C)+\eta_{t}\B_t, \label{lemma 1.1}
	\end{equation}
	where $I$ denotes the identity operator, and $\B_t$ is given by
	\begin{equation}
		\B_t=(S_t-S^\dagger)L_C+(y_t-S_tx_t)\otimes x_t. \label{lemma 1.2}
	\end{equation}
	Furthermore, it holds that $\mathbb{E}_{z_t}[\B_t]=0, \forall t \geq 1$.
\end{lemma}

Set $\prod_{t+1}^{t}(I-\eta_{t}L_C)=I$. By applying induction to equality (\ref{lemma 1.1}) for any $T\in\bn$, we derive an important decomposition as follows,
\begin{equation}\label{form}
	\begin{split}
		S_{T+1}-S^{\dagger}=&(S_{T}-S^{\dagger})(I-\eta_{T}L_C)+\eta_T\B_{T} 
		\\=&(S_{T-1}-S^{\dagger})(I-\eta_{T-1}L_C)(I-\eta_{T}L_C) 
		\\&+\eta_{T-1}\B_{T-1}(I-\eta_{T}L_C)+\eta_T\B_{T} 
		\\=&-S^{\dagger}\prod_{t=1}^{T}(I-\eta_{t}L_C)+\sum_{t=1}^{T}\eta_t\B_{t}\prod_{j=t+1}^{T}(I-\eta_{j}L_C). 
	\end{split}
\end{equation}

The decomposition in \eqref{form} is referred to as the martingale decomposition, as referenced in the studies by \cite{tarres2014online}. Only the second term in (\ref{form}) exhibits randomness.

Recall that the estimation error is given by $\be_{z^{T}}\left[\left\|(S_{T+1}-S^{\dagger})\right\|_{\mathrm{HS}}^{2}\right]$ and the prediction error can be reformulated as $\be_{z^{T}}\left[\left\|(S_{T+1}-S^{\dagger})L_C^{1/2}\right\|_{\mathrm{HS}}^{2}\right]$. In the following proposition, we consider the decomposition of $\be_{z^{T}}\left[\left\|(S_{T+1}-S^{\dagger})L_C^{\alpha}\right\|_{\mathrm{HS}}^{2}\right]$ for $\alpha \geq 0$. Hereinafter, we use the convention $\mathbb{E}_{z^0} \xi = \xi$  for any random variable $\xi$.

\begin{proposition}\label{Proposition error 1}
Let $T\ge1$ and let $S_1,\ldots,S_{T+1}$ be defined by \eqref{sgd}.
Suppose that Assumption~\ref{a3} holds with constant $c>0$.
Then, for every $\alpha\ge0$ satisfying
$S^{\dagger}L_C^\alpha\in\B_{\mathrm{HS}}(\H_1,\H_2)$,
there holds
\begin{equation} \label{eq2}
\be_{z^{T}}\left[\left\|(S_{T+1}-S^{\dagger})L_C^{\alpha}\right\|_{\mathrm{HS}}^{2}\right]\leq
\mathcal{T}_1 + \mathcal{T}_2
\end{equation} where
\begin{equation} \label{errors}
\begin{split} 
&\mathcal{T}_1:=\left\|S^{\dagger}L_C^{\alpha}\prod_{t=1}^{T}(I-\eta_{t}L_C)\right\|_{\mathrm{HS}}^{2},\\
&\mathcal{T}_2:=\sum_{t=1}^T \eta_t^2 \left(c\be_{z^{t-1}}\left[\mathcal{E}(S_{t})-\mathcal{E}(S^{\dagger})\right]+\sqrt{c}\sigma^2\right)\mathrm{Tr}\left(L_C^{1+2\alpha}\prod_{j=t+1}^{T}(I-\eta_{j}L_C)^2\right).\\ 
\end{split}      
\end{equation}
\end{proposition}

The proof of this proposition is deferred to Appendix \ref{Prop 4.2}, in which Assumption \ref{a3} plays an important role. The error decomposition derived herein is critical in streamlining our objective in error analysis, which involves estimating $\mathcal{T}_1$ and $\mathcal{T}_2$ in \eqref{errors}. We will now establish some basic estimates that are crucial for establishing tight bounds on these two terms. The first term $\mathcal{T}_1$ is referred to as the approximation error, while the second term $\mathcal{T}_2$ is known as the cumulative sample error. We bound $\mathcal{T}_1$ and $\mathcal{T}_2$ separately using the estimates developed in the following propositions. The proofs of these propositions are provided in Appendix.

Recall that the weak regularity condition stated in Assumption \ref{a1} asserts that $S^{\dagger}=\J L_C^r$ for some $\J \in \B(\H_1,\H_2)$ and $r>0$. Regarding the first term, specifically $\left\|S^{\dagger}L_C^{\alpha}\prod_{t=1}^{T}(I-\eta_{t}L_C)\right\|_{\mathrm{HS}}^{2}$ in (\ref{eq2}), we utilize the inequality $\|PQ\|_{\mathrm{HS}}\leq\|P\|\|Q\|_{\mathrm{HS}}$ applicable to any Hilbert-Schmidt operator $Q$ and bounded operator $P$. This leads to
\[
\left\|S^{\dagger}L_C^{\alpha}\prod_{t=1}^{T}(I-\eta_{t}L_C)\right\|_{\mathrm{HS}}^{2}
\leq\|\J\|^2
\left\|L_C^{r+\alpha}\prod_{t=1}^{T}(I-\eta_{t}L_C)\right\|_{\mathrm{HS}}^{2}.
\]
Suppose the strong regularity condition in Assumption \ref{a4} is satisfied for some $\widetilde{\J}\in \B_{\mathrm{HS}}(\H_1,\H_2)$ and a positive parameter $\widetilde{r}>0$, ensuring that $S^{\dagger}=\widetilde{\J} L_C^{\widetilde{r}}$. Employing the inequality $\|PQ\|_{\mathrm{HS}}\leq\|P\|_{\mathrm{HS}}\|Q\|$, it follows that
\[
\left\|S^{\dagger}L_C^{\alpha}\prod_{t=1}^{T}(I-  \eta_{t}L_C)\right\|_{\mathrm{HS}}^{2}
\leq\|\widetilde{\J}\|_{\mathrm{HS}}^2
\left\|L_C^{\widetilde{r}+\alpha}\prod_{t=1}^{T}(I-\eta_{t}L_C)\right\|^{2}.
\]
Therefore, we first establish bounds for $\left\|A^\beta\prod_{j=l}^{T}(I-\eta_jA)^2\right\|$ when $A$ is a compact positive operator, and $\beta>0$, in the following lemma. 
	
\begin{lemma} \label{lemma basic}
		Let $A$ be a compact positive operator on a Hilbert space, 
        let $\beta>0$, and let $l,T\in\mathbb N$ satisfy $1\le l\le T$. Given that $\eta_t\|A\|<1$ for any $l\leq t\leq T$, then
		\begin{align}
			\left\|A^\beta\prod_{j=l}^{T}(I-\eta_jA)^2\right\|\leq\left(\frac{\beta}{2e}\right)^\beta\left(\sum_{j=l}^{T}\eta_j\right)^{-\beta}, \label{l1}
		\end{align}
		and
		\begin{align}
			\left\|A^\beta\prod_{j=l}^{T}(I-\eta_jA)^2\right\|\leq2\frac{(\frac{\beta}{2e})^\beta+\|A\|^\beta}{1+(\sum_{j=l}^{T}\eta_j)^\beta}. \label{l2}
		\end{align}
        Bound \eqref{l2} also holds for $l=T+1$, with the empty product
equal to $I$ and the empty sum equal to zero.
	\end{lemma}

 This lemma facilitates the estimation of $\left\|L_C^{\alpha}\prod_{t=1}^{T}(I-\eta_{t}L_C)\right\|_{\mathrm{HS}}^{2}$, as demonstrated in the following proposition.

\begin{proposition} \label{first term}
        Let $\left\{\eta_{t}=\eta_{1}t^{-\theta}\right\}_{t\geq 1}$ with $0\leq\theta<1$ and $\eta_t\|L_C\|<1$. Under Assumption \ref{a2} with $0<s\leq1$, then for any $T\geq1$ and  $\alpha>s/2$, there holds
        \begin{equation}
            \left\|L_C^{\alpha}\prod_{t=1}^{T}(I-\eta_{t}L_C)\right\|^2_{\mathrm{HS}}\leq \mathrm{Tr}(L_C^s)\left(\frac{(2\alpha-s)(1-\theta)}{2e(1-2^{\theta-1})\eta_{1}}\right)^{2\alpha-s}T^{-(1-\theta)(2\alpha-s)}.
        \end{equation}
    \end{proposition}

This proposition is used to bound the approximation error under Assumptions \ref{a1} and \ref{a2}, with $0<s\leq1$, when the step-size is set as $\eta_{t}=\eta_{1}t^{-\theta}$, where $0\leq\theta<1$. An important result is presented next.

\begin{proposition} \label{eta1}
		Let $v>0$, $T\geq2$, and $\left\{\eta_t=\eta_{1}t^{-\theta}\right\}_{t\in \mathbb{N}_T}$ with $\eta_{1}>0$ and $0<\theta<1$.
		\begin{enumerate}
			\item[(1)] When $0<v<1$, there holds
			\[
			\sum_{t=1}^{T-1}\frac{\eta_{t}^2}{1+\left(\sum_{j=t+1}^{T}\eta_{j}\right)^v}
			\leq c_5\begin{cases}
				(T+1)^{1-v-\theta(2-v)}, & \text{ if } 0<\theta<\frac{1}{2}, \\
				(T+1)^{-v/2}\log (T+1), & \text{ if }\theta=\frac{1}{2}, \\
				(T+1)^{-v(1-\theta)}, &\text{ if } \frac{1}{2}<\theta<1.
			\end{cases}
			\]
			\item[(2)] When $v>1$, there holds
			\[
			\sum_{t=1}^{T-1}\frac{\eta_{t}^2}{1+\left(\sum_{j=t+1}^{T}\eta_{j}\right)^v}
			\leq c_5(T+1)^{-\min\{\theta,v(1-\theta)\}}.
			\]
			\item[(3)] When $v=1$, there holds
			\[
			\sum_{t=1}^{T-1}\frac{\eta_{t}^2}{1+\left(\sum_{j=t+1}^{T}\eta_{j}\right)^v}
			\leq c_5\begin{cases}
				(T+1)^{-\theta}\log (T+1) , & \text{ if } 0<\theta\leq\frac{1}{2}, \\
				(T+1)^{-(1-\theta)} , & \text{ if } \frac{1}{2}<\theta<1.
			\end{cases}
			\]
		\end{enumerate}
The constant $c_5=c_5(\eta_{1},v,\theta)$ is independent of $T$ and will be given in the proof. 	
	\end{proposition}

Furthermore, under the conditions of the above proposition, letting $v=1$ implies that
	\begin{equation}
		\sum_{t=1}^{T-1}\frac{\eta_{t}^2}{1+\sum_{j=t+1}^{T}\eta_{j}}
		\leq c_5(\eta_{1},1,\theta)\max\left\{\frac{1}{e\theta},1\right\}, \label{temp6}
	\end{equation}
	which can be easily verified when we notice that \[(T+1)^{-\theta}\log(T+1)\leq\max_{x>0}x^{-\theta}\log x=1/(e\theta).\]

Consider the term $\sqrt{c}\be_{z^{t-1}}\left[\mathbb{E}(S_{t})-\mathcal{E}(S^{\dagger})\right]+\sigma^2$ in the cumulative sample error $\mathcal{T}_2$, where  $c$ is a constant defined in Assumption \ref{a3}. 
To control this term, it suffices to establish a uniform upper bound for \(\mathbb E_{z^{t-1}}[\mathcal E(S_t)]\), which is provided in Propositions \ref{p1} and \ref{p2}.


\begin{proposition} \label{p1}
    Let $\left\{\eta_{t}=\eta_{1}t^{-\theta}\right\}_{t \geq 1}$ with $0<\theta<1$. Suppose that Assumption \ref{a3} holds with $c>0$. If $\eta_1$ satisfies
    \begin{equation} \label{condition 1}
        c\left[3c_5(\eta_{1},1,\theta)\max\left\{\frac{1}{e\theta},1\right\}+\eta_1^2\right]<1
    \end{equation}
    and $\eta_{1}\|L_C\|<1$,
    then, for any $t\geq 1$, there holds
    \begin{equation}
        \mathbb{E}_{z^{t-1}}\left[\mathcal{E}(S_{t})\right]\leq M, \label{bound1}
    \end{equation}
    where $M$ is a constant independent of $t$ and will be given in the proof.
\end{proposition}

We note that condition (\ref{condition 1}) is satisfied by choosing a sufficiently small $\eta_1>0$. In Propositions \ref{eta1} and \ref{p1}, the assumption $0<\theta<1$ aligns with the setting employing decaying step sizes $\eta_{t}=\eta_{1}t^{-\theta}$ for $t\geq 1$. In the next two propositions, we focus on the fixed step sizes, i.e., $\theta=0$ and $\eta_{t}\equiv \eta_{1}$ for $t\in\bn_T$.

\begin{proposition} \label{eta2}
   Let $v>0$, $\eta_1>0$, and $\eta_t=\eta_1$ for
$t=1,\ldots,T$, where $T\ge2$. Then
\begin{equation*}
\sum_{t=1}^{T-1}
\frac{\eta_t^2}
{1+\left(\sum_{j=t+1}^{T}\eta_j\right)^v}
\le c_6
\begin{cases}
\eta_1^{2-v}(T+1)^{1-v}, & 0<v<1,\\
\eta_1\log\bigl(1+\eta_1(T+1)\bigr), & v=1,\\
\eta_1, & v>1,
\end{cases}
\end{equation*}
where $c_6=c_6(v)$ is given by
\begin{equation*}
c_6(v):=
\begin{cases}
\dfrac{1}{1-v}, & 0<v<1,\\
1, & v=1,\\
\dfrac{v}{v-1}, & v>1.
\end{cases}
\end{equation*}
	\end{proposition}
	
Next, we demonstrate that $\mathbb{E}_{z^{t-1}}\left[\mathcal{E}(S_{t})\right]$ remains bounded when $\theta=0$.
	
 \begin{proposition} \label{p2}
		Let $\left\{\eta_t=\eta_1\right\}_{t \in \mathbb{N}_T}$ with $T\geq 2$. Suppose that Assumption \ref{a3} is satisfied with $c>0$ and $\eta_{1}$ satisfies the condition
		\begin{equation}
			\eta_{1}\leq\frac{1}{(1+14c)\log(T+1)}. \label{condition2}
		\end{equation}
		Then, for every $1\le t\le T+1$,
		\begin{equation}
			\be_{z^{t-1}}\left[\mathcal{E}(S_{t})\right]\leq \widetilde{M}, \label{bound2}
		\end{equation}
		where $\widetilde{M}$ is a constant independent of $\eta_1$ or $T$, and will be given in the proof.
	\end{proposition}

\section{Convergence Analysis of Upper Bounds under Weak Regularity Condition}\label{proof1}

In this section, we present the proofs of  Theorem \ref{t1} and Theorem \ref{t2}, which focus on the convergence analysis of the prediction error based on  weak regularity condition in Assumption \ref{a1}. For this purpose, we first establish the following estimate according to the results of Section \ref{section: Error decomposition}.
	
	\begin{proposition} \label{further1}
		Let $\{S_t\}_{t \in \mathbb{N}_{T+1}}$ be defined by (\ref{sgd}),  $\eta_{t}=\eta_{1}t^{-\theta}$ with $0\leq\theta<1$ and $\eta_t\|L_C\|<1$. If Assumption \ref{a1}, Assumption \ref{a2} with $0<s\leq1$ and Assumption \ref{a3} hold, then 
		\begin{equation} \label{eqq1}
			\begin{aligned}
				\be_{z^{T}}&\left[\mathcal{E}(S_{T+1})-\mathcal{E}(S^{\dagger})\right]\leq c_7\eta_{1}^{-(2r+1-s)}(T+1)^{-(2r+1-s)(1-\theta)}
				\\&+c_8\sum_{t=1}^{T}\left(\sqrt{c}\be_{z^{t-1}}\left[\mathcal{E}(S_{t})-\mathcal{E}(S^{\dagger})\right]+\sigma^2\right)\frac{\eta_{t}^{2}}{1+\left(\sum_{j=t+1}^{T}\eta_j\right)^{2-s}},
			\end{aligned}
		\end{equation}
		where $c_7=c_7(\theta,r,s)$ and $c_8=c_8(s)$ are constants independent of $\eta_1$ or $t$, and will be given in the proof.
	\end{proposition}
	
	\begin{proof}
		By the error decomposition (\ref{eq2}) in Proposition \ref{Proposition error 1} with $\alpha=\frac{1}{2}$, there holds
		\begin{equation} 
			\begin{aligned}  \label{temp25}
				\be_{z^{T}}&\left[\mathcal{E}(S_{T+1})-\mathcal{E}(S^{\dagger})\right]
				\leq\left\|S^{\dagger}L_C^{\frac{1}{2}}\prod_{t=1}^{T}(I-\eta_{t}L_C)\right\|_{\mathrm{HS}}^{2}
				\\&+\sum_{t=1}^{T}\eta_{t}^{2}\sqrt{c}\left(\sqrt{c}\be_{z^{t-1}}\left[\mathcal{E}(S_{t})-\mathcal{E}(S^{\dagger})\right]+\sigma^2\right)\mathrm{Tr}\left(L_C^2\prod_{j=t+1}^{T}(I-\eta_{j}L_C)^2\right).
			\end{aligned}  		
		\end{equation}
		We estimate the first term on the right-hand side of the above inequality (\ref{temp25}). By recalling Assumption \ref{a1}, we see that
		\begin{equation}
			\begin{aligned}
				\left\|S^{\dagger}L_C^{\frac{1}{2}}\prod_{t=1}^{T}(I-\eta_{t}L_C)\right\|_{\mathrm{HS}}^{2}&=\left\|\J L_C^{r+\frac{1}{2}}\prod_{t=1}^{T}(I-\eta_{t}L_C)\right\|_{\mathrm{HS}}^{2}
				\\&\leq\left\|\J\right\|^2\left\|L_C^{r+\frac{1}{2}}\prod_{t=1}^{T}(I-\eta_tL_C)\right\|_{\mathrm{HS}}^2. \label{temp26}
			\end{aligned}
		\end{equation}
		Applying Proposition \ref{first term} with $0<s\leq1$ and $\alpha=r+1/2$ to (\ref{temp26}) yields that
		\begin{align*}
			&\left\|S^{\dagger}L_C^{\frac{1}{2}}\prod_{t=1}^{T}(I-\eta_{t}L_C)\right\|_{\mathrm{HS}}^{2}
			\\\leq&\left\|\J\right\|^2\mathrm{Tr}(L_C^s)\left(\frac{(2r+1-s)(1-\theta)}{2e\eta_{1} (1-2^{\theta-1})}\right)^{2r+1-s}T^{-(2r+1-s)(1-\theta)}\\
			\leq& 2^{(2r+1-s)(1-\theta)}\left\|\J\right\|^2\mathrm{Tr}(L_C^s)\left(\frac{(2r+1-s)(1-\theta)}{2e\eta_{1} (1-2^{\theta-1})}\right)^{2r+1-s}(T+1)^{-(2r+1-s)(1-\theta)},
		\end{align*}
		where the fact $T^{-(2r+1-s)(1-\theta)}\leq2^{(2r+1-s)(1-\theta)}(T+1)^{-(2r+1-s)(1-\theta)}$ is used in the last inequality. Set $c_7=2^{(2r+1-s)(1-\theta)}\left\|\J\right\|^2\mathrm{Tr}(L_C^s)\left(\frac{(2r+1-s)(1-\theta)}{2e (1-2^{\theta-1})}\right)^{2r+1-s}$, then
		\begin{equation}
			\left\|S^{\dagger}L_C^{\frac{1}{2}}\prod_{t=1}^{T}(I-\eta_{t}L_C)\right\|_{\mathrm{HS}}^{2}
			\leq c_7\eta_{1}^{-(2r+1-s)}(T+1)^{-(2r+1-s)(1-\theta)}. \label{temp27}
		\end{equation}
		For the second term of (\ref{temp25}), there holds
		\begin{align*}
			\mathrm{Tr}\left(L_C^2\prod_{j=t+1}^{T}(I-\eta_{j}L_C)^2\right)
			&\leq\mathrm{Tr}(L_C^s)\left\|L_C^{2-s}\prod_{j=t+1}^{T}(I-\eta_{j}L_C)^2\right\| \\&\leq\mathrm{Tr}(L_C^s)\frac{2\left(\frac{2-s}{2e}\right)^{2-s}+2\|L_C\|^{2-s}}{1+\left(\sum_{j=t+1}^{T}\eta_j\right)^{2-s}},
		\end{align*}
		where the last step follows from Lemma \ref{lemma basic} with $\beta=2-s$ and $l=t+1$. By setting $c_8=\left(2\left(\frac{2-s}{2e}\right)^{2-s}+2\|L_C\|^{2-s}\right)\sqrt{c}\mathrm{Tr}(L_C^s)$, we obtain that the second term is bounded by		
		\begin{gather}
			c_8\sum_{t=1}^{T}\left(\sqrt{c}\be_{z^{t-1}}\left[\mathcal{E}(S_{t})-\mathcal{E}(S^{\dagger})\right]+\sigma^2\right)\frac{\eta_{t}^{2}}{1+\left(\sum_{j=t+1}^{T}\eta_j\right)^{2-s}}.  \label{temp28}
		\end{gather}
		The proof is completed by combining (\ref{temp27}), (\ref{temp28}) with (\ref{temp25}).	
	\end{proof}

Now we are in the position to prove Theorem \ref{t1}.
	
	\begin{proof}[Proof of Theorem \ref{t1}.]
		From Proposition \ref{p1}, we obtain that
		\[
		\be_{z^{t-1}}\left[\mathcal{E}(S_{t})\right]\leq M.
		\] 
		By applying the error decomposition (\ref{eqq1}) in Proposition \ref{further1}, we have
		\begin{equation} \label{temp29}
			\begin{aligned}
				\be_{z^{T}}&\left[\mathcal{E}(S_{T+1})-\mathcal{E}(S^{\dagger})\right]\leq c_7\eta_{1}^{-(2r+1-s)}(T+1)^{-(2r+1-s)(1-\theta)}
				\\&+c_8\left(\sqrt{c}\left(M-\mathcal{E}(S^{\dagger})\right)+\sigma^2\right)
				\left(\sum_{t=1}^{T-1}\frac{\eta_{t}^{2}}{1+\left(\sum_{j=t+1}^{T}\eta_j\right)^{2-s}}+\eta_{1}^22^{2\theta}(T+1)^{-2\theta}\right),
			\end{aligned}
		\end{equation}
		where in the last inequality (\ref{temp29}) 
  is due to the fact
		\[
		\frac{\eta_{T}^{2}}{1+\left(\sum_{j=T+1}^{T}\eta_j\right)^{2-s}}=\eta_{T}^2=\eta_{1}^2T^{-2\theta}\leq\eta_{1}^22^{2\theta}(T+1)^{-2\theta}.
		\] 
		We split the remainder of our estimates into two cases.
		
  \textbf{Case 1:} Suppose that Assumption \ref{a2} holds with $s=1$.
		By Proposition \ref{eta1} with the case of $v=1$, there holds
		\begin{equation} \label{temp32}
			\sum_{t=1}^{T-1}\frac{\eta_{t}^2}{1+\sum_{j=t+1}^{T}\eta_{j}}\leq c_5\begin{cases}
				(T+1)^{-\theta}\log (T+1) , & \text{ if } 0<\theta\leq\frac{1}{2}, \\
				(T+1)^{-(1-\theta)} , & \text{ if }\frac{1}{2}<\theta<1.
			\end{cases}
		\end{equation}
		Define $f_1(\theta)=-2r(1-\theta)$, $g_1(\theta)=-\min\{\theta,1-\theta\}$, and $h_1(\theta)=-2\theta$. Set the optimal choice of the parameter $\theta$ as
		\begin{equation} \label{temp33}
			\theta=\mathop{\arg\min}_{\theta}\max\{f_1(\theta),g_1(\theta),h_1(\theta)\}=
			\begin{cases}
				\frac{2r}{2r+1}, &\text{ if }r<\frac12,\\
				\frac{1}{2}, &\text{ if }r\geq\frac12. 		
			\end{cases}
		\end{equation}
		Hence, 
  by (\ref{temp32}), (\ref{temp33}) and the (\ref{temp29}) with $s=1$, we have
		\[
		\be_{z^{T}}\left[\mathcal{E}(S_{T+1})-\mathcal{E}(S^{\dagger})\right]
		\leq c_r
		(T+1)^{-\min\{\frac{2r}{2r+1},\frac{1}{2}\}}\log(T+1),
		\]
		where $c_r:=c_7\eta_{1}^{-2r}+c_8\left(\sqrt{c}\left(M-\mathcal{E}(S^{\dagger})\right)+\sigma^2\right)\left(c_5+\eta_{1}^22^{2\theta}\right).$
		
  \textbf{Case 2:} Suppose that Assumption \ref{a2} holds with $0<s<1$. By Proposition \ref{eta1} with the case of $v=2-s$, there holds
		\begin{equation} \label{temp34}
			\sum_{t=1}^{T-1}\frac{\eta_{t}^2}{1+\left(\sum_{j=t+1}^{T}\eta_{j}\right)^{2-s}}\leq c_5(T+1)^{-\min\{\theta,(2-s)(1-\theta)\}}.
		\end{equation}
		Define $f_2(\theta)=-(2r+1-s)(1-\theta)$,  $g_2(\theta)=-\min\{\theta,(2-s)(1-\theta)\}$ and $h_2(\theta)=-2\theta$. Choose 
		\begin{equation} \label{temp35}
			\theta=\mathop{\arg\min}_{\theta}\max\{f_2(\theta),g_2(\theta),h_2(\theta)\}=
			\begin{cases}
				\frac{2r+1-s}{2r+2-s}, &\text{ if } r<\frac{1}{2}, \\
				\frac{2-s}{3-s}, &\text{ if } r\geq\frac{1}{2}.
			\end{cases}
		\end{equation}
		Consequently,  by combining (\ref{temp34}), (\ref{temp35}) with (\ref{temp29}), we obtain
		\[
		\be_{z^{T}}\left[\mathcal{E}(S_{T+1})-\mathcal{E}(S^{\dagger})\right]
		\leq c_{r,s}
		\begin{cases}
			(T+1)^{-\frac{2r+1-s}{2r+2-s}}, &\text{ if }r<\frac{1}{2}, \\
			(T+1)^{-\frac{2-s}{3-s}}, &\text{ if }r\geq\frac12,
		\end{cases}
		\]
		where
		$
		c_{r,s}:=c_7\eta_{1}^{-(2r+1-s)}+c_8\left(\sqrt{c}\left(M-\mathcal{E}(S^{\dagger})\right)+\sigma^2\right)\left(c_5+\eta_{1}^22^{2\theta}\right)$. 
  
Thus, we complete the proof.
\end{proof}
	
Next, we prove Theorem \ref{t2}, where we bound the terms $\be_{z^{t-1}}\left[\mathcal{E}(S_{t})\right]$ and $\sum_{t=1}^{T-1}\frac{\eta_{t}^2}{1+\left(\sum_{j=t+1}^{T}\eta_{j}\right)^v}$ by applying Proposition \ref{p2} and Proposition \ref{eta2} instead of Proposition \ref{p1} and Proposition \ref{eta1}.
	
 \begin{proof}[Proof of Theorem \ref{t2}.]
		From Proposition \ref{p2}, we see that if the condition $\eta_{1}\leq\frac{1}{(1+14c)\log(T+1)}$ is satisfied, it follows that $\be_{z^{t-1}}\left[\mathcal{E}(S_{t})\right]\leq\widetilde{M}$ for any $t\in\bn_T$.
		By applying the error decomposition (\ref{eqq1}) with $\theta=0$ in Proposition \ref{further1}, we have
		\begin{equation} \label{temp36}
			\begin{aligned}
				\be_{z^{T}}&\left[\mathcal{E}(S_{T+1})-\mathcal{E}(S^{\dagger})\right]\leq c_7\eta_{1}^{-(2r+1-s)}(T+1)^{-(2r+1-s)}
				\\&+c_8\left(\sqrt{c}\left(\widetilde{M}-\mathcal{E}(S^{\dagger})\right)+\sigma^2\right)
				\left(\sum_{t=1}^{T-1}\frac{\eta_{t}^{2}}{1+\left(\sum_{j=t+1}^{T}\eta_j\right)^{2-s}}+\eta_{1}^2\right).
			\end{aligned}
		\end{equation}
		
  \textbf{Case 1:} If Assumption \ref{a2} holds with $s=1$. Then, 
		by Proposition \ref{eta2} with the case of $v=1$, there holds
		\begin{equation} \label{temp37}
        \sum_{t=1}^{T-1}
\frac{\eta_t^2}{1+\sum_{j=t+1}^{T}\eta_j}
\le
c_6\eta_1\log\bigl(1+\eta_1(T+1)\bigr).
		\end{equation}
		Choosing $\eta_{1}=\eta_*(T+1)^{-\frac{2r}{2r+1}}$ with $\eta_*\leq\frac{2er}{(1+14c)(2r+1)}$, then
		$\eta_{1}\leq\frac{1}{(1+14c)\log(T+1)}$. Substituting $\eta_1$ into (\ref{temp37}) yields 
		\begin{equation} \label{temp17}
			\begin{aligned}
				\begin{aligned}
c_6\eta_1\log\bigl(1+\eta_1(T+1)\bigr)
&=
c_6\eta_*(T+1)^{-\frac{2r}{2r+1}}
\log\left(1+\eta_*(T+1)^{\frac{1}{2r+1}}\right)\\
&\le
c_6\eta_*(T+1)^{-\frac{2r}{2r+1}}
\left[
\log(1+\eta_*)+\frac{\log(T+1)}{2r+1}
\right]\\
&\le
c_6\eta_*
\left[
\frac{1}{2r+1}+\log(1+\eta_*)
\right]
(T+1)^{-\frac{2r}{2r+1}}\log(T+1).
\end{aligned}
			\end{aligned}
		\end{equation}
		Putting the estimates (\ref{temp37}) and (\ref{temp17}) into (\ref{temp36}) with $s=1$ entails that
		\begin{equation*}
			\be_{z^{T}}\left[\mathcal{E}(S_{T+1})-\mathcal{E}(S^{\dagger})\right]
			\leq \widetilde{c}_{r,1}(T+1)^{-\frac{2r}{2r+1}}\log(T+1),
		\end{equation*}
        where
\begin{equation*}
\begin{aligned}
\widetilde{c}_{r,1}:={}&
c_7\eta_*^{-2r}
+c_8\left(
\sqrt c\left(\widetilde M-\mathcal E(S^\dagger)\right)
+\sigma^2
\right)\\
&\quad\times
\left[
c_6\eta_*
\left(\frac{1}{2r+1}+\log(1+\eta_*)\right)
+\eta_*^2
\right].
\end{aligned}
\end{equation*}
  
\textbf{Case 2:} Suppose that Assumption \ref{a2} holds with $0<s<1$. Applying Proposition \ref{eta2} with $v=2-s$ tells us that
		\begin{equation} \label{temp38}
			\sum_{t=1}^{T-1}\frac{\eta_{t}^2}{1+\left(\sum_{j=t+1}^{T}\eta_{j}\right)^{2-s}}
			\leq c_6\eta_{1}.
		\end{equation} 
		Choosing $\eta_{1}=\eta_{*}(T+1)^{-\frac{2r+1-s}{2r+2-s}}$ with $\eta_{*}\leq\frac{e(2r+1-s)}{(1+14c)(2r+2-s)}$, then $\eta_{1}\leq\frac{1}{(1+14c)\log(T+1)}$. We can obtain from (\ref{temp36}) and (\ref{temp38}) that
		\begin{align*}
			\be_{z^{T}}\left[\mathcal{E}(S_{T+1})-\mathcal{E}(S^{\dagger})\right]
			\leq\widetilde{c}_{r,s}(T+1)^{-\frac{2r+1-s}{2r+2-s}},
		\end{align*}
		where $\widetilde{c}_{r,s}:=c_7\eta_{*}^{-(2r+1-s)}+c_8\left(\sqrt{c}\left(\widetilde{M}-\mathcal{E}(S^{\dagger})\right)+\sigma^2\right)\left(c_6\eta_{*}+\eta_{*}^2\right)$. 
		
  Then we finish the proof. \end{proof}

\section{Convergence Analysis of Upper Bounds under Strong Regularity Condition} \label{proof2}

In this section, we present the proofs of  Theorem \ref{t5}, Theorem \ref{t6}, Theorem \ref{t3}, and Theorem \ref{t4}, which focus on the convergence analysis of the prediction and estimation errors based on strong regularity condition in Assumption \ref{a4}. To this end, we establish the following estimate according to the results of Section \ref{section: Error decomposition}.
	
	\begin{proposition}  \label{eqqq}
    Let $T\ge1$ and let $S_1,\ldots,S_{T+1}$ be generated by
\eqref{sgd}, with $\eta_t=\eta_1t^{-\theta}$,
$\eta_1>0$, $0\le\theta<1$, and $\eta_t\|L_C\|<1$
for $t=1,\ldots,T$.
Suppose that Assumptions~\ref{a4},~\ref{a2} (with $0<s\le1$), and~\ref{a3} hold.
For $\beta\ge0$ and $u\ge0$, define
\begin{equation*}
\Psi_\beta(u):=
\begin{cases}
(1+u^\beta)^{-1}, & \beta>0,\\
1, & \beta=0.
\end{cases}
\end{equation*}
Then, for every $0\le\alpha\le1/2$,
\begin{equation} \label{eqq4}
    \begin{aligned}
\mathbb E_{z^T}\!\left[
\left\|(S_{T+1}-S^\dagger)L_C^\alpha\right\|_{\mathrm{HS}}^2
\right]
&\le
c_9\eta_1^{-2(\widetilde r+\alpha)}
(T+1)^{-2(\widetilde r+\alpha)(1-\theta)}\\
&{}+
c_{10}\sum_{t=1}^T\eta_t^2
\left(
\sqrt c\,\mathbb E_{z^{t-1}}
[\mathcal E(S_t)-\mathcal E(S^\dagger)]
+\sigma^2
\right)
\Psi_{1+2\alpha-s}\!\left(\sum_{j=t+1}^T\eta_j\right),
\end{aligned}
\end{equation}
		where $c_9=c_9(\theta,\widetilde{r},\alpha)$ and $c_{10}=c_{10}(s,\alpha)$ are constants independent of $\eta_{1}$ or $t$, and will be given in the proof.
	\end{proposition}
	
	\begin{proof}
		By the error decomposition (\ref{eq2}) in Proposition \ref{Proposition error 1}, there holds
		\begin{equation} \label{temp39}
			\begin{aligned} 
				&\be_{z^{T}}\left[\left\|\left(S_{T+1}-S^{\dagger}\right)L_C^\alpha\right\|_{\mathrm{HS}}^2\right]
				\leq\left\|S^{\dagger}L_C^\alpha\prod_{t=1}^{T}(I-\eta_{t}L_C)\right\|_{\mathrm{HS}}^{2}
				\\&+\sum_{t=1}^{T}\eta_{t}^{2}\sqrt{c}\left(\sqrt{c}\be_{z^{t-1}}\left[\mathcal{E}(S_{t})-\mathcal{E}(S^{\dagger})\right]+\sigma^2\right)\mathrm{Tr}\left(L_C^{1+2\alpha}\prod_{j=t+1}^{T}(I-\eta_{j}L_C)^2\right). 
			\end{aligned}  		
		\end{equation}
		Since Assumption \ref{a4} holds, we can bound the first term on the right-hand side of (\ref{temp39}) as 
		\begin{equation*} 
			\begin{aligned}
				\left\|S^{\dagger}L_C^\alpha\prod_{t=1}^{T}(I-\eta_{t}L_C)\right\|_{\mathrm{HS}}^{2}
				&=\left\|\widetilde{\J}L_C^{\widetilde{r}+\alpha}\prod_{t=1}^{T}(I-\eta_tL_C)\right\|_{\mathrm{HS}}^2
				\\&\leq\left\|\widetilde{\J}\right\|_{\mathrm{HS}}^2\left\|L_C^{\widetilde{r}+\alpha}\prod_{t=1}^{T}(I-\eta_tL_C)\right\|^2 \\
				&\leq \left\|\widetilde{\J}\right\|_{\mathrm{HS}}^2\left(\frac{\widetilde{r}+\alpha}{e}\right)^{2(\widetilde{r}+\alpha)}\left(\sum_{t=1}^{T}\eta_t\right)^{-2(\widetilde{r}+\alpha)},
			\end{aligned}
		\end{equation*}
		where we have used Lemma \ref{lemma basic} with $\beta=2(\widetilde{r}+\alpha)$ and $l=1$ in the last inequality. 
		For the term $\sum_{t=1}^{T}\eta_t$, if $T\geq2$, it follows that
		\begin{align}  \label{temp41}
			\sum_{t=1}^{T}\eta_t
			=\eta_{1}\sum_{t=1}^{T}t^{-\theta}
			\geq\eta_{1}\int_{T/2}^{T}u^{-\theta}\mathrm{d}u=\frac{\eta_{1}(1-2^{\theta-1})}{1-\theta}T^{1-\theta}.
		\end{align}
		One can easily verify that (\ref{temp41}) is also true for $T=1$. Consequently, we get
  \begin{equation}\label{temp40}
  		\begin{split}
			\left\|S^{\dagger}L_C^\alpha\prod_{t=1}^{T}(I-\eta_{t}L_C)\right\|_{\mathrm{HS}}^{2}	&\leq\left\|\widetilde{\J}\right\|_{\mathrm{HS}}^2\left(\frac{(\widetilde{r}+\alpha)(1-\theta)
			}{e\eta_{1}(1-2^{\theta-1})}\right)^{2(\widetilde{r}+\alpha)}T^{-2(\widetilde{r}+\alpha)(1-\theta)}\\&\leq c_9\eta_{1}^{-2(\widetilde{r}+\alpha)}(T+1)^{-2(\widetilde{r}+\alpha)(1-\theta)},  
		\end{split}
  \end{equation}
		where $c_9(\theta,\widetilde{r},\alpha):=2^{2(\widetilde{r}+\alpha)(1-\theta)}\left\|\widetilde{\J}\right\|_{\mathrm{HS}}^2\left(\frac{(\widetilde{r}+\alpha)(1-\theta)
		}{e(1-2^{\theta-1})}\right)^{2(\widetilde{r}+\alpha)}.$

        To bound the second term of \eqref{temp39}, set $\beta=1+2\alpha-s$.
First suppose that $\beta>0$. For $t<T$, Lemma~4.3 gives
\begin{equation*}
\begin{aligned}
\operatorname{Tr}\!\left(
L_C^{1+2\alpha}
\prod_{j=t+1}^T(I-\eta_jL_C)^2
\right)
&\le
\operatorname{Tr}(L_C^s)
\left\|
L_C^\beta\prod_{j=t+1}^T(I-\eta_jL_C)^2
\right\|\\
&\le
2\operatorname{Tr}(L_C^s)
\left[
\left(\frac{\beta}{2e}\right)^\beta+\|L_C\|^\beta
\right]
\Psi_\beta\!\left(\sum_{j=t+1}^T\eta_j\right).
\end{aligned}
\end{equation*}
For $t=T$, the same bound follows from
$\operatorname{Tr}(L_C^{s+\beta})
\le\operatorname{Tr}(L_C^s)\|L_C\|^\beta$
and $\Psi_\beta(0)=1$.

If $\beta=0$, then necessarily $\alpha=0$ and $s=1$.
Since $0\preceq I-\eta_jL_C\preceq I$ and all these factors
commute with $L_C$, we have, for every $1\le t\le T$,
\begin{equation*}
\operatorname{Tr}\!\left(
L_C\prod_{j=t+1}^T(I-\eta_jL_C)^2
\right)
\le \operatorname{Tr}(L_C).
\end{equation*}
Here the product is understood as the identity when $t=T$.
Define
\begin{equation*}
c_{10}(s,\alpha):=
\begin{cases}
2\sqrt c\,\operatorname{Tr}(L_C^s)
\left[
\left(\dfrac{\beta}{2e}\right)^\beta+\|L_C\|^\beta
\right],
& \beta=1+2\alpha-s>0,\\[1ex]
\sqrt c\,\operatorname{Tr}(L_C),
& \alpha=0,\ s=1.
\end{cases}
\end{equation*}
The second term of \eqref{temp39} is therefore bounded by
\begin{equation} \label{temp42}
c_{10}\sum_{t=1}^T\eta_t^2
\left(
\sqrt c\,\mathbb E_{z^{t-1}}
[\mathcal E(S_t)-\mathcal E(S^\dagger)]
+\sigma^2
\right)
\Psi_{1+2\alpha-s}\!\left(\sum_{j=t+1}^T\eta_j\right).
\end{equation}
Combining \eqref{temp40}, \eqref{temp42}, and \eqref{temp39} completes the proof.
\end{proof}

Now we give the proof of Theorem \ref{t5}.
	
	\begin{proof}[Proof of Theorem \ref{t5}.]
		From Proposition \ref{p1}, we have $\be_{z^{t-1}}\left[\mathcal{E}(S_{t})\right]\leq M$. By the decomposition (\ref{eqq4}) with $\alpha=1/2$ in Proposition \ref{eqqq}, it follows that
		\begin{equation}  \label{temp47}
			\begin{aligned}
				\be_{z^{T}}&\left[\mathcal{E}(S_{T+1})-\mathcal{E}(S^{\dagger})\right]
				\leq c_9\eta_{1}^{-(2\widetilde{r}+1)}(T+1)^{-(2\widetilde{r}+1)(1-\theta)}
				\\&+c_{10}\left(\sqrt{c}\left(M-\mathcal{E}(S^{\dagger})\right)+\sigma^2\right)\left(\sum_{t=1}^{T-1}\frac{\eta_{t}^{2}}{1+\left(\sum_{j=t+1}^{T}\eta_j\right)^{2-s}}+2^{2\theta}\eta_{1}^2(T+1)^{-2\theta}\right),
			\end{aligned}
		\end{equation}
		where the last inequality is due to the fact that
$\eta_{T}^2=\eta_{1}^2T^{-2\theta}\leq2^{2\theta}\eta_{1}^2(T+1)^{-2\theta}$.

\textbf{Case 1:} Suppose that Assumption \ref{a2} holds with $s=1$. Through Proposition \ref{eta1} with the case of $v=1$, $\be_{z^{T}}\left[\mathcal{E}(S_{T+1})-\mathcal{E}(S^{\dagger})\right]$ is bounded by
\begin{equation*}  
			\begin{aligned}
				&c_9\eta_{1}^{-(2\widetilde{r}+1)}(T+1)^{-(2\widetilde{r}+1)(1-\theta)}
				+c_{10}\left(\sqrt{c}\left(M-\mathcal{E}(S^{\dagger})\right)+\sigma^2\right)
				\\&\times \left[2^{2\theta}\eta_{1}^2(T+1)^{-2\theta}+c_5\begin{cases}
					(T+1)^{-\theta}\log (T+1) , & \text{if }0<\theta\leq\frac{1}{2} \\
					(T+1)^{-(1-\theta)} , & \text{if }\frac{1}{2}<\theta<1
				\end{cases}\right],
			\end{aligned}
		\end{equation*}
		Choose $\theta=1/2$, then 
		\begin{equation*}
			\be_{z^{T}}\left[\mathcal{E}(S_{T+1})-\mathcal{E}(S^{\dagger})\right]
			\leq c_{\widetilde{r},s}'(T+1)^{-\frac{1}{2}}\log(T+1),
		\end{equation*}
		where $c_{\widetilde{r},s}':=c_9\eta_{1}^{-(2\widetilde{r}+1)}+c_{10}\left(\sqrt{c}\left(M-\mathcal{E}(S^{\dagger})\right)+\sigma^2\right)\left(c_5+2^{2\theta}\eta_{1}^2\right).$

\textbf{Case 2:} Suppose that Assumption \ref{a2} holds with $0<s<1$. Through Proposition \ref{eta1} with the case of $v=2-s>1$, $\be_{z^{T}}\left[\mathcal{E}(S_{T+1})-\mathcal{E}(S^{\dagger})\right]$ is bounded by
		\begin{equation*}  
			\begin{aligned}
				&c_9\eta_{1}^{-(2\widetilde{r}+1)}(T+1)^{-(2\widetilde{r}+1)(1-\theta)}
				+c_{10}\left(\sqrt{c}\left(M-\mathcal{E}(S^{\dagger})\right)+\sigma^2\right)
				\\&\times
				\left[c_5(T+1)^{-\min\{\theta,(2-s)(1-\theta)\}}+2^{2\theta}\eta_{1}^2(T+1)^{-2\theta}\right].
			\end{aligned}
		\end{equation*}
		If $1-2\widetilde{r}<s<1$, choose $\theta=\frac{2-s}{3-s}$, else if $0<s\leq1-2\widetilde{r}$, choose $\theta=\frac{2\widetilde{r}+1}{2\widetilde{r}+2}$. Then there holds
		\begin{equation*}
			\be_{z^{T}}\left[\mathcal{E}(S_{T+1})-\mathcal{E}(S^{\dagger})\right]
			\leq c_{\widetilde{r},s}'(T+1)^{-\theta},
		\end{equation*}
		where $c_{\widetilde{r},s}':=c_9\eta_{1}^{-(2\widetilde{r}+1)}+c_{10}\left(\sqrt{c}\left(M-\mathcal{E}(S^{\dagger})\right)+\sigma^2\right)\left(c_5+2^{2\theta}\eta_{1}^2\right).$ 

  Thus, we finish the proof.
	\end{proof}

 Next, we prove Theorem \ref{t6}.
	
	\begin{proof}[Proof of Theorem \ref{t6}.]
		From Proposition \ref{p2}, if the condition $\eta_{1}\leq\frac{1}{(1+14c)\log(T+1)}$ is satisfied, we have $\be_{z^{t-1}}\left[\mathcal{E}(S_{t})\right]\leq\widetilde{M}$ for any $t\in\bn_T$.
		Applying the error decomposition (\ref{eqq4}) with $\alpha=1/2$ and $\theta=0$ in Proposition \ref{eqqq} yields that
		\begin{equation} \label{temp48}
			\begin{aligned}
				\be_{z^{T}}&\left[\mathcal{E}(S_{T+1})-\mathcal{E}(S^{\dagger})\right]
				\leq c_9\eta_{1}^{-(2\widetilde{r}+1)}(T+1)^{-(2\widetilde{r}+1)}
				\\&+c_{10}\left(\sqrt{c}\left(\widetilde{M}-\mathcal{E}(S^{\dagger})\right)+\sigma^2\right)\left(\sum_{t=1}^{T-1}\frac{\eta_{t}^{2}}{1+\left(\sum_{j=t+1}^{T}\eta_j\right)^{2-s}}+\eta_{1}^2\right).
			\end{aligned}
		\end{equation} 
  

\textbf{Case 1:}
Suppose that Assumption~\ref{a2} holds with $s=1$.
By Proposition~\ref{eta2} with $v=1$,
\begin{equation*}
\begin{aligned}
\mathbb E_{z^T}
\left[\mathcal E(S_{T+1})-\mathcal E(S^\dagger)\right]
&\le
c_9\eta_1^{-(2\widetilde r+1)}
(T+1)^{-(2\widetilde r+1)}\\
&\quad+
c_{10}\left(
\sqrt c\left(\widetilde M-\mathcal E(S^\dagger)\right)
+\sigma^2
\right)\\
&\qquad\times
\left[
c_6\eta_1\log\bigl(1+\eta_1(T+1)\bigr)+\eta_1^2
\right].
\end{aligned}
\end{equation*}
Choose
$\eta_1=\eta_*(T+1)^{-(2\widetilde r+1)/(2\widetilde r+2)}$
with
$\eta_*\le
e(2\widetilde r+1)/[(1+14c)(2\widetilde r+2)]$,
so that
$\eta_1\le[(1+14c)\log(T+1)]^{-1}$.
For this choice,
\begin{equation*}
\begin{aligned}
\eta_1\log\bigl(1+\eta_1(T+1)\bigr)
&\le
\eta_*(T+1)^{-\frac{2\widetilde r+1}{2\widetilde r+2}}
\left[
\log(1+\eta_*)
+\frac{\log(T+1)}{2\widetilde r+2}
\right]\\
&\le
\eta_*
\left[
\frac{1}{2\widetilde r+2}+\log(1+\eta_*)
\right]
(T+1)^{-\frac{2\widetilde r+1}{2\widetilde r+2}}
\log(T+1).
\end{aligned}
\end{equation*}
Consequently,
\begin{equation*}
\mathbb E_{z^T}
\left[\mathcal E(S_{T+1})-\mathcal E(S^\dagger)\right]
\le
c''_{\widetilde r,1}
(T+1)^{-\frac{2\widetilde r+1}{2\widetilde r+2}}
\log(T+1),
\end{equation*}
where
\begin{equation*}
\begin{aligned}
c''_{\widetilde r,1}:={}&
c_9\eta_*^{-(2\widetilde r+1)}
+c_{10}\left(
\sqrt c\left(\widetilde M-\mathcal E(S^\dagger)\right)
+\sigma^2
\right)\\
&\quad\times
\left[
c_6\eta_*
\left(\frac{1}{2\widetilde r+2}+\log(1+\eta_*)\right)
+\eta_*^2
\right].
\end{aligned}
\end{equation*}

\textbf{Case 2:} Suppose that Assumption \ref{a2} holds with $0<s<1$. Proposition \ref{eta2} with the case of $v=2-s>1$ tells us that
		\begin{equation} \label{temp49}
			\sum_{t=1}^{T-1}\frac{\eta_{t}^2}{1+\left(\sum_{j=t+1}^{T}\eta_{j}\right)^{2-s}}
			\leq c_6\eta_{1}.
		\end{equation}
		Substituting (\ref{temp49}) into (\ref{temp48}), we obtain 
		\begin{align*}
			\be_{z^{T}}\left[\mathcal{E}(S_{T+1})-\mathcal{E}(S^{\dagger})\right]
			\leq&
			c_9\eta_{1}^{-(2\widetilde{r}+1)}(T+1)^{-(2\widetilde{r}+1)}
			\\&+c_{10}\left(\sqrt{c}\left(\widetilde{M}-\mathcal{E}(S^{\dagger})\right)+\sigma^2\right)\left(c_6\eta_{1}+\eta_{1}^2\right).
		\end{align*}
		By choosing $\eta_{1}=\eta_{*}(T+1)^{-\frac{2\widetilde{r}+1}{2\widetilde{r}+2}}$ with $\eta_{*}\leq\frac{e(2\widetilde{r}+1)}{(1+14c)(2\widetilde{r}+2)}$ such that $\eta_{1}\leq\frac{1}{(1+14c)\log(T+1)}$, it follows that
		\begin{align*}
			\be_{z^{T}}\left[\mathcal{E}(S_{T+1})-\mathcal{E}(S^{\dagger})\right]
			\leq c_{\widetilde{r},s}''(T+1)^{-\frac{2\widetilde{r}+1}{2\widetilde{r}+2}},
		\end{align*}
		where $c_{\widetilde{r},s}'':=c_9\eta_{*}^{-(2\widetilde{r}+1)}+c_{10}\left(\sqrt{c}\left(\widetilde{M}-\mathcal{E}(S^{\dagger})\right)+\sigma^2\right)\left(c_6\eta_{*}+\eta_{*}^2\right)$.
  
		The proof is thus completed.		
	\end{proof}

 At the end of this section, we prove Theorem \ref{t3} and Theorem \ref{t4}.

\begin{proof}[Proof of Theorem \ref{t3}.]
		From Proposition \ref{p1}, we have $\be_{z^{t-1}}\left[\mathcal{E}(S_{t})\right]\leq M$.
		Since Assumption \ref{a4}, \ref{a2} (with $0<s<1$), and \ref{a3} hold, applying (\ref{eqq4}) in Proposition \ref{eqqq} with $\alpha=0$ and Proposition \ref{eta1} with the case of $v=1-s$, we obtain	
		\begin{equation}  \label{temp44}
			\begin{aligned}
				&\be_{z^{T}}\left[\left\|S_{T+1}-S^{\dagger}\right\|_{\mathrm{HS}}^2\right]
				\leq c_9\eta_{1}^{-2\widetilde{r}}(T+1)^{-2\widetilde{r}(1-\theta)}
				+c_{10}\left(\sqrt{c}\left(M-\mathcal{E}(S^{\dagger})\right)+\sigma^2\right)
				\\&\times \left[2^{2\theta}\eta_{1}^2(T+1)^{-2\theta}+c_5(\eta_{1},1-s,\theta)\begin{cases}
					(T+1)^{s-\theta(1+s)}, & \text{if }0<\theta<\frac{1}{2} \\
					(T+1)^{-(1-s)/2}\log (T+1), & \text{if } \theta=\frac{1}{2} \\
					(T+1)^{-(1-s)(1-\theta)}, & \text{if }\frac{1}{2}<\theta<1
				\end{cases}\right],
			\end{aligned}
		\end{equation}
		where in the last inequality is due to the fact $\eta_{T}^2=\eta_{1}^2T^{-2\theta}\leq2^{2\theta}\eta_{1}^2(T+1)^{-2\theta}$.
		Then, define $f_3(\theta)=-2\widetilde{r}(1-\theta)$,
        \begin{equation*}
            g_3(\theta)=\begin{cases}
            s-\theta(1+s), & \text{if } 0<\theta<\frac{1}{2}, \\
            -(1-s)/2, & \text{if } \theta=\frac{1}{2}, \\
            -(1-s)(1-\theta), & \text{if } \frac{1}{2}<\theta<1,
            \end{cases} 
        \end{equation*}
        and  $h_3(\theta)=-2\theta$.
		Choose 
		\begin{equation} \label{temp43}
			\theta=\mathop{\arg\min}_{\theta}\max\{f_3(\theta),g_3(\theta),h_3(\theta)\}=
			\begin{cases}
				\frac{2\widetilde{r}+s}{1+2\widetilde{r}+s}, & \text{if }\widetilde{r}<\frac{1-s}{2}, \\
				\frac{1}{2}, &\text{if } \widetilde{r}\geq\frac{1-s}{2}.
			\end{cases}
		\end{equation}
		Hence, by substituting (\ref{temp43}) into (\ref{temp44}), 
		\[
		\be_{z^{T}}\left[\left\|S_{T+1}-S^{\dagger}\right\|_{\mathrm{HS}}^2\right]
		\leq c_{\widetilde{r},s}
		\begin{cases}
			(T+1)^{-\frac{2\widetilde{r}}{1+2\widetilde{r}+s}}, &\text{if }\widetilde{r}<\frac{1-s}{2}, \\
			(T+1)^{-\frac{1-s}{2}}\log(T+1), &\text{if }\widetilde{r}\geq\frac{1-s}{2},
		\end{cases}
		\]
		where $c_{\widetilde{r},s}:=c_9\eta_{1}^{-2\widetilde{r}}+c_{10}\left[\sqrt{c}\left(M-\mathcal{E}(S^{\dagger})\right)+\sigma^2\right]\left(c_5(\eta_{1},1-s,\theta)+2^{2\theta}\eta_{1}^2\right).$

  Thus, we complete the proof.
		\end{proof}

	\begin{proof}[Proof of Theorem \ref{t4}.]
		From Proposition \ref{p2}, if the condition $\eta_{1}\leq\frac{1}{(1+14c)\log(T+1)}$ is satisfied, we have $\be_{z^{t-1}}\left[\mathcal{E}(S_{t})\right]\leq\widetilde{M}$ for any $t\in\bn_T$.         
		Applying the error decomposition (\ref{eqq4}) with $\alpha=0$ and $\theta=0$ in Proposition \ref{eqqq}, and using $\Psi_{1-s}(0)=1$ yields that
\begin{equation} \label{temp45}
    \begin{aligned}
\mathbb E_{z^T}&\left[
\|S_{T+1}-S^\dagger\|_{\mathrm{HS}}^2
\right]
\le
c_9\eta_1^{-2\widetilde r}(T+1)^{-2\widetilde r}\\
&+
c_{10}
\left(
\sqrt c\bigl(\widetilde M-\mathcal E(S^\dagger)\bigr)
+\sigma^2
\right)
\left[
\sum_{t=1}^{T-1}\eta_t^2
\Psi_{1-s}\!\left(\sum_{j=t+1}^T\eta_j\right)
+\eta_1^2
\right].
\end{aligned}
\end{equation}


\textbf{Case 1:}
Suppose that Assumption~\ref{a2} holds with $s=1$.
Since $\Psi_0(u)=1$ for all $u\ge0$ and
$c_{10}(1,0)=\sqrt c\,\operatorname{Tr}(L_C)$,
inequality \eqref{temp45} gives
\begin{equation}
\begin{aligned}
\mathbb E_{z^T}\!\left[
\|S_{T+1}-S^\dagger\|_{\mathrm{HS}}^2
\right]
&\le
c_9\eta_1^{-2\widetilde r}(T+1)^{-2\widetilde r}\\
&\quad+
\sqrt c\,\operatorname{Tr}(L_C)
\left(
\sqrt c\bigl(\widetilde M-\mathcal E(S^\dagger)\bigr)
+\sigma^2
\right)T\eta_1^2.
\end{aligned}
\end{equation}
Choose
$\eta_1=\eta_*(T+1)^{-(2\widetilde r+1)/(2\widetilde r+2)}$
with
\begin{equation*}
0<\eta_*\le
\frac{e(2\widetilde r+1)}
{(1+14c)(2\widetilde r+2)},
\end{equation*}
so that
$\eta_1\le[(1+14c)\log(T+1)]^{-1}$.
It follows that
\begin{equation*}
\mathbb E_{z^T}\!\left[
\|S_{T+1}-S^\dagger\|_{\mathrm{HS}}^2
\right]
\le
\widetilde c_{\widetilde r,1}
(T+1)^{-\widetilde r/(\widetilde r+1)},
\end{equation*}
where
\begin{equation*}
\begin{aligned}
\widetilde c_{\widetilde r,1}:={}&
c_9(0,\widetilde r,0)\eta_*^{-2\widetilde r}+
\sqrt c\,\operatorname{Tr}(L_C)
\left(
\sqrt c\bigl(\widetilde M-\mathcal E(S^\dagger)\bigr)
+\sigma^2
\right)\eta_*^2.
\end{aligned}
\end{equation*}

 \textbf{Case 2:} Suppose that $0<s<1$.
Then $\Psi_{1-s}(u)=(1+u^{1-s})^{-1}$.
Proposition~\ref{eta2} with $v=1-s$ gives
		\begin{equation} \label{temp46}
			\sum_{t=1}^{T-1}\frac{\eta_{t}^2}{1+\left(\sum_{j=t+1}^{T}\eta_{j}\right)^{1-s}}
			\leq c_6\eta_{1}^{1+s}(T+1)^{s}.
		\end{equation}
		Substituting (\ref{temp46}) into (\ref{temp45})  yields 
		\begin{align*}
			\be_{z^{T}}\left[\left\|S_{T+1}-S^{\dagger}\right\|_{\mathrm{HS}}^2\right]
			\leq&
			c_9\eta_{1}^{-2\widetilde{r}}(T+1)^{-2\widetilde{r}}
			\\&+c_{10}\left(\sqrt{c}\left(\widetilde{M}-\mathcal{E}(S^{\dagger})\right)+\sigma^2\right)\left(c_6\eta_{1}^{1+s}(T+1)^{s}+\eta_{1}^2\right).
		\end{align*}
		By choosing $\eta_{1}=\eta_{*}(T+1)^{-\frac{2\widetilde{r}+s}{1+2\widetilde{r}+s}}$ with $\eta_{*}\leq\frac{e(2\widetilde{r}+s)}{(1+14c)(1+2\widetilde{r}+s)}$ such that $\eta_{1}\leq\frac{1}{(1+14c)\log(T+1)}$, we have
		\begin{align*}
			\be_{z^{T}}\left[\left\|S_{T+1}-S^{\dagger}\right\|_{\mathrm{HS}}^2\right]
			\leq\widetilde{c}_{\widetilde{r},s}(T+1)^{-\frac{2\widetilde{r}}{1+2\widetilde{r}+s}},
		\end{align*}
		where $\widetilde{c}_{\widetilde{r},s}:=c_9\eta_{*}^{-2\widetilde{r}}+c_{10}\left(\sqrt{c}\left(\widetilde{M}-\mathcal{E}(S^{\dagger})\right)+\sigma^2\right)\left(c_6\eta_{*}^{1+s}+\eta_{*}^2\right)$.
	
 We thus complete the proof.
\end{proof}
	
\section{Proofs of Minimax Lower Bounds} \label{Section:lower bounds}

In this section, we prove Theorems \ref{thm-minimax1} and \ref{thm-minimax2} over
$\mathcal P_\omega$ and $\widetilde{\mathcal P}_\omega$,
respectively. The weak-regularity proof uses matrix packing
with an increasing number of output directions, while the
strong-regularity proof uses vector packing in a
one-dimensional output subspace.

We use the following Varshamov--Gilbert bound;
see \cite{tsybakov2004introduction} and \cite[Lemma 13.5]{duchi2016lecture}.


\begin{lemma} \label{minimax1}
		For every integer $m\geq8$, there exists a set $\Lambda\subseteq\H_m=\{-1,1\}^m$ such that $\lvert\Lambda\rvert\geq e^{\frac m8}$ and
		\begin{equation*}
			\|\iota-\iota'\|_1=2\sum_{j=1}^{m}\mathbf{1}_{\{\iota_j\not=\iota_j'\}}\geq\frac{m}{2}
		\end{equation*}
		for any $\iota\not=\iota'$ with $\iota,\iota'\in\Lambda$.
	\end{lemma}

The following
lemma refines the Varshamov--Gilbert bound by combining a spectral norm estimate for
Rademacher matrices with a maximal-packing argument. Its proof is deferred to Appendix
\ref{Proof of Lemma minimax4}.
\begin{lemma} \label{minimax4}
    There exists a universal constant $C_*>0$ such that, for all sufficiently
large $m$, there exists a set
$\Lambda\subset \mathcal{H}_{m^2}:=\{-1,1\}^{m\times m}$
such that $|\Lambda|\ge e^{m^2/8}$ and, for any $\iota\ne \iota'$ with
$\iota,\iota'\in\Lambda$,
\[
\|\iota-\iota'\|_1
=
2\sum_{i,j=1}^m {\bf 1}_{\{\iota_{ij}\ne \iota'_{ij}\}}
\ge \frac{m^2}{2}.
\]
Moreover, every $\iota\in\Lambda$ satisfies
\[
\|\iota\|\le C_*\sqrt m,
\]
where $\|\iota\|$ denotes the operator norm of the matrix $\iota$ as a linear map
from $\mathbb{R}^m$ to $\mathbb{R}^m$.
\end{lemma}

Let us introduce several symbols and definitions. Consider a set of probability measures $\P=\{\rho_S : S \in \mathcal{M}\}$ on a measurable space $(X, \mathcal{F})$, where each probability measure $\rho \in \P$ is determined by a specific $S \in \mathcal{M}$ with $\mathcal{M}$ being a nonempty set.  Let $d: \mathcal{M} \times \mathcal{M} \rightarrow [0, +\infty)$ represent a (semi-)distance. The Kullback–Leibler divergence between two probability measures $\rho_1$ and $\rho_2$ on $(X, \mathcal{F})$, denoted by $D_{kl}(\rho_1 \parallel \rho_2)$, is defined as
\[
D_{kl}(\rho_1 \parallel \rho_2) = \int_{X} \log\left(\frac{\mathrm{d}\rho_1}{\mathrm{d}\rho_2}\right) \mathrm{d}\rho_1,
\]
provided that $\rho_1$ is absolutely continuous with respect to $\rho_2$, and $D_{kl}(\rho_1 \parallel \rho_2) = \infty$ otherwise. Then
\[
D_{kl}(\rho_1^{\otimes T} \parallel \rho_2^{\otimes T}) = TD_{kl}(\rho_1 \parallel \rho_2)
\]
holds for any positive integer $T \in \mathbb{N}$.

The following proposition is referenced in \cite[Theorem 2.5]{tsybakov2004introduction} and \cite[Proposition 6.1]{blanchard2018optimal}.

\begin{proposition} \label{minimax2}
		Let $n\geq2$, assume that $\M$ contains $n+1$ elements $S_0,\cdots,S_n$ such that
		\begin{enumerate}
			\item[(1)] There exists $\zeta>0$, for any $0\leq i<j\leq n$, $d(S_i,S_j)\geq2\zeta$.
			\item[(2)] For any $j=1,2\cdots,n$, $\rho_j$ is absolutely continuous with respect to $\rho_0$, and 
			\[
			\frac{1}{n}\sum_{j=1}^{n}D_{kl}(\rho_j\parallel \rho_0)\leq w\log n
			\]
			for some $0<w<\frac{1}{8}$.
		\end{enumerate}
		Then, there holds
		\[
		\inf_{S_{\bf{z}}}\max_{0\leq j\leq n}\mathbb{P}_{\bf z \sim\rho_j}(d(S_{\bf{z}},S_j)\geq\zeta)\geq
		\frac{\sqrt{n}}{\sqrt{n}+1}\left(1-2w-\sqrt{\frac{2w}{\log n}}\right),
		\]
		where the infimum is taken over all estimators $S_{\bf{z}}$ based on the sample $\bf{z}$.
	\end{proposition}

Our proofs of the minimax lower bounds are based on Lemmas \ref{minimax1}--\ref{minimax4} and Proposition \ref{minimax2}.

 \subsection{Minimax Lower Bound  under Weak Regularity Condition} \label{Minimax Lower Bound  under Weak Regularity Condition}
	
 

Let $\M$ in Proposition \ref{minimax2} be $\mathcal{B}(\H_{1},\H_{2})$. For any $S,S'\in\mathcal{B}(\H_1,\H_2)$, define the semi-distance $d(S,S')=\left\|(S-S')L_C^{\frac12}\right\|_\mathrm{HS}$. One can easily see that 
	$d(S,S^{\dagger})^2=\mathcal{E}(S)-\mathcal{E}(S^{\dagger})$.
	

Fix $\omega\in\Omega$, and let $(\phi_k)_{k\ge1}$ be an
orthonormal basis of $\H_1$. Set
\[
q:=\min\left\{\frac12,\left(\frac Q2\right)^{1/s}\right\}.
\]
For each sufficiently large integer $m$, define
\[
\lambda_k=
\begin{cases}
q\,m^{-1/s},&1\le k\le2m,\\
0,&k>2m,
\end{cases}
\qquad
x=\sum_{k=1}^{2m}\sqrt{\lambda_k}\,\xi_k\phi_k,
\]
where the $\xi_k$ are independent standard Gaussian variables.
Then $L_C=\sum_{k=1}^{2m}\lambda_k\phi_k\otimes\phi_k$ and
\[
\operatorname{Tr}(L_C)=2q\,m^{1-1/s}\le1,
\qquad
\operatorname{Tr}(L_C^s)=2q^s\le Q.
\]
Moreover, $x$ satisfies Assumption \ref{a3} with constant $3$, and hence with the prescribed \(c\ge3\).
The same input construction will be used in Subsection \ref{Minimax Lower Bound  under Strong Regularity Condition}.
Let $(f_j)_{j\ge1}$ be an orthonormal basis of $\H_2$,
and choose the noise independently of $x$ as follows:
 \begin{equation}\label{epsilon}
     \epsilon=\sum_{l=1}^{m}\epsilon_l f_l\in \mathrm{span}\{f_l:1\leq l\leq m\},
 \end{equation}
 where $\epsilon_l\sim\N(0,\frac{\sigma^2}{m})$ is a centered independent Gaussian random variable for $l=1,\cdots,m$.  We write $\epsilon\sim\N(\bf{0_m},\frac{\sigma^2}{m}I_m)$ for simplicity. It is clear that $\mathbb{E}\left[\|\epsilon\|_{\H_{2}}^2\right]=\sigma^2$. The input and noise laws are fixed within each packing, but
may vary with $m$, while the parameter tuple $\omega$ remains fixed. We fix the distribution of $\epsilon$. Then, for any $S\in\mathcal{B}(\H_{1},\H_{2})$, $S$ specifies a distribution of $(x,y)$ via $y=Sx+\epsilon$. The entirety of distributions determined by $S$ with the fixed distributions of $x$ and $\epsilon$ as above are denoted by $\mathcal{P}$. For $S_{(i)}:=J_{(i)}L_C^r\in\mathcal{B}(\H_1,\H_2)$ with $\|J_{(i)}\|\leq R$, we define the corresponding joint distribution $\rho_i$ of $(x,y)$ on $\H_1\times\H_2$ through $y=J_{(i)}L_C^{r}x+\epsilon$. 
By construction, $\rho_i\in\mathcal P_\omega$.
Subsequently, we aim to identify $\zeta=\zeta(m)$, $n=n(m)$, and a set of $\rho_j\in\mathcal{P}_{\omega}$ determined by $S_{(j)}$, each dependent on $m$ for $j=0,\cdots, n$ as per Proposition \ref{minimax2}, to ascertain the lower bound of
    $$\inf_{S_{\bf z}}\max_{0\leq j\leq n}\mathbb{P}_{{\bf z} \sim\rho_j^{\otimes T}}\left(\left\|(S_{\bf z}-S_{(j)})L_C^{\frac12}\right\|_\mathrm{HS}\geq\zeta\right).$$
    Then we employ the inequality 
\begin{equation} \label{minimax inequality}
    \begin{split}
        &\inf_{S_{\bf z}}\sup_{\rho\in\mathcal{P}_{\omega}}\mathbb{P}_{{\bf z} \sim\rho^{\otimes T}}\left(\left\|(S_{\bf z}-S^\dagger) L_C^{\frac12}\right\|_\mathrm{HS}\geq\zeta\right)\\
       &\quad \geq\inf_{S_{\bf{z}}}\max_{0\leq j\leq n}\mathbb{P}_{{\bf{z}} \sim\rho_j^{\otimes T}}\left(\left\|(S_{\bf{z}}-S_{(j)})L_C^{\frac12}\right\|_\mathrm{HS}\geq\zeta\right)
    \end{split}
\end{equation} 
    to demonstrate the minimax lower bound,
	where the distribution on the right side with $\epsilon\sim\N(\bf{0_m},\frac{\sigma^2}{m}I_m)$ and the distribution of $x$ are previously determined, while the distribution $\rho$ on the left side only requires that $\epsilon$ and $x$ satisfy the conditions of $\mathcal{P}_{\omega}$, and that $y=S^\dagger x+\epsilon$. 
 Finally, we choose suitable $m=m(T)$ to achieve the desired results.
 
At the end of this subsection, we prove Theorem \ref{thm-minimax1} by using Proposition  \ref{minimax2} combined with the following Proposition \ref{minimax3}. We will give the proof of  Proposition \ref{minimax3} in Appendix \ref{Prop 7.4} based on Lemma \ref{minimax4}.
	
\begin{proposition} \label{minimax3}
		For every sufficiently large integer $m$,
        there exist some $J_{(1)},\cdots,J_{(L)}\in\B(\H_1,\H_{2})$ such that $L\geq e^{m^2/8}$ and for any distinct $i,j\in\{1,\ldots,L\}$,
		\begin{enumerate}
			\item[(1)] $d(S_{(i)},S_{(j)})\geq2^{-\frac{2r+1}{2s}}q^{r+1/2}\frac{R}{C_*}m^{-\frac{2r+1}{2s}+1/2}$, where $S_{(i)}=J_{(i)}L_C^r$.
			\item[(2)] 
            $\rho_i\in\mathcal P_\omega$.
			\item[(3)] $D_{kl}\left(\rho_i^{\otimes T}\parallel \rho_j^{\otimes T}\right)\leq\frac{2R^2}{C_*^2\sigma^2}q^{2r+1}m^{-\frac{2r+1}{s}+2}T$.
		\end{enumerate}
	\end{proposition}

Now we are in the position to prove Theorem \ref{thm-minimax1}.

\begin{proof}[Proof of Theorem \ref{thm-minimax1}]
		Let $J_{(1)},\cdots,J_{(L)}$ be given in Proposition \ref{minimax3}. Let		
		\[\zeta=2^{-\frac{2r+1+2s}{2s}}q^{r+1/2}\frac{R}{C_*}m^{-\frac{2r+1}{2s}+1/2},\]
		$w=\frac{1}{16}<\frac{1}{8}$ and $n=L-1$ in Proposition \ref{minimax2}, then $d(S_{(i)},S_{(j)})\geq 2\zeta$ for any $1\leq i<j\leq L$. The condition 
		\[
		\frac{1}{n}\sum_{j=2}^{n+1}D_{kl}(\rho_j^{\otimes T}\parallel \rho_1^{\otimes T})\leq w\log n
		\]
		in Proposition \ref{minimax2} is satisfied when
		\begin{equation} \label{condition3}
			\frac{2R^2}{C_*^2\sigma^2}q^{2r+1}m^{-\frac{2r+1}{s}+2}T\leq \frac{1}{16}\log (L-1). 
		\end{equation}
		Since $L\geq e^{m^2/8}$, it is obvious that $\log (L-1)\geq\frac{m^2}{16}$. Therefore, if
		\[
		m^{\frac{2r+1}{s}}\geq\frac{512R^2}{C_*^2\sigma^2}q^{2r+1}T,
		\]
		the condition \eqref{condition3} is satisfied. For $T$ sufficiently large, we choose $m$ as $m=8\lceil(\frac{512R^2}{C_*^2\sigma^2}q^{2r+1}T)^{\frac{s}{2r+1}}\rceil$. By Proposition \ref{minimax2}, it follows that
		\begin{equation*}
			\begin{split}
			     &\inf_{S_{\bf z}}\max_{1\leq j\leq n+1}\mathbb{P}_{\bf z \sim\rho_j^{\otimes T}}\left(\left\|(S_{\bf z}-S_{(j)})L_C^{\frac12}\right\|_\mathrm{HS}\geq2^{-\frac{2r+1+2s}{2s}}q^{r+1/2}\frac{R}{C_*}m^{-\frac{2r+1}{2s}+1/2}\right)
				\\
				&\quad \geq
				\frac{\sqrt{n}}{\sqrt{n}+1}\left(1-2w-\sqrt{\frac{2w}{\log n}}\right)
				\geq\frac{e^{\frac{m}{32}}}{e^{\frac{m}{32}}+1}\left(\frac{7}{8}-\sqrt{\frac{2}{m}}\right)
				\geq\frac{3e^{\frac{1}{4}}}{8e^{\frac{1}{4}}+8}.
			\end{split}
		\end{equation*}
		Recall that $\mathcal{E}(S_{\bf z})-\mathcal{E}(S^{\dagger})=\left\|(S_{\bf z}-S_{(j)})L_C^{\frac12}\right\|_\mathrm{HS}^2$. Applying \eqref{minimax inequality}, there holds  
		\begin{equation*}
                \inf_{S_{\bf z}}\sup_{\rho\in\mathcal{P}_{\omega}}\mathbb{P}_{\bf z \sim\rho^{\otimes T}}\left(\mathcal{E}(S_{\bf z})-\mathcal{E}(S^{\dagger})\geq\gamma_\omega T^{-\frac{1+2r-s}{2r+1}}\right)
			\geq\frac{3e^{\frac{1}{4}}}{8e^{\frac{1}{4}}+8},
		\end{equation*}
		where infimum is taken over all estimators $S_{\bf z}$ based on the sample $\bf z$ and $\gamma_\omega>0$ is a constant independent of $T$.
        Since $\frac{3e^{\frac{1}{4}}}{8e^{\frac{1}{4}}+8}$ is independent of $T$ and $\omega$, we deduce that
        \[
        \inf_{\omega\in\Omega}\liminf_{T\to \infty}\inf_{S_{\bf z}}\sup_{\rho\in\mathcal{P}_{\omega}}\mathbb{P}_{\bf z \sim\rho^{\otimes T}}\left(\mathcal{E}(S_{\bf z})-\mathcal{E}(S^{\dagger})\geq\gamma_\omega T^{-\frac{1+2r-s}{2r+1}}\right)
			>0.
        \]
        
        The proof is thus finished.
	\end{proof}
	
	\subsection{Minimax Lower Bound  under Strong Regularity Condition} \label{Minimax Lower Bound  under Strong Regularity Condition}

 In this subsection, we provide the proof for the minimax lower bound of $\left\|(S-S^\dagger)L_C^{\alpha}\right\|_{\mathrm{HS}}^2$  under the strong regularity condition. The prediction and estimation errors correspond to $\alpha=1/2$ and $\alpha=0$, respectively.

 The proof idea for this scenario does not fundamentally differ from that in the case where $\H_2=\mathbb{R}$. Specifically, we further assume that the range of $S^\dagger$ is contained in $\operatorname{span}\{f_1\}$ and that $\epsilon$ follows a fixed Gaussian distribution, which leads to a subset of $\widetilde{\mathcal{P}}_{\omega}$, denoted by $\mathcal{P}^\prime_{\omega}$ for any $\omega\in\widetilde{\Omega}.$ To elaborate, we define $\mathcal{P}^\prime_{\omega}$ as follows: Recall the conditions 1, 2, $3^\prime$, and 4 for the joint distribution of observation $(x,y)$, which are delineated in Subsection \ref{subsection: minimax}. Then the distribution $\mathcal{P}$ is parameterized by $\omega$ such that the conditions 1, 2, $3^\prime$, 4 for $\widetilde{\mathcal{P}}_{\omega}$ are satisfied. Additionally, it is stipulated that $\mathrm{Ran}(S^\dagger)\subseteq\mathrm{span}(f_1)$ and $\epsilon=\epsilon_1f_1$ where $\epsilon_1\sim\N(0,\sigma^2)$ is Gaussian on $\mathbb{R}$ and $f_1$ is a fixed unit vector in $\mathcal{H}_2$. Under the assumption of $\mathrm{Ran}(S^\dagger)\subseteq\mathrm{span}(f_1)$, the condition $3'$ in Subsection \ref{subsection: minimax} can be represented as
\begin{itemize}
        \item[$3^{\prime\prime}$.] There exist $g^\dagger,\beta^\dagger\in \H_1$ such that
$\beta^\dagger=L_C^{\widetilde r}g^\dagger$,
$S^\dagger x=\langle\beta^\dagger,x\rangle_{\H_1}f_1$,
and $\|g^\dagger\|_{\H_1}\le R$.
\end{itemize}
Indeed, one may take $\widetilde J=f_1\otimes g^\dagger$,
for which
$\|\widetilde J\|_{\mathrm{HS}}=\|g^\dagger\|_{\H_1}\le R$.
For any fixed $\omega \in \widetilde{\Omega}$, the collection of distributions $\rho$ that satisfy conditions 1, 2, $3^{\prime\prime}$, and 4, as well as $\epsilon=\epsilon_1f_1$ where $\epsilon_1$ is distributed according to $\N(0,\sigma^2)$ for the joint distribution of observations $(x,y)$, is denoted as $\mathcal{P}^\prime_{\omega}$. This specification, due to the additional constraints on the image space of $S^\dagger$ and the distribution of $\epsilon$, qualifies it as a subset of $\widetilde{\mathcal{P}}_{\omega}$.

Establishing a lower bound for a specific case immediately provides a lower bound for the general case within our framework. Specifically, for any $\zeta \geq 0$ and $\omega$ within the extended parameter space $\widetilde{\Omega}$, the following inequality holds
\begin{equation} \label{m1}
    \inf_{S_{\mathbf{z}}}\sup_{\rho \in \widetilde{\mathcal{P}}_{\omega}}\mathbb{P}_{\mathbf{z} \sim \rho^{\otimes T}}\left(\left\|(S_{\mathbf{z}}-S^\dagger) L_C^{\alpha}\right\|_{\mathrm{HS}} \geq \zeta\right)
    \geq \inf_{S_{\mathbf{z}}}\sup_{\rho \in \mathcal{P}^\prime_{\omega}}\mathbb{P}_{\mathbf{z} \sim \rho^{\otimes T}}\left(\left\|(S_{\mathbf{z}}-S^\dagger) L_C^{\alpha}\right\|_{\mathrm{HS}} \geq \zeta\right).
\end{equation}

Let $\pi_1: \mathcal{H}_{2} \rightarrow \mathcal{H}_{2}$ be the projection operator defined by $\pi_1(y) = \langle y, f_1 \rangle f_1$ for any $y \in \mathcal{H}_2$. 
Let $S_{\mathbf z}$ take values in $\mathcal B(\H_1,\H_2)$
when $\alpha=1/2$, and in
$\mathcal B_{\mathrm{HS}}(\H_1,\H_2)$ when $\alpha=0$.
Since
$\pi_1S_{\mathbf z}
=f_1\otimes(S_{\mathbf z}^*f_1)$
has rank at most one, we have
\[
\left\|(S_{\mathbf{z}}-S^\dagger)L_C^{\alpha}\right\|_{\mathrm{HS}} \geq \left\|(\pi_1 S_{\mathbf{z}}-S^\dagger)L_C^{\alpha}\right\|_{\mathrm{HS}}.
\]

Therefore, on the right-hand side of inequality \eqref{m1}, it suffices to consider the infimum over all $S_{\mathbf{z}}$ with $\mathrm{Ran}(S_{\mathbf{z}}) \subseteq \mathrm{span}\{f_1\}$. According to the Riesz Representation Theorem, a unique $\beta_{\mathbf{z}} \in \mathcal{H}_1$ exists such that $S_{\mathbf{z}}x = \langle \beta_{\mathbf{z}}, x \rangle f_1$. Since
\[
\left\|(S_{\mathbf{z}}-S^\dagger)L_C^{\alpha}\right\|_{\mathrm{HS}} = \left\|L_C^{\alpha}(\beta_{\mathbf{z}}-\beta^\dagger)\right\|_{\mathcal{H}_1},
\]it is then sufficient to estimate:
\begin{equation} \label{m2}
    \inf_{\beta_{\mathbf{z}}}\sup_{\rho \in \mathcal{P}^\prime_{\omega}}\mathbb{P}_{\mathbf{z} \sim \rho^{\otimes T}}\left(\left\|L_C^{\alpha}(\beta_{\mathbf{z}}-\beta^\dagger)\right\|_{\mathcal{H}_1} \geq \zeta\right).
\end{equation}

Now, we begin to estimate the lower bound \eqref{m2}. 
For each $m\ge8$, restrict the input distribution to the
Gaussian construction in Subsection \ref{Minimax Lower Bound  under Weak Regularity Condition}, with the fixed
parameters $s$ and $Q$ of $\omega\in\widetilde\Omega$.
Thus all candidates in the packing share the same $L_C$. Define $\mathcal{M} = \mathcal{H}_1$ and introduce a semi-distance $d: \mathcal{M} \times \mathcal{M} \rightarrow [0, \infty)$, defined by $d(\beta_1, \beta_2) = \left\|L_C^{\alpha}(\beta_1 - \beta_2)\right\|_{\mathcal{H}_1}$, as presented in Proposition \ref{minimax2}.

\begin{proposition} \label{minimax5}
		For any fixed $m\geq8$, there exist $\beta_1,\cdots,\beta_L\in\H_1$ such that $L\geq e^{m/8}$ and for any distinct $i,j\in\{1,\ldots,L\}$,
		\begin{enumerate}
			\item[(1)] $d(\beta_i,\beta_j)\geq2^{-\frac{\widetilde{r}+\alpha}{s}}q^{\widetilde{r}+\alpha}Rm^{-\frac{\widetilde{r}+\alpha}{s}}$, where $\beta_i=L_C^{\widetilde{r}}g_i$.
			\item[(2)] 
            $\rho_i\in\mathcal P'_\omega$.
			\item[(3)] $D_{kl}\left(\rho_i^{\otimes T}\parallel \rho_j^{\otimes T}\right)\leq\frac{2R^2}{\sigma^2}q^{2\widetilde{r}+1}m^{-\frac{2\widetilde{r}+1}{s}}T$.
		\end{enumerate}
	\end{proposition}
	
	The proof of this proposition can be found in Appendix \ref{Prop 7.5}. Now we are ready to prove Theorem \ref{thm-minimax2}.
	
	\begin{proof}[Proof of Theorem \ref{thm-minimax2}]
		Let $\beta_1,\cdots,\beta_L$ be given in Proposition \ref{minimax5}. Set	
		\[\zeta=2^{-\frac{\widetilde{r}+\alpha+s}{s}}q^{\widetilde{r}+\alpha}Rm^{-\frac{\widetilde{r}+\alpha}{s}}\]
		$w=\frac{1}{16}<\frac{1}{8}$ and $n=L-1$ in Proposition \ref{minimax2}. The condition 
		\[
		\frac{1}{n}\sum_{j=2}^{n+1}D_{kl}(\rho_j^{\otimes T}\parallel \rho_1^{\otimes T})\leq w\log n
		\]
		in Proposition \ref{minimax2} is satisfied if
		\begin{equation} \label{condition99}
			\frac{2R^2}{\sigma^2}q^{2\widetilde{r}+1}m^{-\frac{2\widetilde{r}+1}{s}}T\leq \frac{1}{16}\log (L-1). 
		\end{equation}
		Since $L\geq e^{m/8}$ and $m\geq8$, there holds $\log (L-1)\geq\frac{m}{16}$. Therefore, if
		\[
		m^{\frac{2\widetilde{r}+1+s}{s}}\geq\frac{512R^2}{\sigma^2}q^{2\widetilde{r}+1}T,
		\]
		the condition \eqref{condition99} is satisfied. We choose $m=8\lceil(\frac{512R^2}{\sigma^2}q^{2\widetilde{r}+1}T)^{\frac{s}{2\widetilde{r}+1+s}}\rceil$. By Proposition \ref{minimax2}, we have
		\begin{equation*}
			\begin{split}
				&\inf_{\beta_{\bf z}}\max_{1\leq j\leq n+1}\mathbb{P}_{{\bf z}\sim\rho_j^{\otimes T}}
                \left(\left\|L_C^{\alpha}(\beta_{\bf z}-\beta_j)\right\|_{\H_1}\geq2^{-\frac{\widetilde{r}+\alpha+s}{s}}q^{\widetilde{r}+\alpha}Rm^{-\frac{\widetilde{r}+\alpha}{s}}\right)
				\\
				&\geq
				\frac{\sqrt{n}}{\sqrt{n}+1}\left(1-2w-\sqrt{\frac{2w}{\log n}}\right)
				\\
				&\geq\frac{e^{\frac{m}{32}}}{e^{\frac{m}{32}}+1}\left(\frac{7}{8}-\sqrt{\frac{2}{m}}\right)
				\geq\frac{3e^{\frac{1}{4}}}{8e^{\frac{1}{4}}+8}.
			\end{split}
		\end{equation*}
		Therefore, for any $\omega\in\widetilde{\Omega}$, 
		\begin{equation*}
                \inf_{\beta_{\bf z}}\sup_{\rho\in\P'_\omega}\mathbb{P}_{{\bf z}\sim\rho^{\otimes T}}\left(\left\|L_C^{\alpha}(\beta_{\bf z}-\beta^\dagger)\right\|_{\H_1}^2\geq\gamma_\omega^\prime T^{-\frac{2\widetilde{r}+2\alpha}{1+2\widetilde{r}+s}}\right)\geq\frac{3e^{\frac{1}{4}}}{8e^{\frac{1}{4}}+8},
		\end{equation*}
            where $\gamma_\omega^\prime$ is a constant independent of $T$.
            
            We finish this proof by applying \eqref{m1} and \eqref{m2} and setting $\alpha=1/2$ or $0$, which yields the minimax lower bounds of prediction error and estimation error respectively.
	\end{proof}

\section{Conclusion and Discussion} \label{Conclusion and Discussion}

In this work, we developed a convergence and minimax analysis of SGD for linear operator regression between separable Hilbert spaces, with particular emphasis on the distinction between weak and strong regularity conditions. The strong regularity condition in Assumption \ref{a4} was introduced in the literature \cite{mollenhauer2022learning}, but may be unnecessarily restrictive when only prediction error is considered.
Under the remaining assumptions and step-size conditions of
the corresponding theorems, Assumption \ref{a1} suffices for the
prediction bounds in place of Assumption \ref{a4}. 
However, Assumption \ref{a1} does not guarantee that the operator is a Hilbert-Schmidt operator, which makes it infeasible to characterize and compute the estimation error using the Hilbert-Schmidt norm. In such cases, Assumption \ref{a4} provides a sufficient regularity condition for deriving the stated Hilbert–Schmidt estimation rates. Additionally, \cite{mollenhauer2022learning} did not employ Assumption \ref{a2} to characterize the smoothness of the input. We incorporate this assumption to achieve the established minimax lower bounds. 
For nonlinear regression targets, Theorem~\ref{Thm:nonlinear} establishes convergence
to a bounded best linear approximation satisfying the corresponding
regularity condition. Under the finite fourth-moment condition on the
approximation residual and the stated step-size restrictions, the
prediction and Hilbert--Schmidt estimation bounds retain the rates
of their corresponding linear results.

Theorems \ref{t4} and \ref{thm-minimax2} establish the minimax-optimal
Hilbert--Schmidt estimation rate under strong regularity
for $0<s\le1$; the lower bound already holds with
one-dimensional output.
Under weak regularity, the matrix packing exploits an
increasing number of output directions while keeping
$\|J\|$ bounded.
Indeed, the operators in \eqref{J_i} satisfy
$\|J_{(i)}\|\le R$ but
$\|J_{(i)}\|_{\mathrm{HS}}=R\sqrt m/C_*$,
even though every $S_{(i)}$ has finite rank.

Regarding the generality of the model \eqref{linear} and our approach, we provide the following remarks. Our results are applicable to many cases, such as regression with vector-valued RKHS, functional linear models, regression with scalar-valued RKHS, and the approximation of nonlinear operators using linear operators. In fact, our method can be extended to more general scenarios. Our subsequent companion work \cite{yang2025learning} studies regularized SGD with operator-valued kernels. In contrast, the present paper focuses on unregularized direct operator regression and its weak/strong regularity-condition analysis.
In practice, data is often collected in the form of high-dimensional vectors rather than entire functions. However, studying at the function level is able to capture the essential characteristics of variables without being affected by specific discretization or sampling methods. Our future work involves incorporating an encoder-decoder structure to transform the problem into a regression problem between finite-dimensional spaces through sampling.

	\appendix
\section*{Appendix: Proofs of Propositions and Lemmas}\label{section: appendix}
\setcounter{equation}{0}
\setcounter{theorem}{0}
\addtocounter{section}{1}
The appendix includes proofs of some propositions, lemmas, and Theorem \ref{Thm:nonlinear} that appeared in the main body of this paper. 
\subsection{Proof of Proposition \ref{1}} \label{Appendix A}

We first prove the following proposition.

\begin{proposition} \label{prop A1}
    For any fixed $c>0$, the following three statements are equivalent.
\begin{enumerate}
\item[(1)]
For every $f\in\H_1$,
\begin{equation*}
\mathbb{E}\bigl[\langle x,f\rangle_{\H_1}^4\bigr]
\le
c\left(
\mathbb{E}\bigl[\langle x,f\rangle_{\H_1}^2\bigr]
\right)^2.
\end{equation*}

\item[(2)]
For every separable Hilbert space $\H$ and every bounded linear
operator $A:\H_1\to\H$,
\begin{equation*}
\mathbb{E}\bigl[\|Ax\|_{\H}^4\bigr]
\le
c\left(\mathbb{E}\bigl[\|Ax\|_{\H}^2\bigr]\right)^2.
\end{equation*}

\item[(3)]
There exists a nontrivial separable Hilbert space $\H$ such that,
for every bounded linear operator $A:\H_1\to\H$,
\begin{equation*}
\mathbb{E}\bigl[\|Ax\|_{\H}^4\bigr]
\le
c\left(\mathbb{E}\bigl[\|Ax\|_{\H}^2\bigr]\right)^2.
\end{equation*}
\end{enumerate}
\end{proposition}
	
\begin{proof}
Assume (1), and let $\H$ be a separable Hilbert space and
$A:\H_1\to\H$ a bounded linear operator. Let
$(e_j)_{j\in\mathcal{I}}$ be an orthonormal basis of $\H$, where
$\mathcal{I}$ is finite or countable, and set
$Z_j=\langle Ax,e_j\rangle_{\H}
=\langle x,A^*e_j\rangle_{\H_1}$.
By Parseval's identity, Tonelli's theorem, Cauchy--Schwarz, and (1),
\begin{equation*}
\begin{aligned}
\mathbb{E}\|Ax\|_{\H}^4
&=\sum_{i,j\in\mathcal{I}}\mathbb{E}[Z_i^2Z_j^2]\\
&\le
\sum_{i,j\in\mathcal{I}}
\left(\mathbb{E}Z_i^4\,\mathbb{E}Z_j^4\right)^{1/2}\\
&\le
c\left(\sum_{j\in\mathcal{I}}\mathbb{E}Z_j^2\right)^2
=c\left(\mathbb{E}\|Ax\|_{\H}^2\right)^2.
\end{aligned}
\end{equation*}
The right-hand side is finite because
$\mathbb{E}\|Ax\|_{\H}^2
\le\|A\|^2\mathbb{E}\|x\|_{\H_1}^2<\infty$.
Thus, (1) implies (2).

Statement (2) implies (3) by taking $\H=\mathbb{R}$.
To prove that (3) implies (1), choose a unit vector $g\in\H$.
For any $f\in\H_1$, define
$Ah=\langle h,f\rangle_{\H_1}g$ for $h\in\H_1$.
Then $A$ is bounded and
$\|Ax\|_{\H}=|\langle x,f\rangle_{\H_1}|$, so (3) yields (1).
All implications preserve the constant $c$.
\end{proof}

	Next, we prove that random variable in Hilbert spaces satisfies the moment condition in Assumption \ref{a3} if it is strictly sub-Gaussian. First, we give some notations and definitions.
	
 \begin{definition}[Sub-Gaussian random variable in Hilbert spaces \cite{chen2021hanson}] \label{def1}
		Let $x$ be a random variable in some real separable Hilbert space $\H$ and $\Gamma:\H\rightarrow\H$ be a positive semi-definite trace class linear operator. Then, $x$ is sub-Gaussian with respect to $\Gamma$ if there exists a constant $\alpha>0$ such that for any $f\in\H$,
		\begin{equation*}
			\mathbb{E}\left[e^{\langle f,x-\mathbb{E}[x]\rangle}\right]\leq e^{\alpha^2\langle\Gamma f,f\rangle/2}.
		\end{equation*}
	\end{definition}
	We denote the covariance operator of $x$ by  $\Sigma:\H \rightarrow \H$ with
	\[
	\Sigma=\mathbb{E}\left[(x-\mathbb{E}[x])\otimes (x-\mathbb{E}[x])\right].
	\]
	The next definition is a natural extension of strictly sub-Gaussian random variables in $\mathbb{R}$ \cite{buldygin2000metric}.
	\begin{definition}[Strictly sub-Gaussian random variable in Hilbert spaces] \label{def2}
		Let $x$ be a sub-Gaussian random variable in Definition \ref{def1}, if there holds
		\begin{equation*}
			\alpha^2\Gamma=\Sigma,
		\end{equation*}
		then, $x$ is strictly sub-Gaussian.
	\end{definition}
	Then, we give a sufficient condition of Assumption \ref{a3}.
	\begin{proposition}
		Let $x$ be a sub-Gaussian random variable in Definition \ref{def1}, if 
		\begin{equation*}
			\Gamma\preceq C\Sigma
		\end{equation*}
	for some fixed constant $C>0$, then there exists a constant $c>0$, such that for any $f\in \H$,
	\begin{equation*}
		\mathbb{E}\left[\left\langle x-\mathbb{E}[x],f\right\rangle^4\right]
		\leq c\left(\mathbb{E}\left[\left\langle x-\mathbb{E}[x],f\right\rangle^2\right]\right)^2.
	\end{equation*}
	\end{proposition}
	\begin{proof}
    It suffices to prove the stated inequality for the centered random
variable $x-\mathbb{E}[x]$, which we denote by $x$ throughout this
proof. Fix $f\in\H$. If $\langle\Gamma f,f\rangle=0$, the
sub-Gaussian bound implies $\langle x,f\rangle=0$ almost surely,
and the conclusion is immediate. We therefore assume
$\langle\Gamma f,f\rangle>0$ and first establish a tail bound for
$\langle x,f\rangle$. 

		By the definition of sub-Gaussian random variable, there holds
		\begin{equation} \label{tail bound}
			\begin{aligned}
				\mathbb{P}\left(\langle x,f\rangle\geq t\right)
				&\leq\inf_{\lambda>0}e^{-\lambda t}\mathbb{E}\left[e^{\lambda\langle x,f\rangle}\right]
				\\&\leq\inf_{\lambda>0}e^{\frac{\lambda^2\alpha^2}{2}\langle\Gamma f,f\rangle-\lambda t}
				\\&=e^{-\frac{t^2}{2\alpha^2\langle \Gamma f,f\rangle}}
			\end{aligned}
		\end{equation}
		Similarly we have $\mathbb{P}\left(\langle x,f\rangle\leq -t\right)\leq e^{-\frac{t^2}{2\alpha^2\langle \Gamma f,f\rangle}}$. Using (\ref{tail bound}) we have
		\begin{equation*}
			\begin{aligned}
				\mathbb{E}\left[\langle x,f\rangle^4\right]&=
				\int_{0}^{\infty}\mathbb{P}\left(\lvert\langle x,f\rangle\rvert\geq t^{1/4}\right)\mathrm{d}t
				\\&=\int_{0}^{\infty}2t\mathbb{P}\left(\lvert\langle x,f\rangle\rvert\geq t^{1/2}\right)\mathrm{d}t
				\\&\leq4\int_{0}^{\infty}te^{-\frac{t}{2\alpha^2\langle \Gamma f,f\rangle}}\mathrm{d}t
				\\&=16\alpha^4\langle \Gamma f,f\rangle^2.
			\end{aligned}
		\end{equation*}
		Since $\Gamma\preceq C\Sigma$,
		\begin{equation*}
			\langle \Gamma f,f\rangle\leq C\langle \Sigma f,f\rangle=C\mathbb{E}\left[\langle x,f\rangle^2\right].
		\end{equation*}
		Therefore,
		\begin{equation*}
			\mathbb{E}\left[\langle x,f\rangle^4\right]
			\leq16\alpha^4C^2\left(\mathbb{E}\left[\langle x,f\rangle^2\right]\right)^2,
		\end{equation*}
		which completes our proof.
	\end{proof}

We now return to the original, not necessarily centered random variable
$x$ and verify the strictly sub-Gaussian case. For $f\in\H$, set
$Z=\langle x-\mathbb{E}[x],f\rangle$,
$m=\langle\mathbb{E}[x],f\rangle$, and
$v=\mathbb{E}Z^2=\langle\Sigma f,f\rangle$.
By Definitions~\ref{def1} and \ref{def2},
$\mathbb{E}e^{tZ}\le e^{vt^2/2}$ for every $t\in\mathbb{R}$.
The moment generating function is finite in a neighborhood of zero,
and its Taylor expansion gives
\begin{equation*}
0\le e^{vt^2/2}-\mathbb{E}e^{tZ}
=-\frac{\mathbb{E}Z^3}{6}t^3
+\frac{3v^2-\mathbb{E}Z^4}{24}t^4+o(t^4).
\end{equation*}
Considering both positive and negative $t$ tending to zero yields
$\mathbb{E}Z^3=0$ and $\mathbb{E}Z^4\le3v^2$. Consequently,
\begin{equation*}
\begin{aligned}
\mathbb{E}[\langle x,f\rangle^4]
&=\mathbb{E}(Z+m)^4
=\mathbb{E}Z^4+6m^2v+m^4\\
&\le3v^2+6m^2v+m^4
\le3(v+m^2)^2\\
&=3\left(\mathbb{E}[\langle x,f\rangle^2]\right)^2.
\end{aligned}
\end{equation*}
Taking $\H=\H_1$ and applying Proposition~\ref{prop A1} shows that Assumption~\ref{a3}
holds with $c=3$, without requiring $\mathbb{E}[x]=0$.
This proves the sufficient condition in Proposition~\ref{1}.


	\subsection{Proof of Proposition \ref{2}} \label{Appendix B}
	
	\begin{proof}
		Let $\H_0=\mathrm{span}\{K(x,\cdot)y:x\in\mathcal{X},y\in\H_2\}$ and $\B_0=\mathrm{span}\{y\otimes\phi(x):x\in\mathcal{X},y\in\H_2\}$, then it's clear that $\H_K=\overline{\H_0}$ and $\B_{\mathrm{HS}}(\H_1,\H_2)=\overline{\B_0}$. 
		
		We define the mapping $T_0:\H_0\rightarrow\B_0,\  \sum_{i=1}^{n}\alpha_iK(x_i,\cdot)y_i\mapsto\sum_{i=1}^{n}\alpha_iy_i\otimes\phi(x_i)$ for any $n\in\bn$, $x_1,\cdots x_n\in\mathcal{X}$ and $y_1,\cdots y_n\in\H_2$. It's easy to verify that $T_0$ is well-defined and a linear operator. Moreover, for any $x,\widetilde{x}\in\mathcal{X}$  and $y,\widetilde{y}\in\H_2$,
		\begin{equation*}
			\begin{aligned}
				\left\langle T_0(K(x,\cdot)y),T_0(K(\widetilde{x},\cdot)\widetilde{y})\right\rangle_{\mathrm{HS}}
				&=\langle y\otimes\phi(x),\widetilde{y}\otimes\phi(\widetilde{x})\rangle_{\mathrm{HS}}
				=\mathcal{K}(x,\widetilde{x})\langle y,\widetilde{y}\rangle_{\H_{2}} \\
				&=\langle \mathcal{K}(x,\widetilde{x})y,\widetilde{y}\rangle_{\H_{2}}=\langle K(x,\cdot)y,K(\widetilde{x},\cdot)\widetilde{y}\rangle_{\H_K}.
			\end{aligned}
		\end{equation*}
		By extending $T_0$ to $T:\H_K\rightarrow\B_{\mathrm{HS}}(\H_1,\H_2)$, we conclude that $\H_K$ is isometric to $\B_{\mathrm{HS}}(\H_1,\H_2)$.
		
		Then, we prove $h(x)=(Th)(\phi(x))$. For any $y\in\H_2$,
		\begin{align*}
			\langle y,h(x)\rangle_{\H_{2}}&=\langle K(x,\cdot)y,h\rangle_{\H_K}=\left\langle T(K(x,\cdot)y),Th\right\rangle_{\mathrm{HS}}=\langle y\otimes\phi(x),Th\rangle_{\mathrm{HS}} \\
			&=\mathrm{Tr}\left((y\otimes\phi(x))^*(Th)\right)=\langle y, (Th)\phi(x)\rangle_{\H_{2}},
		\end{align*}
		where the property $\langle y,f(x)\rangle_{\H_{2}}=\langle K(x,\cdot)y,f\rangle_{\H_K}$ is used in the first equality. Thus, we can conclude that $h(x)=(Th)\phi(x)$. The uniqueness obviously holds. The proof is then finished.
	\end{proof}

\subsection{Proof of Lemma \ref{lemma4.1}} \label{Lemma 4.1}
\begin{proof}
		The equality (\ref{lemma 1.1}) is easily verified by the form of algorithm (\ref{sgd}).
		Note that $S_t$ is dependent on $Z^{t-1}$ and independent of $z_t=(x_t,y_t)$, by taking expectation with respect to $z_t$, we have
		\begin{align*}
			\mathbb{E}_{z_t}[\B_t]&=(S_t-S^\dagger)L_C+\mathbb{E}_{z_t}\left[(y_t-S_tx_t)\otimes x_t\right]\\
			&=(S_t-S^\dagger)L_C+\mathbb{E}_{z_t}\left[\left((S^\dagger-S_t)x_t+\epsilon_t\right)\otimes x_t\right].
		\end{align*}
		Since $\mathbb{E}[\epsilon_t]=0$ and $\epsilon_t$ is independent of $x_t$, $\mathbb{E}_{z_t}[\epsilon_t\otimes x_t]=\mathbb{E}_{x_t}\left[\mathbb{E}_{\epsilon_t}[\epsilon_t]\otimes x_t\right]=0$. Hence,
		\[
		\mathbb{E}_{z_t}[\B_t]=(S_t-S^\dagger)L_C+(S^\dagger-S_t)L_C=0.
		\]
		This proves the lemma.
	\end{proof}

\subsection{Proof of Proposition \ref{Proposition error 1}}
\label{Prop 4.2}
\begin{proof}
		From the martingale decomposition (\ref{form}) we have
		\[
		(S_{T+1}-S^{\dagger})L_C^{\alpha}=-S^{\dagger}L_C^{\alpha}\prod_{t=1}^{T}(I-\eta_{t}L_C)+\sum_{t=1}^{T}\eta_t\B_{t}L_C^{\alpha}\prod_{j=t+1}^{T}(I-\eta_{j}L_C).
		\]		
		By the equality (\ref{form2}), the error of $S_{T+1}$ can be rewritten as
		\begin{align*}
			&\be_{z^{T}}\left[\left\|(S_{T+1}-S^{\dagger})L_C^{\alpha}\right\|_{\mathrm{HS}}^{2}\right]
			\\&=\be_{z^{T}}\left[\left\|-S^{\dagger}L_C^{\alpha}\prod_{t=1}^{T}(I-\eta_{t}L_C)+\sum_{t=1}^{T}\eta_t\B_{t}L_C^{\alpha}\prod_{j=t+1}^{T}(I-\eta_{j}L_C)\right\|_{\mathrm{HS}}^{2}\right]
			\\&=\left\|S^{\dagger}L_C^{\alpha}\prod_{t=1}^{T}(I-\eta_{t}L_C)\right\|_{\mathrm{HS}}^{2}+\be_{z^{T}}\left[\left\|\sum_{t=1}^{T}\eta_t\B_{t}L_C^{\alpha}\prod_{j=t+1}^{T}(I-\eta_{j}L_C)\right\|_{\mathrm{HS}}^{2}\right]
			\\&\ +
			2\be_{z^{T}}\left[\left\langle-S^{\dagger}L_C^{\alpha}\prod_{t=1}^{T}(I-\eta_{t}L_C),\sum_{t=1}^{T}\eta_t\B_{t}L_C^{\alpha}\prod_{j=t+1}^{T}(I-\eta_{j}L_C)\right\rangle_{\mathrm{HS}}   \right].
		\end{align*}
		Using the vanishing property $\mathbb{E}_{z_t}[\B_t]=0$, we get 
		\begin{align*}
			&\be_{z^{T}}\left[\left\langle-S^{\dagger}L_C^{\alpha}\prod_{t=1}^{T}(I-\eta_{t}L_C),\sum_{t=1}^{T}\eta_t\B_{t}L_C^{\alpha}\prod_{j=t+1}^{T}(I-\eta_{j}L_C)\right\rangle_{\mathrm{HS}}   \right]
			\\=&\left\langle-S^{\dagger}L_C^{\alpha}\prod_{t=1}^{T}(I-\eta_{t}L_C),\sum_{t=1}^{T}\eta_t\be_{z^{T}}[\B_{t}]L_C^{\alpha}\prod_{j=t+1}^{T}(I-\eta_{j}L_C)\right\rangle_{\mathrm{HS}}=0.
		\end{align*}
		Therefore,
		\begin{align} 
			&\be_{z^{T}}\left[\left\|(S_{T+1}-S^{\dagger})L_C^{\alpha}\right\|_{\mathrm{HS}}^{2}\right]
			=\left\|S^{\dagger}L_C^{\alpha}\prod_{t=1}^{T}(I-\eta_{t}L_C)\right\|_{\mathrm{HS}}^{2} \nonumber
			\\&+\be_{z^{T}}\left[\left\|\sum_{t=1}^{T}\eta_t\B_{t}L_C^{\alpha}\prod_{j=t+1}^{T}(I-\eta_{j}L_C)\right\|_{\mathrm{HS}}^{2}\right]. \label{eq1}
		\end{align}	
		Denote by $\A$ the second term on the right hand side of (\ref{eq1}). Next we bound $\A$,
		which can be rewritten as
		\[
		\sum_{t=1}^{T}\sum_{t^\prime=1}^{T}\be_{z^{T}}\left[\left\langle\eta_t\B_{t}L_C^{\alpha}\prod_{j=t+1}^{T}(I-\eta_{j}L_C),\eta_{t^\prime}\B_{t^\prime}L_C^{\alpha}\prod_{j^\prime=t^\prime+1}^{T}(I-\eta_{j^\prime}L_C)\right\rangle_{\mathrm{HS}}    \right].
		\]
		By the vanishing property of $\B_t$, for $t>t^\prime$, there holds
		\begin{align*}
			&\be_{z^{T}}\left[\left\langle\B_{t}L_C^{\alpha}\prod_{j=t+1}^{T}(I-\eta_{j}L_C),\B_{t^\prime}L_C^{\alpha}\prod_{j^\prime=t^\prime+1}^{T}(I-\eta_{j^\prime}L_C)\right\rangle_{\mathrm{HS}}\right]
			\\ &=\be_{z^{t-1}}\be_{z_{t}}\left[\left\langle\B_{t}L_C^{\alpha}\prod_{j=t+1}^{T}(I-\eta_{j}L_C),\B_{t^\prime}L_C^{\alpha}\prod_{j^\prime=t^\prime+1}^{T}(I-\eta_{j^\prime}L_C)\right\rangle_{\mathrm{HS}}\right]
			\\ &=\be_{z^{t-1}}\left[\left\langle\be_{z_{t}}\B_{t}L_C^{\alpha}\prod_{j=t+1}^{T}(I-\eta_{j}L_C),\B_{t^\prime}L_C^{\alpha}\prod_{j^\prime=t^\prime+1}^{T}(I-\eta_{j^\prime}L_C)\right\rangle_{\mathrm{HS}}\right]=0.
		\end{align*}
		By the symmetry of $t,t^\prime$, the above equality also holds if $t<t^\prime$.
		Consequently, it follows that
		\begin{eqnarray}
			&&\A=\sum_{t=1}^{T}\be_{z^{T}}\left[\left\|\eta_t\B_{t}L_C^{\alpha}\prod_{j=t+1}^{T}(I-\eta_{j}L_C)\right\|_{\mathrm{HS}}^{2}\right]. \label{temp1}
		\end{eqnarray}
		Recall the definition of $\B_t$ in (\ref{lemma 1.2}) and use the vanishing property $\mathbb{E}_{z_t}[\B_t]=0$ again, we have
		$
		(S_t-S^\dagger)L_C=-\mathbb{E}_{z_t}[(y_t-S_tx_t)\otimes x_t]
		$ and
		$\B_{t}=-\mathbb{E}_{z_t}[(y_t-S_tx_t)\otimes x_t]+(y_t-S_tx_t)\otimes x_t$. Denote $\eta_t\left[(y_t-S_tx_t)\otimes x_t\right]L_C^{\alpha}\prod_{j=t+1}^{T}(I-\eta_{j}L_C)$ by $\B$ for simplicity, then substituting this into (\ref{temp1}) yields that
		\begin{align*}
			\A&=\sum_{t=1}^{T}\be_{z^{T}}\left[\left\|-\mathbb{E}_{z_t}[\B]+\B\right\|_{\mathrm{HS}}^{2}\right]=\sum_{t=1}^{T}\be_{z^{t-1}}\be_{z_{t}}\left[\left\|-\mathbb{E}_{z_t}[\B]+\B\right\|_{\mathrm{HS}}^{2}\right]
			\\&\leq \sum_{t=1}^{T}\be_{z^{T}}\left[\left\|\B\right\|_{\mathrm{HS}}^{2}\right]
			=\sum_{t=1}^{T}\be_{z^{T}}\left[\left\|\left[\eta_{t}(y_{t}-S_{t}x_{t})\otimes{}x_{t}\right]L_C^{\alpha}\prod_{j=t+1}^{T}(I-\eta_{j}L_C)\right\|_{\mathrm{HS}}^{2}\right].
		\end{align*}
		Recall that $\{e_i\}_{i\geq1}$ is an orthonormal basis of Hilbert space $\H_1$, by the definition of the Hilbert-schmidt norm, we have
  \begin{equation}
      \begin{aligned}
\A&\leq\sum_{t=1}^{T}\be_{z^{T}}\left[\sum_{i\geq1}\left\|\left(\eta_{t}(y_{t}-S_{t}x_{t})\otimes{}x_{t}\right)L_C^{\alpha}\prod_{j=t+1}^{T}(I-\eta_{j}L_C)e_i\right\|_{\H_{2}}^{2}\right]
			\\ &=\sum_{t=1}^{T}\be_{z^{T}}\left[\sum_{i\geq1}\left\|\eta_{t}(y_{t}-S_{t}x_{t})\right\|_{\H_{2}}^2\left\langle x_t,L_C^{\alpha}\prod_{j=t+1}^{T}(I-\eta_{j}L_C)e_i\right\rangle_{\H_{1}}^2\right]\\ &=\sum_{t=1}^{T}\eta_{t}^{2}\be_{z^{T}}\left[\left\|S_{t}x_{t}-y_{t}\right\|_{\H_{2}}^{2}\left\|L_C^{\alpha}\prod_{j=t+1}^{T}(I-\eta_{j}L_C)x_{t}\right\|_{\H_{1}}^{2}\right] 
			\\&=\sum_{t=1}^{T}\eta_{t}^{2}\be_{z^{t-1}}\be_{x_t}\left[\left(\left\|(S_{t}-S^\dagger)x_{t}\right\|^2_{\H_{2}}+\sigma^2\right)\left\|L_C^{\alpha}\prod_{j=t+1}^{T}(I-\eta_{j}L_C)x_{t}\right\|_{\H_{1}}^{2}\right], \label{temp2}
		\end{aligned}
  \end{equation}where the last equality we have taken expectation over $\epsilon_{t}$ and used
		\[
		\be_{\epsilon_{t}}[\left\|S_{t}x_{t}-y_{t}\right\|_{\H_{2}}^{2}]=\left\|(S_{t}-S^\dagger)x_{t}\right\|^2_{\H_{2}}+\sigma^2
		\]
		and the fact that $\epsilon_{t}$ is independent of $x_t$.
		Furthermore, By applying Cauchy-Schwarz inequality to the above inequality (\ref{temp2}), it follows that $\A$ is bounded by
		\begin{align*}
			&\sum_{t=1}^{T}\eta_{t}^{2}\left(\be_{z^{t-1}}\sqrt{\be_{x_t}\left\|(S_{t}-S^\dagger)x_{t}\right\|_{\H_{2}}^4}+\sigma^2\right)
			\left(\be_{x_t}\left\|L_C^{\alpha}\prod_{j=t+1}^{T}(I-\eta_{j}L_C)x_{t}\right\|_{\H_{1}}^{4}\right)^{1/2}
			\\\leq&\sum_{t=1}^{T}\eta_{t}^{2}\sqrt{c}\left(\sqrt{c}\be_{z^{t-1}}\be_{x_t}\left\|(S_{t}-S^\dagger)x_{t}\right\|_{\H_{2}}^2+\sigma^2\right)
			\be_{x_t}\left\|L_C^{\alpha}\prod_{j=t+1}^{T}(I-\eta_{j}L_C)x_{t}\right\|_{\H_{1}}^{2},
		\end{align*}
        where the last inequality follows from Assumption~\ref{a3} applied to
the bounded operators $S_t-S^{\dagger}$ and
$L_C^\alpha\prod_{j=t+1}^{T}(I-\eta_jL_C)$.
For the former, the moment bound is applied conditionally on the
first $t-1$ samples, using the independence of $x_t$ from these samples.
        Recall that for any self-adjoint linear operator $A$,  $\be_{x_t}\left\|A x_{t}\right\|_{\H_{1}}^{2}=\be_{x_t}\left[\mathrm{Tr}(A x_t\otimes x_tA)\right]=\mathrm{Tr}(A L_CA)$. Therefore, by setting $A=L_C^{\alpha}\prod_{j=t+1}^{T}(I-\eta_{j}L_C)$ and recalling the equivalent expression of the prediction error (\ref{form1}), we have
		\begin{equation} \label{eq11}
			\A\leq\sum_{t=1}^{T}\eta_{t}^{2}\sqrt{c}\left(\sqrt{c}\be_{z^{t-1}}\left[\mathcal{E}(S_{t})-\mathcal{E}(S^{\dagger})\right]+\sigma^2\right)\mathrm{Tr}\left(L_C^{1+2\alpha}\prod_{j=t+1}^{T}(I-\eta_{j}L_C)^{2}\right).
		\end{equation}
		Finally, the proof is completed by combining (\ref{eq1}) and (\ref{eq11}).
	\end{proof}

	\subsection{Proof of Lemma \ref{lemma basic}}\label{Lemma 4.3}
	\begin{proof}
        The case $A=0$ is immediate; assume henceforth that $A\ne0$. 
		Recall the definition of the operator norm and $\eta_{j}\|A\|<1$ for all $l\leq j\leq T$, we have
		\[
		\left\|A^\beta\prod_{j=l}^{T}(I-\eta_jA)^2\right\|
		\leq
		\sup_{0\leq u\leq\|A\|}\left\{u^\beta\prod_{j=l}^{T}(I-\eta_ju)^2\right\}.
		\]
		Since $1-\eta_ju\leq\exp\{-\eta_ju\}$ and the function $h(u)=u^\beta\exp\{-uv\}$ achieves its maximum at $u=\beta/v$, it follows that
		\begin{equation*}
			\begin{aligned}
				\left\|A^\beta\prod_{j=l}^{T}(I-\eta_jA)^2\right\|
				\leq
				\sup_{0\leq u\leq\|A\|}\left\{u^\beta\exp\left(-2\sum_{j=l}^{T}\eta_{j}u\right)\right\}
				\leq
				\left(\frac{\beta}{2e}\right)^\beta\left(\sum_{j=l}^{T}\eta_{j}\right)^{-\beta}.
			\end{aligned}
		\end{equation*}
		On the other hand, it is obvious that $\left\|A^\beta\prod_{j=l}^{T}(I-\eta_jA)^2\right\|
		\leq\|A\|^\beta$. Using the inequality $\min\{a,b\}\leq\frac{2ab}{a+b}$ for any $a>0$ and $b>0$ yields that
		\begin{equation*}
			\begin{aligned}
				\left\|A^\beta\prod_{j=l}^{T}(I-\eta_jA)^2\right\|
				&\leq\min\left\{\left(\frac{\beta}{2e}\right)^\beta\left(\sum_{j=l}^{T}\eta_{j}\right)^{-\beta},\|A\|^\beta\right\}
				\\&\leq\frac{2\|A\|^\beta\left(\frac{\beta}{2e}\right)^\beta\left(\sum_{j=l}^{T}\eta_{j}\right)^{-\beta}}{\|A\|^\beta+\left(\frac{\beta}{2e}\right)^\beta\left(\sum_{j=l}^{T}\eta_{j}\right)^{-\beta}}
				\\&
				\leq\frac{2}{\left(\frac{\beta}{2e}\right)^{-\beta}\left(\sum_{j=l}^{T}\eta_{j}\right)^{\beta}+\|A\|^{-\beta}}
				\\&\leq
				2\frac{(\frac{\beta}{2e})^\beta+\|A\|^\beta}{1+(\sum_{j=l}^{T}\eta_j)^\beta}.
			\end{aligned}
		\end{equation*}
		This completes the proof. 
	\end{proof}

\subsection{Proof of Proposition \ref{first term}}
\label{Prop 4.4}
\begin{proof}
		
		Since $L_C$ is a self-adjoint operator, by the properties of the trace and the Hilbert-Schmidt norm, we see that
		\begin{align}
			\left\|L_C^{\alpha}\prod_{t=1}^{T}(I-\eta_{t}L_C)\right\|^2_{\mathrm{HS}}
			&=\mathrm{Tr}\left(L_C^{2\alpha}\prod_{t=1}^{T}(I-\eta_tL_C)^2\right) \nonumber
			\\\ &\leq\mathrm{Tr}(L_C^s)\left\|L_C^{2\alpha-s}\prod_{t=1}^{T}(I-\eta_tL_C)^2\right\|, \nonumber
		\end{align}
		where the last inequality follows from Assumption \ref{a2}. Then, by applying Lemma \ref{lemma basic} with $A=L_C$, $l=1$ and $\beta=2\alpha-s$ entails that 
		\begin{equation}
			\left\|L_C^{\alpha}\prod_{t=1}^{T}(I-\eta_{t}L_C)\right\|^2_{\mathrm{HS}}
			\leq \mathrm{Tr}(L_C^s)\left(\frac{2\alpha-s}{2e}\right)^{2\alpha-s}\left(\sum_{t=1}^{T}\eta_t\right)^{-(2\alpha-s)}. \label{temp3}
		\end{equation}
		It remains to bound $\left(\sum_{t=1}^{T}\eta_t\right)^{-(2\alpha-s)}$. To this end, for any $T\geq2$, we have the following estimation,
		\begin{align}
			\sum_{t=1}^{T}\eta_t
			=\eta_{1}\sum_{t=1}^{T}t^{-\theta}
			\geq\eta_{1}\int_{T/2}^{T}u^{-\theta}\mathrm{d}u=\frac{\eta_1(1-2^{\theta-1})}{1-\theta}T^{1-\theta}. \label{temp4}
		\end{align}
		By $1-\theta\geq1-2^{\theta-1}$ for $0\leq\theta<1$, the inequality (\ref{temp4}) also holds when $T=1$. We end the proof by substituting (\ref{temp4}) into (\ref{temp3}),
		\begin{eqnarray*}
			\left\|L_C^{\alpha}\prod_{t=1}^{T}(I-\eta_{t}L_C)\right\|^2_{\mathrm{HS}}
			&\leq&\mathrm{Tr}(L_C^s)\left(\frac{2\alpha-s}{2e}\right)^{2\alpha-s}\left(\frac{\eta_1(1-2^{\theta-1})}{1-\theta}T^{1-\theta}\right)^{-(2\alpha-s)}
			\\&=&\mathrm{Tr}(L_C^s)\left(\frac{(2\alpha-s)(1-\theta)}{2e(1-2^{\theta-1})\eta_{1}}\right)^{2\alpha-s}T^{-(1-\theta)(2\alpha-s)}.
		\end{eqnarray*} The proof is finished.
	\end{proof}

\subsection{Proof of Proposition \ref{eta1}} \label{Prop 4.5}
We first prove the following lemma.
 
	\begin{lemma} \label{re1}
		Let $0<\theta<1$, $v>0$, define 
		\[
		\A_1=\int_{1}^{t/2}\frac{u^{-2\theta}}{1+\left(t^{1-\theta}-u^{1-\theta}\right)^v}du
		\]
		and
		\[
		\A_2=\int_{t/2}^{t}\frac{u^{-2\theta}}{1+\left(t^{1-\theta}-u^{1-\theta}\right)^v}du.
		\]
		Then, there exist constants $c_1, c_2>0$, such that for any $t\geq3$,
		\begin{equation*}
			\A_1\leq c_1
			\begin{cases}
				t^{1-v-\theta(2-v)}, & \text{if } 0<\theta<\frac{1}{2}, \\
				t^{-v/2}\log t, &  \mbox{if } \theta=\frac{1}{2}, \\
				t^{-v(1-\theta)}, &  \mbox{if } \frac{1}{2}<\theta<1,
			\end{cases}
		\end{equation*}
		and 
		\begin{equation*}
			\A_2\leq c_2
			\begin{cases}
				t^{-\theta}, & \mbox{ if }v>1, \\
				t^{-\theta}\log t, & \mbox{ if }v=1, \\
				t^{1-v-\theta(2-v)}, & \mbox{ if }0<v<1,
			\end{cases}
		\end{equation*}
		where $c_1=c_1(\theta,v)$, $c_2=c_2(\theta,v)$ are independent of $t$. 
	\end{lemma}
	
	\begin{proof}
		{\textbf Bound $\A_1$:}
		Since $u\leq t/2$, we have
		\begin{equation*}
			\A_1\leq\frac{1}{1+\left(t^{1-\theta}-(t/2)^{1-\theta}\right)^v}\int_{1}^{t/2}u^{-2\theta}du.
		\end{equation*}
		Note that 
		\[
		\int_{1}^{t/2}u^{-2\theta}du\leq
		\begin{cases}
			\frac{(t/2)^{1-2\theta}}{1-2\theta}, & \text{if }0<\theta<\frac{1}{2}, \\
			\log t, &\text{if } \theta=\frac{1}{2}, \\
			\frac{1}{2\theta-1}, &\text{if } \frac{1}{2}<\theta<1.
		\end{cases}
		\]
		Therefore,
		\begin{align*}
			\A_1&\leq \frac{t^{-v(1-\theta)}}{(1-2^{\theta-1})^v}
			\begin{cases}
				\frac{(t/2)^{1-2\theta}}{1-2\theta}, & \text{if }0<\theta<\frac{1}{2}, \\
				\log t, &\text{if } \theta=\frac{1}{2}, \\
				\frac{1}{2\theta-1}, &\text{if } \frac{1}{2}<\theta<1,
			\end{cases} \\
			& \leq c_1
			\begin{cases}
				t^{1-v-\theta(2-v)}, & \text{if }0<\theta<\frac{1}{2}, \\
				t^{-v/2}\log t, & \text{if }\theta=\frac{1}{2}, \\
				t^{-v(1-\theta)}, & \text{if } \frac{1}{2}<\theta<1,
			\end{cases}
		\end{align*}
		where $c_1=(1-2^{\theta-1})^{-v}$ when $\theta=\frac{1}{2}$, and $c_1=(1-2^{\theta-1})^{-v}/\vert2\theta-1\vert$ otherwise. 
  
		\textbf{Bound $\A_2$:} 
		We change the variable as $\xi=t^{1-\theta}-u^{1-\theta}$, thus we have $d\xi=-(1-\theta)u^{-\theta}du$, and
  \begin{equation} \label{integral}
  \begin{aligned}
			\A_2& =\int_{0}^{t^{1-\theta}-(t/2)^{1-\theta}}\frac{u^{-\theta}}{(1+\xi^v)(1-\theta)}d\xi
			\leq\frac{(t/2)^{-\theta}}{1-\theta}\int_{0}^{t^{1-\theta}(1-2^{\theta-1})}\frac{1}{1+\xi^v}d\xi.
		\end{aligned}
  \end{equation}

        By estimating the integral (\ref{integral}), we obtain
\begin{equation*}
\mathcal A_2
\le
\frac{(t/2)^{-\theta}}{1-\theta}
\begin{cases}
\dfrac{v}{v-1}, & v>1,\\
\log\bigl(1+(1-2^{\theta-1})t^{1-\theta}\bigr), & v=1,\\
\dfrac{(1-2^{\theta-1})^{1-v}}{1-v}
t^{(1-\theta)(1-v)}, & 0<v<1.
\end{cases}
\end{equation*}
For $v=1$, since $t\ge3$,
\begin{equation*}
\begin{aligned}
\log\bigl(1+(1-2^{\theta-1})t^{1-\theta}\bigr)
&\le
\log(2-2^{\theta-1})+(1-\theta)\log t\\
&\le
\bigl[1-\theta+\log(2-2^{\theta-1})\bigr]\log t.
\end{aligned}
\end{equation*}
Consequently, the asserted bound holds with
\begin{equation*}
c_2(\theta,v)
=
\frac{2^\theta}{1-\theta}
\begin{cases}
\dfrac{v}{v-1}, & v>1,\\
1-\theta+\log(2-2^{\theta-1}), & v=1,\\
\dfrac{(1-2^{\theta-1})^{1-v}}{1-v}, & 0<v<1.
\end{cases}
\end{equation*}


  We complete the proof. \end{proof}

	\begin{proof}[Proof of Proposition \ref{eta1}]
		One can easily verify that
		\begin{gather*}
			\sum_{j=t+1}^{T}\eta_{j}=\eta_{1}\sum_{j=t+1}^{T}j^{-\theta}\geq\eta_{1}\int_{t+1}^{T+1}u^{-\theta}du=\frac{\eta_{1}}{1-\theta}\left((T+1)^{1-\theta}-(t+1)^{1-\theta}\right).
		\end{gather*}
		Therefore,
  \begin{equation}
		\begin{aligned}
  			\sum_{t=1}^{T-1}\frac{\eta_{t}^2}{1+\left(\sum_{j=t+1}^{T}\eta_{j}\right)^v}
			&\leq\eta_{1}^2\sum_{t=1}^{T-1}\frac{t^{-2\theta}}{1+(\frac{\eta_{1}}{1-\theta})^v\left[(T+1)^{1-\theta}-(t+1)^{1-\theta}\right]^v} \\
			&\leq \frac{\eta_{1}^2}{\min\{1,(\frac{\eta_{1}}{1-\theta})^v\}}\sum_{t=1}^{T-1}\frac{t^{-2\theta}}{1+\left[(T+1)^{1-\theta}-(t+1)^{1-\theta}\right]^v}. \label{l} 
		\end{aligned}
\end{equation}
		Recall that for any $t\geq1$, $t^{-2\theta}\leq3^{2\theta}(t+2)^{-2\theta}$, substituting this into (\ref{l}) entails that
		\begin{align*}
			\sum_{t=1}^{T-1}\frac{\eta_{t}^2}{1+\left(\sum_{j=t+1}^{T}\eta_{j}\right)^v}
			\leq\frac{\eta_{1}^23^{2\theta}}{\min\{1,(\frac{\eta_{1}}{1-\theta})^v\}}\sum_{t=1}^{T-1}\frac{(t+2)^{-2\theta}}{1+\left[(T+1)^{1-\theta}-(t+1)^{1-\theta}\right]^v}.
		\end{align*}	
		Note that for any $t\geq1$, there holds \[\frac{(t+2)^{-2\theta}}{1+\left[(T+1)^{1-\theta}-(t+1)^{1-\theta}\right]^v}\leq
		\int_{t+1}^{t+2}\frac{u^{-2\theta}}{1+\left[(T+1)^{1-\theta}-u^{1-\theta}\right]^v}du.\]
		Hence,
		\begin{align*}
			\sum_{t=1}^{T-1}\frac{\eta_{t}^2}{1+\left(\sum_{j=t+1}^{T}\eta_{j}\right)^v}
			\leq\frac{\eta_{1}^23^{2\theta}}{\min\{1,(\frac{\eta_{1}}{1-\theta})^v\}}\int_{2}^{T+1}\frac{u^{-2\theta}}{1+\left[(T+1)^{1-\theta}-u^{1-\theta}\right]^v}du.
		\end{align*}	
		Define $\A=\int_{2}^{T+1}\frac{u^{-2\theta}}{1+\left[(T+1)^{1-\theta}-u^{1-\theta}\right]^v}du$. If we choose $t=T+1\geq3$ in $\A_1$ and $\A_2$ in Lemma \ref{re1}, then $\A\leq\A_1+\A_2$. Thus, by Lemma \ref{re1}, we have $\A$ is bounded by
		\begin{align*}
			c_1
			\begin{cases}
				(T+1)^{1-v-\theta(2-v)}, &\text{if } 0<\theta<\frac{1}{2}, \\
				(T+1)^{-v/2}\log (T+1), &\text{if } \theta=\frac{1}{2}, \\
				(T+1)^{-v(1-\theta)}, &\text{if } \frac{1}{2}<\theta<1,
			\end{cases}
			\ +\ 
			c_2
			\begin{cases}
				(T+1)^{-\theta}, &\text{if } v>1, \\
				(T+1)^{-\theta}\log (T+1), &\text{if } v=1, \\
				(T+1)^{1-v-\theta(2-v)}, &\text{if } 0<v<1.
			\end{cases}
		\end{align*}
  
		\textbf{Case 1:} If $0<v<1$.		
		Note that when $\theta=1/2$, $-v/2=1-v-\theta(2-v)$, and when $1/2<\theta<1$, $-v(1-\theta)>1-v-\theta(2-v)$. So we have
		\[
		\A\leq(c_1+c_2)
		\begin{cases}
			(T+1)^{1-v-\theta(2-v)}, &\text{if } 0<\theta<\frac{1}{2}, \\
			(T+1)^{-v/2}\log (T+1), &\text{if } \theta=\frac{1}{2}, \\
			(T+1)^{-v(1-\theta)}, &\text{if } \frac{1}{2}<\theta<1.
		\end{cases}
		\]
  
		\textbf{Case 2:} If $v>1$.
		Note that when $0<\theta<1/2$, $1-v-\theta(2-v)<-\theta$ and $v(1-\theta)>\theta$, which implies that $\A\leq(c_1+c_2)(T+1)^{-\min\{\theta,v(1-\theta)\}}$ when $\theta\not=1/2$.	
		If $\theta=1/2$, recall that
		\[
		\frac{(T+1)^{-v/2}\log (T+1)}{(T+1)^{-1/2}}
		\leq \max_{x>0} \frac{x^{-v/2}\log x}{x^{-1/2}}=\frac{2}{e(v-1)},
		\]
		where the maximum is achieved at $x=e^{2/(v-1)}$. So $\A\leq
		(\frac{2}{e(v-1)}c_1+c_2)(T+1)^{-1/2}$.
		Consequently,
		\[
		\A\leq c_3(T+1)^{-\min\{\theta,v(1-\theta)\}},
		\]
		where $c_3=c_1+c_2$ when $\theta\not=1/2$ and $c_3=\frac{2}{e(v-1)}c_1+c_2$ when $\theta=1/2$. 
  
		\textbf{Case 3:} If $v=1$.
		It is obvious that $1-v-\theta(2-v)=-\theta$ when $0<\theta<1/2$ and $-v/2=-\theta$ when $\theta=1/2$. And we can easily verify that $\max_{x>0} \frac{x^{-\theta}\log x}{x^{-(1-\theta)}}=\frac{1}{e(2\theta-1)},$ where the maximum is achieved at $x=e^{\frac{1}{2\theta-1}}$.
		So we have
		\[
		\A\leq c_4
		\begin{cases}
			(T+1)^{-\theta}\log (T+1) , &\text{if } 0<\theta\leq\frac{1}{2}, \\
			(T+1)^{-(1-\theta)} , & \text{if }\frac{1}{2}<\theta<1,
		\end{cases}
		\] 
		where $c_4=c_1+\frac{1}{e(2\theta-1)}c_2$ when $\frac12<\theta<1$ and $c_4=c_1+c_2$ otherwise.
        We complete the proof by taking $c_5(\eta_1,v,\theta)$
to be $\frac{\eta_{1}^23^{2\theta}}{\min\{1,(\frac{\eta_{1}}{1-\theta})^v\}}$
times $c_1+c_2$, $c_3$, and $c_4$ in Cases 1, 2, and 3,
respectively.
	\end{proof}

\subsection{Proof of Proposition \ref{p1}} \label{Prop 4.6}
\begin{proof}
		For brevity, denote $3c_5(\eta_{1},1,\theta)\max\{\frac{1}{e\theta},1\}+\eta_{1}^2$ by $\xi$.
		We choose 
		\[M=\frac{\|S^\dagger\|^2+\mathcal{E}(S^\dagger)+\left|\sqrt{c}\sigma^2-c\mathcal{E}(S^{\dagger})\right|\xi}{1-c\xi}.\]
		Since $S_1={\bf 0}$, we know from (\ref{form2}) and the fact that $\|L_C^{1/2}\|_{\mathrm{HS}}^2=\mathrm{Tr}(L_C)\leq1$ that
		\begin{align*}
			\mathcal{E}(S_{1})&=\mathcal{E}(S^\dagger)+\|(S_1-S^\dagger)L_C^{\frac12}\|^2_{\mathrm{HS}}=\mathcal{E}(S^\dagger)+\|S^\dagger L_C^{\frac12}\|^2_{\mathrm{HS}}
			\\&\leq\mathcal{E}(S^\dagger)+\|S^\dagger\|^2\leq M,
		\end{align*}
		which implies (\ref{bound1}) holds for $t=1$. We prove (\ref{bound1}) by induction. Assume this holds for $1\leq t\leq k$. For $t=k+1$, by the error decomposition (\ref{eq2}) in Proposition \ref{Proposition error 1}, there holds
		\begin{equation}
			\begin{aligned}
				&\be_{z^{k}}\left[\mathcal{E}(S_{k+1})\right]=\be_{z^{k}}\left[\mathcal{E}(S_{k+1})-\mathcal{E}(S^\dagger)\right]+\mathcal{E}(S^\dagger)
				\\ \leq&
				\left\|S^{\dagger}L_C^{\frac{1}{2}}\prod_{t=1}^{k}(I-\eta_{t}L_C)\right\|_{\mathrm{HS}}^{2} 
				+\sum_{t=1}^{k}\eta_{t}^{2}\sqrt{c}\left(\sqrt{c}\be_{z^{t-1}}\left[\mathcal{E}(S_{t})-\mathcal{E}(S^{\dagger})\right]
				+\sigma^2\right) 
				\\ &\times\mathrm{Tr}\left(L_C^2\prod_{j=t+1}^{k}(I-\eta_{j}L_C)^2\right)
				+\mathcal{E}(S^\dagger). \label{temp9}
			\end{aligned}
		\end{equation}
		Denote the first term and the second term on the right-hand side of the inequality (\ref{temp9}) by $\A_1$ and $\A_2$, respectively. Next, we bound them separately. 
  
		\textbf{Bound $\A_1$:} Since $\mathrm{Tr}(L_C)\leq1$, it follows that
		\begin{equation}	\label{temp7}
			\begin{aligned}
				\A_1&\leq\|S^\dagger\|^2\left\|L_C^{\frac{1}{2}}\prod_{t=1}^{k}(I-\eta_{t}L_C)\right\|_{\mathrm{HS}}^{2}
				=\|S^\dagger\|^2\mathrm{Tr}\left(L_C\prod_{t=1}^{k}(I-\eta_{t}L_C)^2\right) 
				\\ &\leq\|S^\dagger\|^2\left\|\prod_{t=1}^{k}(I-\eta_{t}L_C)\right\|^2\mathrm{Tr}(L_C)\leq\|S^\dagger\|^2. 
			\end{aligned}
		\end{equation}

		\textbf{Bound $\A_2$:} Note that 
            \begin{equation}
            \begin{aligned}
			\mathrm{Tr}\left(L_C^2\prod_{j=t+1}^{k}(I-\eta_{j}L_C)^2\right)&\leq
			\mathrm{Tr}(L_C)\left\|L_C\prod_{j=t+1}^{k}(I-\eta_{j}L_C)^2\right\| 
			\\
			&\leq\left(\frac{1}{e}+2\right)\frac{1}{1+\sum_{j=t+1}^{k}\eta_{j}}, \label{temp22}
		\end{aligned}
            \end{equation}
		where the last inequality (\ref{temp22}) we have used the facts $\mathrm{Tr}(L_C)\leq1$, $\|L_C\|\leq1$ and Lemma \ref{lemma basic} with $l=t+1,T=k,\beta=1$. Then, by applying the induction assumption, $\A_2$ is bounded by	
		\begin{gather*}
			\sqrt{c}\left(\sqrt{c}\left(M-\mathcal{E}(S^{\dagger})\right)	+\sigma^2\right)\left[\left(\frac{1}{e}+2\right)\sum_{t=1}^{k-1}\frac{\eta_{t}^2}{1+\sum_{j=t+1}^{k}\eta_{j}}+\eta_k^2\right].
		\end{gather*}
		By Proposition \ref{eta1} with $v=1$ and the inequality (\ref{temp6}), $\A_2$ is bounded by
		\begin{gather}
			\left(cM+\sqrt{c}\sigma^2-c\mathcal{E}(S^{\dagger})\right)\left[3c_5(\eta_{1},1,\theta)\max\{\frac{1}{e\theta},1\}+\eta_{1}^2\right]. \label{temp8}
		\end{gather}
		Putting (\ref{temp7}) and (\ref{temp8}) into (\ref{temp9}) tells us that $	\be_{z^{k}}\left[\mathcal{E}(S_{k+1})\right]$ can be bounded by
		\begin{align*}
			\|S^\dagger\|^2+\mathcal{E}(S^\dagger)+\left(cM+\left|\sqrt{c}\sigma^2-c\mathcal{E}(S^{\dagger})\right|\right)\xi\leq M,
		\end{align*}
		which advances the induction and completes the proof.	
	\end{proof}
 
\subsection{Proof of Proposition \ref{eta2}} \label{Appendix D}

\begin{proof}
Since $u\mapsto(1+u^v)^{-1}$ is decreasing on $[0,\infty)$,
we have
\begin{equation*}
\begin{aligned}
\sum_{t=1}^{T-1}
\frac{\eta_t^2}
{1+\left(\sum_{j=t+1}^{T}\eta_j\right)^v}
&=\eta_1^2\sum_{k=1}^{T-1}\frac{1}{1+(k\eta_1)^v}\\
&\le
\eta_1\int_0^{\eta_1(T-1)}\frac{du}{1+u^v}.
\end{aligned}
\end{equation*}
For any $a\ge0$,
\begin{equation*}
\int_0^a\frac{du}{1+u^v}
\le
\begin{cases}
\dfrac{a^{1-v}}{1-v}, & 0<v<1,\\
\log(1+a), & v=1,\\
\dfrac{v}{v-1}, & v>1.
\end{cases}
\end{equation*}
Indeed, for $0<v<1$, we use $(1+u^v)^{-1}\le u^{-v}$
for $u>0$. The case $v=1$ follows by direct integration.
For $v>1$, we use
\begin{equation*}
\int_0^a\frac{du}{1+u^v}
\le
\int_0^\infty\frac{du}{1+u^v}
\le
\int_0^1du+\int_1^\infty u^{-v}\,du
=\frac{v}{v-1}.
\end{equation*}
Applying these bounds with $a=\eta_1(T-1)$ and using
$T-1\le T+1$ proves the claim with the stated constant $c_6(v)$.
\end{proof}

 
\subsection{Proof of Proposition \ref{p2}} \label{Prop 4.8}
\begin{proof}
		We choose 
		\[\widetilde{M}=2\left(\|S^\dagger\|^2+\mathcal{E}(S^\dagger)\right)+\frac{\left|\sigma^2-\sqrt{c}\mathcal{E}(S^\dagger)\right|}{\sqrt{c}}.\]
		When $t=1$, since $S_1=0$, we know from (\ref{form2}) and $\|L_C^{1/2}\|_{\mathrm{HS}}^2=\mathrm{Tr}(L_C)\leq1$ that
		\begin{align*}
			\mathcal{E}(S_{1})&=\mathcal{E}(S^\dagger)+\|(S_1-S^\dagger)L_C^{\frac12}\|^2_{\mathrm{HS}}=\mathcal{E}(S^\dagger)+\|S^\dagger L_C^{\frac12}\|^2_{\mathrm{HS}}
			\\&\leq\mathcal{E}(S^\dagger)+\|S^\dagger\|^2\leq\widetilde{M}.
		\end{align*}
        Assume that the claimed bound holds for $1\le t\le k$,
where $1\le k\le T$. For $t=k+1$, Proposition~\ref{Proposition error 1} gives
		\begin{equation}
			\begin{aligned}
			&\be_{z^{k}}\left[\mathcal{E}(S_{k+1})\right]=\be_{z^{k}}\left[\mathcal{E}(S_{k+1})-\mathcal{E}(S^\dagger)\right]+\mathcal{E}(S^\dagger) 
				\\ \leq&
				\left\|S^{\dagger}L_C^{\frac{1}{2}}(I-\eta_{1}L_C)^k\right\|_{\mathrm{HS}}^{2} 
				+\sum_{t=1}^{k}\eta_{t}^{2}\sqrt{c}\left(\sqrt{c}\be_{z^{t-1}}\left[\mathcal{E}(S_{t})-\mathcal{E}(S^{\dagger})\right]
				+\sigma^2\right)
				\\ &\times\mathrm{Tr}\left(L_C^2(I-\eta_{1}L_C)^{2(k-t)}\right)
				+\mathcal{E}(S^\dagger). \label{temp10}
			\end{aligned}
		\end{equation}
		Denote the first term and the second term on the right-hand side of the inequality (\ref{temp10}) by $\B_1$ and $\B_2$, respectively.
		We bound $\B_1$ as before,
		\begin{equation}
			\B_1\leq\|S^\dagger\|^2\left\|L_C^{\frac{1}{2}}(I-\eta_{1}L_C)^k\right\|_{\mathrm{HS}}^{2}\leq\|S^\dagger\|^2\left\|(I-\eta_{1}L_C)^k\right\|^2\mathrm{Tr}(L_C)\leq\|S^\dagger\|^2. \label{temp11}
		\end{equation}
		Note that 
		\begin{align}
			\mathrm{Tr}\left(L_C^2(I-\eta_{1}L_C)^{2(k-t)}\right)&\leq
			\mathrm{Tr}(L_C)\left\|L_C(I-\eta_{1}L_C)^{2(k-t)}\right\| \nonumber
			\\
			&\leq\left(\frac{1}{e}+2\right)\frac{1}{1+(k-t)\eta_{1}}, \label{temp23}
		\end{align}
		where the last inequality (\ref{temp23}) we have used $\mathrm{Tr}(L_C)\leq1$, $\|L_C\|\leq1$ and Lemma \ref{lemma basic} with $l=t+1,T=k$ and $\beta=1$. Then, by applying the induction assumption, $\B_2$ is bounded by	
		\begin{gather*}
			\sqrt{c}\left(\sqrt{c}\left(\widetilde{M}-\mathcal{E}(S^{\dagger})\right)	+\sigma^2\right)\left[\left(\frac{1}{e}+2\right)\sum_{t=1}^{k-1}\frac{\eta_{t}^2}{1+\sum_{j=t+1}^{k}\eta_{j}}+\eta_k^2\right].
		\end{gather*}
        
        For $k\ge2$, applying Proposition~\ref{eta2} with $v=1$ and $T=k$,
and using $1/e+2\le3$, yields
\begin{equation*}
\begin{aligned}
\mathcal{B}_2
\le
\sqrt c\left(
\sqrt c\left(\widetilde M-\mathcal E(S^\dagger)\right)
+\sigma^2
\right)
\left[
3c_6(1)\eta_1\log\bigl(1+\eta_1(k+1)\bigr)
+\eta_1^2
\right].
\end{aligned}
\end{equation*}
The same bound holds for $k=1$, since the sum over
$1\le t\le k-1$ is empty.
The step-size condition implies $0<\eta_1<1$.
Since $k\le T$ and $T\ge2$,
\begin{equation*}
\log\bigl(1+\eta_1(k+1)\bigr)
\le\log(T+2)\le2\log(T+1),
\qquad
\eta_1^2\le\eta_1\log(T+1).
\end{equation*}
Recalling that $c_6(1)=1$, we obtain
\begin{equation*}
\begin{aligned}
3c_6(1)\eta_1\log\bigl(1+\eta_1(k+1)\bigr)+\eta_1^2
&\le7\eta_1\log(T+1)\\
&\le\frac{7}{1+14c}
\le\frac{1}{2c}.
\end{aligned}
\end{equation*}

		Thus, we have
		\begin{equation}
			\B_2\leq\frac{1}{2}\widetilde{M}+\frac{\left|\sigma^2-\sqrt{c}\mathcal{E}(S^\dagger)\right|}{2\sqrt{c}}. \label{temp24}
		\end{equation}
		Consequently, plugging (\ref{temp11}) and (\ref{temp24}) into (\ref{temp10}) yields the desired result,
		\begin{align*}
			\be_{z^{k}}\left[\mathcal{E}(S_{k+1})\right]\leq\|S^\dagger\|^2+\mathcal{E}(S^\dagger)+\frac{\widetilde{M}}{2}+\frac{\left|\sigma^2-\sqrt{c}\mathcal{E}(S^\dagger)\right|}{2\sqrt{c}}\leq \widetilde{M}.
		\end{align*}
		The proof is complete.
	\end{proof}

 \subsection{Proof of Theorem \ref{Thm:nonlinear}} \label{Appendix:nonlinear}

We first establish the excess-risk identity and the martingale
decomposition for model \eqref{nonlinear}, and then verify the modified
step-size conditions. By the normal equation \eqref{non1} and the
definition of $\delta(x)$, we have
\begin{equation} \label{non2}
    \mathbb{E}[\delta(x) \otimes x] = 0,
\end{equation}
which enables us to extend the proof of the convergence analysis of the SGD algorithm from linear to nonlinear model. For any $S \in \mathcal{B}(\mathcal{H}_1, \mathcal{H}_2)$, there holds
\begin{equation} \label{nonlinear1}
    \begin{aligned}
        \mathcal{E}(S) - \mathcal{E}(S^\dagger)
        &= \mathbb{E}\left[\|Sx - y\|_{\mathcal{H}_2}^2\right] - \mathbb{E}\left[\|S^\dagger x - y\|_{\mathcal{H}_2}^2\right] \\
        &= \mathbb{E}\left[\|(S - S^\dagger)x\|_{\mathcal{H}_2}^2\right] + 2\mathbb{E}\left[\langle y - S^\dagger x, (S^\dagger - S)x\rangle_{\mathcal{H}_2}\right].
    \end{aligned}
\end{equation}
Calculating $\mathbb{E}\left[\langle y - S^\dagger x, (S^\dagger - S)x\rangle_{\mathcal{H}_2}\right]$ separately yields
\begin{equation} \label{nonlinear2}
    \begin{aligned}
        \mathbb{E}\left[\langle y - S^\dagger x, (S^\dagger - S)x\rangle_{\mathcal{H}_2}\right]
        &= \mathbb{E}\left[\langle \delta(x), (S^\dagger - S)x\rangle_{\mathcal{H}_2}\right] \\
        &= \sum_{i \geq 1} \mathbb{E}\left[\langle \delta(x), f_i\rangle_{\mathcal{H}_2}\langle(S^\dagger - S)x, f_i\rangle_{\mathcal{H}_2}\right] \\
        &= \sum_{i \geq 1} \mathbb{E}\left[\langle(\delta(x) \otimes x)(S^\dagger - S)^*f_i, f_i\rangle_{\mathcal{H}_2}\right] \\
        &= 0,
    \end{aligned}
\end{equation}
where we utilize the equality \eqref{non2}. Subsequently, substituting \eqref{nonlinear2} into \eqref{nonlinear1} results in
\begin{equation} \label{nonlinear3}
    \mathcal{E}(S) - \mathcal{E}(S^\dagger)
    = \left\|(S - S^\dagger)L_C^{1/2}\right\|^2_{\mathrm{HS}}.
\end{equation}
Hence, the expression for \(\mathcal{E}(S) - \mathcal{E}(S^\dagger)\) in the nonlinear model, given by \eqref{nonlinear3}, matches the form established in the linear model, as shown in \eqref{form2}.


Lemma \ref{lemma4.1} also holds for the nonlinear model. To prove this, we only need to verify that $\mathbb{E}_{z_t}[\mathcal{B}_t] = 0, \forall t \geq 1$, where $\mathcal{B}_t = (S_t - S^\dagger)L_C + (y_t - S_t x_t) \otimes x_t$. Given $y_t = S^\dagger x_t + \delta(x_t) + \epsilon_t$, direct calculation shows
\begin{equation*}
    \begin{aligned}
        \mathbb{E}_{z_t}[\mathcal{B}_t]
        &= \mathbb{E}_{z_t}\left[(S_t - S^\dagger)L_C + \left[(S^\dagger - S_t)x_t + \delta(x_t)+ \epsilon_t\right] \otimes x_t\right] \\
        &= \mathbb{E}_{z_t}[\delta(x_t) \otimes x_t] = 0,
    \end{aligned}
\end{equation*}
where the fact \eqref{non2} has been utilized. Hence, the martingale decomposition \eqref{form} and the
orthogonality arguments in the proof of Proposition~\ref{Proposition error 1} remain valid.

With $\nu_\delta^2$ as defined in Theorem~\ref{Thm:nonlinear}, the argument of
Proposition~\ref{Proposition error 1} yields the following bound for every $\alpha\ge0$
such that
$S^\dagger L_C^\alpha\in\B_{\mathrm{HS}}(\H_1,\H_2)$:
\begin{equation*}
\mathbb{E}_{z^T}\left[\left\|(S_{T+1} - S^{\dagger})L_C^{\alpha}\right\|_{\mathrm{HS}}^2\right] \leq \mathcal{T}_1 + \mathcal{T}_2,
\end{equation*}
where
\begin{equation*}
\begin{split}
&\mathcal{T}_1 := \left\|S^{\dagger}L_C^{\alpha}\prod_{t=1}^{T}(I - \eta_{t}L_C)\right\|_{\mathrm{HS}}^2,\\
&\mathcal{T}_2 := 3\sum_{t=1}^T \eta_t^2 \left(c\mathbb{E}_{z^{t-1}}\left[\mathcal{E}(S_{t}) - \mathcal{E}(S^{\dagger})\right] + \sqrt{c}(\nu_\delta^2 + \sigma^2)\right)\mathrm{Tr}\left(L_C^{1+2\alpha}\prod_{j=t+1}^{T}(I - \eta_{j}L_C)^2\right).
\end{split}
\end{equation*}
In the proof of Proposition~\ref{Proposition error 1}, the upper bound in \eqref{temp2} is
replaced by the following bound, using model \eqref{nonlinear} and
$\|u+v+w\|^2\le3(\|u\|^2+\|v\|^2+\|w\|^2)$:
\begin{equation*}
3\sum_{t=1}^{T}\eta_{t}^{2}\mathbb{E}_{z^{t-1}}\mathbb{E}_{x_t}\left[\left(\left\|(S_{t}-S^\dagger)x_{t}\right\|^2_{\mathcal H_2} + \left\|\delta(x_t)\right\|^2_{\mathcal H_2} + \sigma^2\right)\left\|L_C^{\alpha}\prod_{j=t+1}^{T}(I - \eta_{j}L_C)x_{t}\right\|_{\mathcal{H}_{1}}^2\right].
\end{equation*}
Here the additional term involving \(\delta(x_t)\) is controlled by Cauchy--Schwarz inequality and Assumption \ref{a3} as
\[
\mathbb{E}_{x_t}\left[\|\delta(x_t)\|_{\mathcal H_2}^2\|Ax_t\|_{\mathcal H_1}^2\right]
\leq \nu_\delta^2\left(\mathbb{E}_{x_t}\|Ax_t\|_{\mathcal H_1}^4\right)^{1/2}
\leq \sqrt{c}\nu_\delta^2\mathbb{E}_{x_t}\|Ax_t\|_{\mathcal H_1}^2,
\]
where \(A=L_C^{\alpha}\prod_{j=t+1}^{T}(I - \eta_{j}L_C)\).


It remains to verify that the modified step-size conditions give
uniform bounds on the prediction risk. Put
\begin{equation*}
D_\delta:=\nu_\delta^2+\sigma^2,
\qquad
q_t:=\mathbb{E}_{z^{t-1}}
[\mathcal{E}(S_t)-\mathcal{E}(S^\dagger)],
\end{equation*}
and, for $k\ge1$, define
\begin{equation*}
H_k:=
\sum_{t=1}^k\eta_t^2
\operatorname{Tr}\!\left(
L_C^2\prod_{j=t+1}^k(I-\eta_jL_C)^2
\right).
\end{equation*}
The preceding error decomposition with $\alpha=1/2$ implies
\begin{equation*}
q_{k+1}
\le
\|S^\dagger\|^2+
3\left(c\max_{1\le t\le k}q_t+\sqrt c\,D_\delta\right)H_k.
\end{equation*}
Also, $q_1=\|S^\dagger L_C^{1/2}\|_{\mathrm{HS}}^2
\le\|S^\dagger\|^2$.
Using $\operatorname{Tr}(L_C)\le1$, $\|L_C\|\le1$,
and Lemma~\ref{lemma basic} for $t<k$, while treating $t=k$ separately,
gives
\begin{equation*}
H_k
\le
3\sum_{t=1}^{k-1}
\frac{\eta_t^2}{1+\sum_{j=t+1}^k\eta_j}
+\eta_k^2.
\end{equation*}

For decaying step sizes, Proposition~\ref{eta1} and \eqref{temp6} yield
$H_k\le\xi$ for every $k\ge1$, where
\begin{equation*}
\xi:=
3c_5(\eta_1,1,\theta)
\max\left\{\frac{1}{e\theta},1\right\}
+\eta_1^2.
\end{equation*}
Under $3c\xi<1$, induction in the preceding recurrence gives
\begin{equation*}
q_t\le Q_\delta:=
\frac{\|S^\dagger\|^2+3\sqrt c\,D_\delta\xi}
{1-3c\xi},
\qquad t\ge1.
\end{equation*}

For constant step sizes $\eta_t=\eta_1$ within a horizon
$T\ge2$, assume
$\eta_1\le[(1+42c)\log(T+1)]^{-1}$.
For $2\le k\le T$, Proposition~\ref{eta2} with $v=1$ gives
\begin{equation*}
\begin{aligned}
H_k
&\le
3\eta_1\log\bigl(1+\eta_1(k+1)\bigr)+\eta_1^2\\
&\le7\eta_1\log(T+1)
\le\frac{7}{1+42c}
\le\frac{1}{6c}.
\end{aligned}
\end{equation*}
The same bound holds for $k=1$, since the preceding sum is empty.
In the second inequality we used $\eta_1\le1$,
$\log(1+\eta_1(k+1))\le\log(T+2)\le2\log(T+1)$,
and $\eta_1^2\le\eta_1\log(T+1)$.
Induction therefore yields
\begin{equation*}
q_t\le\widetilde Q_\delta:=
2\|S^\dagger\|^2+\frac{D_\delta}{\sqrt c},
\qquad 1\le t\le T+1.
\end{equation*}
For the choices in part (2) of Theorem~\ref{Thm:nonlinear}, the condition
$\eta_*\le ea/(1+42c)$ ensures the required bound on $\eta_1$,
because
$\sup_{u>0}u^{-a}\log u=1/(ea)$.

Thus, the nonlinear analogues of the risk bounds in
Propositions~\ref{p1} and~\ref{p2} hold with
\begin{equation*}
M_\delta=\mathcal{E}(S^\dagger)+Q_\delta,
\qquad
\widetilde M_\delta=
\mathcal{E}(S^\dagger)+\widetilde Q_\delta,
\end{equation*}
respectively. Substituting these uniform bounds into the
cumulative sample term changes only its multiplicative constants.
The approximation estimates and spectral-sum estimates in
Sections~\ref{proof1} and~\ref{proof2} remain unchanged, including the separate
treatment of $\alpha=0$ and $s=1$.
Repeating the corresponding choices of step-size exponents
therefore gives the same powers of $T+1$ and the same logarithmic
factors as in Theorems~\ref{t1}--\ref{t4}. This completes the proof.

\subsection{Proof of Lemma \ref{minimax4}} \label{Proof of Lemma minimax4}
\begin{proof}
Let $\widetilde\iota=(\widetilde\iota_{ij})_{1\le i,j\le m}$ be a random
matrix with independent Rademacher entries, i.e.,
$\mathbb P(\widetilde\iota_{ij}=1)=\mathbb P(\widetilde\iota_{ij}=-1)=1/2$. We first show that, for a sufficiently
large universal constant $C_*>0$, there holds that
\[
\mathbb P\left(
\|\widetilde\iota\|>C_*\sqrt m
\right)\le \frac12 .
\]

Let $\mathcal N$ be a $1/4$-net of the unit sphere in $\mathbb R^m$ satisfying
$|\mathcal N|\le 9^m$ (see \cite[Example 5.8]{wainwright2019high}). That is, for every unit vector $u\in\mathbb R^m$, there exists
$u_0\in\mathcal N$ such that $\|u-u_0\|_2\le 1/4$. 

Set $M=\max_{u,v\in\mathcal N}|u^\top \widetilde\iota v|.$
For any unit vectors $u,v\in\mathbb R^m$, choose $u_0,v_0\in\mathcal N$ such that
$\|u-u_0\|_2\le 1/4$ and $\|v-v_0\|_2\le 1/4$. Then
\[
|u^\top \widetilde\iota v|
\le
|u_0^\top \widetilde\iota v_0|
+
|(u-u_0)^\top \widetilde\iota v|
+
|u_0^\top \widetilde\iota (v-v_0)|
\le
M+\frac12\|\widetilde\iota\|.
\]
Taking the supremum over all unit vectors $u,v$ gives that $\|\widetilde\iota\|\le 2M.$

For fixed $u,v\in\mathcal N$,
\[
u^\top \widetilde\iota v
=
\sum_{i,j=1}^m \widetilde\iota_{ij}u_i v_j
\]
is a Rademacher sum with $\sum_{i,j=1}^m u_i^2v_j^2=1.$ Thus, by Hoeffding's inequality \cite[Proposition 2.5]{wainwright2019high},
\[
\mathbb P\left(|u^\top \widetilde\iota v|\ge t\right)
\le 2\exp\left(-\frac{t^2}{2}\right).
\]Therefore, 
\begin{equation} \label{temp50}
    \begin{aligned}
\mathbb P\left(
\|\widetilde\iota\|>C_*\sqrt m
\right)&\leq\mathbb P\left(
M>\frac{C_*}{2}\sqrt m
\right)
\\&\le
2|\mathcal N|^2
\exp\left(-\frac{C_*^2m}{8}\right)\le
2\exp\left(2m\log 9-\frac{C_*^2m}{8}\right).
    \end{aligned}
\end{equation}
Choosing $C_*$ sufficiently large, the right-hand side is no larger than $1/2$.

Define
\[
\mathcal G_m
=
\left\{
\iota\in\mathcal{H}_{m^2}:
\|\iota\|\le C_*\sqrt m
\right\}.
\]
The estimate \eqref{temp50} implies $|\mathcal G_m|\ge 2^{m^2-1}.$

Now let $\Lambda$ be a maximal subset of $\mathcal G_m$ such that any two distinct
elements $\iota,\iota'\in\Lambda$ satisfy
\begin{equation*} 
    \|\iota-\iota'\|_1>\frac{m^2}{2}.
\end{equation*}
By maximality, the $\|\cdot\|_1$-balls of radius $m^2/2$ centered at elements of
$\Lambda$ cover $\mathcal G_m$. Since
\[
\|\iota-\iota'\|_1
=
2\sum_{i,j=1}^m {\bf 1}_{\{\iota_{ij}\ne \iota'_{ij}\}},
\]
each such ball has cardinality at most
\[
\sum_{k=0}^{\lfloor m^2/4\rfloor}{m^2\choose k}.
\]
For any $t>0$,
\[
\sum_{k=0}^{\lfloor m^2/4\rfloor}{m^2\choose k}
\le
e^{tm^2/4}\sum_{k=0}^{m^2} {m^2\choose k}e^{-tk}
=
e^{tm^2/4}(1+e^{-t})^{m^2}.
\]
Taking $t=\log 3$ gives
\[
\sum_{k=0}^{\lfloor m^2/4\rfloor}{m^2\choose k}
\le
\exp\left(m^2\left(\log4-\frac{3}{4}\log 3\right)\right).
\]
Then, we obtain
\begin{equation*}
    |\Lambda|\geq\frac{2^{m^2-1}}{\exp\left(m^2\left(\log4-\frac{3}{4}\log 3\right)\right)}=
\exp\left(\left(\frac{3}{4}\log 3-\log 2\right)m^2-\log 2\right).
\end{equation*}
Since $\frac{3}{4}\log 3-\log 2>1/8$, it follows that, for all sufficiently large $m$,
\[
|\Lambda|\ge e^{m^2/8}.
\]

The proof is complete.
\end{proof}

\subsection{Proof of Proposition \ref{minimax3}} \label{Prop 7.4}
\begin{proof}
		Let $\iota^{(1)},\iota^{(2)}\cdots\iota^{(L)}\in\left\{-1,1\right\}^{m^2}$ be given by Lemma \ref{minimax4} with $L\geq e^{m^2/8}$. Define $J_{(i)}\in\mathcal{B}(\H_1,\H_2)$ for $i=1,\cdots,L$ as
		\begin{equation} \label{J_i}
			J_{(i)}\phi_k=\begin{cases}
				0, & \text{ if }k\leq m, \\
				\frac{R}{C_*\sqrt{m}}\sum_{l=1}^{m}\iota_{k-m,l}^{(i)}f_{l}, &\text{ if }m+1\leq k\leq2m, \\
				0, &\text{ if }k>2m.
			\end{cases}
		\end{equation}
		Therefore, $\|J_{(i)}\|\leq R$ and $\mathrm{Ran}(J_{(i)})\subseteq \mathrm{span}\{f_j:1\leq j\leq m\}$. 
        Together with the input construction in Subsection \ref{Minimax Lower Bound  under Weak Regularity Condition},
this gives $\rho_i\in\mathcal P_\omega$. Then, recall that $\{\phi_k\}_{k\geq1}$ is the sequence of orthonormal eigenvector of $L_C$ with $L_C\phi_k=\lambda_k\phi_k$,
  we deduce that
		\begin{equation*}
			\begin{aligned}
				S_{(i)}x&=\sum_{k\geq1}\langle x,\phi_k\rangle J_{(i)}L_C^r\phi_k
				=\sum_{k\geq1}\langle x,\phi_k\rangle \lambda_k^rJ_{(i)}\phi_k\\
				&=\sum_{k= m+1}^{2m}\langle x,\phi_k\rangle \lambda_k^r\frac{R}{C_*\sqrt{m}}\sum_{l=1}^{m}\iota_{k-m,l}^{(i)}f_{l}\\
				&=\sum_{l=1}^{m}\left(\frac{R}{C_*\sqrt{m}}\sum_{k=1}^{m}\langle x,\phi_{k+m}\rangle\iota_{k,l}^{(i)}\lambda_{k+m}^r\right)f_{l}.
			\end{aligned}
		\end{equation*}
		Thus, $y=S_{(i)}x+\epsilon=\sum_{l=1}^{m}\left(\frac{R}{C_*\sqrt{m}}\sum_{k=1}^{m}\langle x,\phi_{k+m}\rangle\iota_{k,l}^{(i)}\lambda_{k+m}^r+\epsilon_l\right)f_{l}.$		
		Given $\epsilon$ in \eqref{epsilon}, the conditional distribution $\rho_i(y\mid x)$ can be treated as a gaussian distribution on $\mathbb{R}^m$, i.e. $y\mid x\sim\N(\theta^{(i)},\frac{\sigma^2}{m}I_m)$, where $\theta_l^{(i)}=\frac{R}{C_*\sqrt{m}}\sum_{k=1}^{m}\langle x,\phi_{k+m}\rangle\iota_{k,l}^{(i)}\lambda_{k+m}^r$ for any $1\leq l\leq m$. So,
		\begin{equation*}
			\begin{aligned}
				D_{kl}(\rho_i\parallel \rho_j)=&\mathbb{E}_x\left[D_{kl}\left(\rho_i(\cdot\mid x)\parallel \rho_j(\cdot\mid x)\right)\right]\\
				=&\mathbb{E}_x\left[\frac{m}{2\sigma^2}\sum_{l=1}^{m}\left(\frac{R}{C_*\sqrt{m}}\sum_{k=1}^{m}\langle x,\phi_{k+m}\rangle\left(\iota_{k,l}^{(i)}-\iota_{k,l}^{(j)}\right)\lambda_{k+m}^r\right)^2\right]\\
				=&\frac{R^2}{2C_*^2\sigma^2}\sum_{l=1}^{m}\sum_{k_1,k_2=1}^{m}\left(\iota_{k_1,l}^{(i)}-\iota_{k_1,l}^{(j)}\right)\left(\iota_{k_2,l}^{(i)}-\iota_{k_2,l}^{(j)}\right)\lambda_{k_1+m}^r\lambda_{k_2+m}^r\\
				&\times\mathbb{E}_x\left[\langle x,\phi_{k_1+m}\rangle\langle x,\phi_{k_2+m}\rangle\right].
			\end{aligned}
		\end{equation*}
		Since $\mathbb{E}_x\left[\langle x,\phi_{k_1+m}\rangle\langle x,\phi_{k_2+m}\rangle\right]=\langle L_C\phi_{k_1+m},\phi_{k_2+m}\rangle$, we see that
		\begin{equation*}
			\begin{aligned}		
				D_{kl}(\rho_i\parallel \rho_j)\leq&\frac{R^2}{2C_*^2\sigma^2}\sum_{l=1}^{m}\sum_{k=1}^{m}\left(\iota_{k,l}^{(i)}-\iota_{k,l}^{(j)}\right)^2\lambda_{k+m}^{2r+1}\\
				\leq&\frac{2R^2}{C_*^2\sigma^2}m^2\lambda_{m}^{2r+1}
				\leq\frac{2R^2}{C_*^2\sigma^2}q^{2r+1}m^{-\frac{2r+1}{s}+2}.
			\end{aligned}
		\end{equation*}
		Therefore,
		\begin{equation*}
			D_{kl}\left(\rho_i^{\otimes T}\parallel \rho_j^{\otimes T}\right)\leq\frac{2R^2}{C_*^2\sigma^2}q^{2r+1}m^{-\frac{2r+1}{s}+2}T.
		\end{equation*}
		Next,
		\begin{equation*}
			\begin{aligned}
				d(S_{(i)},S_{(j)})^2&=\left\|(J_{(i)}-J_{(j)})L_C^{r+\frac12}\right\|_\mathrm{HS}^2\\
				&=\sum_{k\geq1}\left\|(J_{(i)}-J_{(j)})L_C^{r+\frac12}\phi_k\right\|^2\\
				&=\sum_{k\geq1}\lambda_k^{2r+1}\left\|(J_{(i)}-J_{(j)})\phi_k\right\|^2\\
				&=\sum_{k=m+1}^{2m}\lambda_k^{2r+1}\frac{R^2}{C_*^2m}\sum_{l=1}^{m}
				\left(\iota_{k-m,l}^{(i)}-\iota_{k-m,l}^{(j)}\right)^2,
			\end{aligned}
		\end{equation*}	
            where the last equality follows from the definition of $J_{(i)}$ in \eqref{J_i}.
		By Lemma \ref{minimax4},
		\begin{equation*}
			\begin{aligned}
				d(S_{(i)},S_{(j)})^2\geq&\lambda_{2m}^{2r+1}\frac{R^2}{C_*^2m}\sum_{k=m+1}^{2m}\sum_{l=1}^{m}
				\left(\iota_{k-m,l}^{(i)}-\iota_{k-m,l}^{(j)}\right)^2 \\
				=&4\lambda_{2m}^{2r+1}\frac{R^2}{C_*^2m}\sum_{k=m+1}^{2m}\sum_{l=1}^{m}\mathbf{1}_{\{\iota_{k-m,l}^{(i)}\not=\iota_{k-m,l}^{(j)}\}} \\
				\geq&\frac{\lambda_{2m}^{2r+1}R^2}{C_*^2}m\geq\frac{2^{-\frac{2r+1}{s}}q^{2r+1}R^2}{C_*^2}m^{-\frac{2r+1}{s}+1}.
			\end{aligned}
		\end{equation*}		
		Thus, there holds that
		\begin{equation*}
			d(S_{(i)},S_{(j)})\geq\frac{2^{-\frac{2r+1}{2s}}q^{r+1/2}R}{C_*}m^{-\frac{2r+1}{2s}+1/2},
		\end{equation*}
		which completes the proof.	
	\end{proof}
\subsection{Proof of Proposition \ref{minimax5}}
\label{Prop 7.5}
\begin{proof}
		Let $\iota^{(1)},\iota^{(2)}\cdots\iota^{(L)}$ be given by Lemma \ref{minimax1} with $L\geq e^{m/8}$. We define $\beta_i\in\H_1$ for $i=1,\cdots,L$ as follows,
		\begin{equation*}
			\beta_i=\sum_{k=m+1}^{2m}\frac{R}{\sqrt{m}}\lambda_k^{\widetilde{r}}\iota_{k-m}^{(i)}\phi_k,
		\end{equation*}
		where $\phi_k$ is the corresponding eigenvector of $\lambda_k$ and $g_i=\sum_{k=m+1}^{2m}\frac{R}{\sqrt{m}}\iota_{k-m}^{(i)}\phi_k$ with $\|g_i\|=R$ and $\beta_i=L_C^{\widetilde{r}}g_i$. 
Let $\widetilde J_{(i)}=f_1\otimes g_i$, so that
$\|\widetilde J_{(i)}\|_{\mathrm{HS}}=R$.
Let $\rho_i$ be the distribution of $(x,y)$ with
$y=\langle\beta_i,x\rangle_{\H_1}f_1+\epsilon$,
where $x$ and $\epsilon$ are chosen as in Subsection \ref{Minimax Lower Bound  under Strong Regularity Condition}.
Then $\rho_i\in\mathcal P'_\omega$, and        
$\langle y,f_1\rangle_{\H_2}\mid x\sim\N(\langle\beta_i,x\rangle_{\H_1},\sigma^2)$ and $\langle y,f_j\rangle_{\H_2}=0$ if $j\geq2$. Therefore, for any $1\leq i\not=j\leq L$,
		\begin{equation*}
			\begin{aligned}
				D_{kl}(\rho_i^{\otimes T}\parallel \rho_j^{\otimes T})&=T\mathbb{E}_x\left[D_{kl}\left(\rho_i(y\mid x)\parallel \rho_j(y\mid x)\right)\right]\\
				&=\frac{T}{2\sigma^2}\mathbb{E}_x\left[\langle\beta_i-\beta_j,x\rangle_{\H_1}^2\right] \\
				&=\frac{T}{2\sigma^2}\left\|L_C^{\frac12}(\beta_i-\beta_j)\right\|_{\H_1}^2\\
				&=\frac{T}{2\sigma^2}\sum_{k=m+1}^{2m}\frac{R^2}{m}\lambda_k^{2\widetilde{r}+1}\left(\iota_{k-m}^{(i)}-\iota_{k-m}^{(j)}\right)^2\\
				&\leq\frac{2TR^2}{\sigma^2}\lambda_m^{2\widetilde{r}+1}
				\leq\frac{2TR^2}{\sigma^2}q^{2\widetilde{r}+1}m^{-\frac{2\widetilde{r}+1}{s}}.
			\end{aligned}
		\end{equation*}
		Next,
		\begin{equation*}
			\begin{aligned}
				d(\beta_i,\beta_j)^2&=\left\|L_C^{\alpha}(\beta_i-\beta_j)\right\|_{\H_1}^2
				=\frac{R^2}{m}\sum_{k=m+1}^{2m}\lambda_k^{2\widetilde{r}+2\alpha}\left(\iota_{k-m}^{(i)}-\iota_{k-m}^{(j)}\right)^2\\
				&\geq\frac{4R^2}{m}\lambda_{2m}^{2\widetilde{r}+2\alpha}\sum_{k=m+1}^{2m}\mathbf{1}_{\{\iota_{k-m}^{(i)}\not=\iota_{k-m}^{(j)}\}}\\
				&\geq 2^{-\frac{2\widetilde{r}+2\alpha}{s}}q^{2\widetilde{r}+2\alpha}R^2m^{-\frac{2\widetilde{r}+2\alpha}{s}},
			\end{aligned}
		\end{equation*}
		where in the last inequality Lemma \ref{minimax1} has been used.
            Thus
		\[
		d(\beta_i,\beta_j)\geq2^{-\frac{\widetilde{r}+\alpha}{s}}q^{\widetilde{r}+\alpha}Rm^{-\frac{\widetilde{r}+\alpha}{s}},
		\]
		which completes the proof.
	\end{proof}

\bibliographystyle{plain}
\bibliography{main}

\begin{thebibliography}{10}

\bibitem{li2020neural}
Anima Anandkumar, Kamyar Azizzadenesheli, Kaushik Bhattacharya, Nikola Kovachki, Zongyi Li, Burigede Liu, and Andrew Stuart.
\newblock Neural operator: Graph kernel network for partial differential equations.
\newblock In {\em ICLR 2020 Workshop on Integration of Deep Neural Models and Differential Equations}, 2020.

\bibitem{batlle2024kernel}
Pau Batlle, Matthieu Darcy, Bamdad Hosseini, and Houman Owhadi.
\newblock Kernel methods are competitive for operator learning.
\newblock {\em Journal of Computational Physics}, 496:112549, 2024.

\bibitem{bharucha1960random}
AT~Bharucha-Reid.
\newblock On random solutions of {Fredholm} integral equations.
\newblock {\em Bulletin of the American Mathematical Society}, 66(2):104--109, 1960.

\bibitem{bhattacharya2021model}
Kaushik Bhattacharya, Bamdad Hosseini, Nikola~B Kovachki, and Andrew~M Stuart.
\newblock Model reduction and neural networks for parametric {PDEs}.
\newblock {\em The SMAI Journal of Computational Mathematics}, 7:121--157, 2021.

\bibitem{blanchard2018optimal}
Gilles Blanchard and Nicole M{\"u}cke.
\newblock Optimal rates for regularization of statistical inverse learning problems.
\newblock {\em Foundations of Computational Mathematics}, 18:971--1013, 2018.

\bibitem{bosq2000linear}
Denis Bosq.
\newblock {\em Linear Processes in Function Spaces: Theory and Applications}, volume 149.
\newblock Springer Science \& Business Media, 2000.

\bibitem{boulle2023elliptic}
Nicolas Boull{\'e}, Diana Halikias, and Alex Townsend.
\newblock Elliptic {PDE} learning is provably data-efficient.
\newblock {\em Proceedings of the National Academy of Sciences}, 120(39):e2303904120, 2023.

\bibitem{bouziani2024structure}
Nacime Bouziani and Nicolas Boull{\'e}.
\newblock Structure-preserving operator learning.
\newblock {\em arXiv preprint arXiv:2410.01065}, 2024.

\bibitem{brogat2022vector}
Luc Brogat-Motte, Alessandro Rudi, C{\'e}line Brouard, Juho Rousu, and Florence d’Alch{\'e} Buc.
\newblock Vector-valued least-squares regression under output regularity assumptions.
\newblock {\em The Journal of Machine Learning Research}, 23(344):1--50, 2022.

\bibitem{brouard2016fast}
C{\'e}line Brouard, Huibin Shen, Kai D{\"u}hrkop, Florence d'Alch{\'e} Buc, Sebastian B{\"o}cker, and Juho Rousu.
\newblock Fast metabolite identification with input output kernel regression.
\newblock {\em Bioinformatics}, 32(12):i28--i36, 2016.

\bibitem{brouard2016input}
C{\'e}line Brouard, Marie Szafranski, and Florence d'Alch{\'e} Buc.
\newblock Input output kernel regression: Supervised and semi-supervised structured output prediction with operator-valued kernels.
\newblock {\em Journal of Machine Learning Research}, 17(176):1--48, 2016.

\bibitem{buldygin2000metric}
Valeri{\u\i}~Vladimirovich Buldygin and IU~V Kozachenko.
\newblock {\em Metric Characterization of Random Variables and Random Processes}, volume 188.
\newblock American Mathematical Society, 2000.

\bibitem{cai2021deepm}
Shengze Cai, Zhicheng Wang, Lu~Lu, Tamer~A Zaki, and George~Em Karniadakis.
\newblock Deepm\&mnet: Inferring the electroconvection multiphysics fields based on operator approximation by neural networks.
\newblock {\em Journal of Computational Physics}, 436:110296, 2021.

\bibitem{cai2012minimax}
T~Tony Cai and Ming Yuan.
\newblock Minimax and adaptive prediction for functional linear regression.
\newblock {\em Journal of the American Statistical Association}, 107(499):1201--1216, 2012.

\bibitem{caponnetto2008universal}
Andrea Caponnetto, Charles~A Micchelli, Massimiliano Pontil, and Yiming Ying.
\newblock Universal multi-task kernels.
\newblock {\em The Journal of Machine Learning Research}, 9:1615--1646, 2008.

\bibitem{carmeli2006vector}
Claudio Carmeli, Ernesto De~Vito, and Alessandro Toigo.
\newblock Vector valued reproducing kernel {Hilbert spaces of integrable functions and Mercer theorem}.
\newblock {\em Analysis and Applications}, 4(04):377--408, 2006.

\bibitem{carmeli2010vector}
Claudio Carmeli, Ernesto De~Vito, Alessandro Toigo, and Veronica Umanit{\'a}.
\newblock Vector valued reproducing kernel {Hilbert} spaces and universality.
\newblock {\em Analysis and Applications}, 8(01):19--61, 2010.

\bibitem{chen1995universal}
Tianping Chen and Hong Chen.
\newblock Universal approximation to nonlinear operators by neural networks with arbitrary activation functions and its application to dynamical systems.
\newblock {\em IEEE Transactions on Neural Networks}, 6(4):911--917, 1995.

\bibitem{chen2022online}
Xiaming Chen, Bohao Tang, Jun Fan, and Xin Guo.
\newblock Online gradient descent algorithms for functional data learning.
\newblock {\em Journal of Complexity}, 70:101635, 2022.

\bibitem{chen2021hanson}
Xiaohui Chen and Yun Yang.
\newblock {Hanson--Wright inequality in Hilbert spaces with application to K-means clustering for non-Euclidean data}.
\newblock {\em Bernoulli}, 27(1):586--614, 2021.

\bibitem{ciliberto2016consistent}
Carlo Ciliberto, Lorenzo Rosasco, and Alessandro Rudi.
\newblock A consistent regularization approach for structured prediction.
\newblock {\em Advances in neural information processing systems}, 29, 2016.

\bibitem{ciliberto2020general}
Carlo Ciliberto, Lorenzo Rosasco, and Alessandro Rudi.
\newblock A general framework for consistent structured prediction with implicit loss embeddings.
\newblock {\em The Journal of Machine Learning Research}, 21(1):3852--3918, 2020.

\bibitem{crambes2013asymptotics}
Christophe Crambes and Andr{\'e} Mas.
\newblock Asymptotics of prediction in functional linear regression with functional outputs.
\newblock {\em Bernoulli}, 19(5B):2627--2651, 2013.

\bibitem{da2014introduction}
Giuseppe Da~Prato.
\newblock {\em Introduction to Stochastic Analysis and Malliavin Calculus}, volume~13.
\newblock Springer, 2014.

\bibitem{di2021deeponet}
Patricio~Clark Di~Leoni, Lu~Lu, Charles Meneveau, George~Em Karniadakis, and Tamer~A Zaki.
\newblock Neural operator prediction of linear instability waves in high-speed boundary layers.
\newblock {\em Journal of Computational Physics}, 474:111793, 2023.

\bibitem{MR3519927}
Aymeric Dieuleveut and Francis Bach.
\newblock Nonparametric stochastic approximation with large step-sizes.
\newblock {\em The Annals of Statistics}, 44(4):1363--1399, 2016.

\bibitem{dieuleveut2017harder}
Aymeric Dieuleveut, Nicolas Flammarion, and Francis Bach.
\newblock Harder, better, faster, stronger convergence rates for least-squares regression.
\newblock {\em The Journal of Machine Learning Research}, 18(1):3520--3570, 2017.

\bibitem{duchi2016lecture}
John Duchi.
\newblock Lecture notes for statistics 311/electrical engineering 377, February 2016.
\newblock Stanford University.

\bibitem{dunford1988linear}
Nelson Dunford and Jacob~T Schwartz.
\newblock {\em Linear Operators, Part 1: General Theory}, volume~10.
\newblock John Wiley \& Sons, 1988.

\bibitem{guo2022capacity}
Xin Guo, Zheng-Chu Guo, and Lei Shi.
\newblock Capacity dependent analysis for functional online learning algorithms.
\newblock {\em Applied and Computational Harmonic Analysis}, 67:101567, 2023.

\bibitem{guo2022rates}
Xin Guo, Junhong Lin, and Ding-Xuan Zhou.
\newblock {Rates of convergence of randomized Kaczmarz algorithms in Hilbert spaces}.
\newblock {\em Applied and Computational Harmonic Analysis}, 61:288--318, 2022.

\bibitem{guo2023optimality}
Zheng-Chu Guo, Andreas Christmann, and Lei Shi.
\newblock Optimality of robust online learning.
\newblock {\em Foundations of Computational Mathematics}, 24(5):1455--1483, 2024.

\bibitem{guo2019fast}
Zheng-Chu Guo and Lei Shi.
\newblock Fast and strong convergence of online learning algorithms.
\newblock {\em Advances in Computational Mathematics}, 45(5):2745--2770, 2019.

\bibitem{gupta2021multiwavelet}
Gaurav Gupta, Xiongye Xiao, and Paul Bogdan.
\newblock Multiwavelet-based operator learning for differential equations.
\newblock {\em Advances in neural information processing systems}, 34:24048--24062, 2021.

\bibitem{jin2022mionet}
Pengzhan Jin, Shuai Meng, and Lu~Lu.
\newblock Mionet: Learning multiple-input operators via tensor product.
\newblock {\em SIAM Journal on Scientific Computing}, 44(6):A3490--A3514, 2022.

\bibitem{kadri2010nonlinear}
Hachem Kadri, Emmanuel Duflos, Philippe Preux, St{\'e}phane Canu, and Manuel Davy.
\newblock Nonlinear functional regression: A functional {RKHS} approach.
\newblock In {\em Proceedings of the Thirteenth International Conference on Artificial Intelligence and Statistics}, pages 374--380. JMLR Workshop and Conference Proceedings, 2010.

\bibitem{kadri2016operator}
Hachem Kadri, Emmanuel Duflos, Philippe Preux, St{\'e}phane Canu, Alain Rakotomamonjy, and Julien Audiffren.
\newblock Operator-valued kernels for learning from functional response data.
\newblock {\em The Journal of Machine Learning Research}, 17(20):1--54, 2016.

\bibitem{kadri2013generalized}
Hachem Kadri, Mohammad Ghavamzadeh, and Philippe Preux.
\newblock A generalized kernel approach to structured output learning.
\newblock In {\em International Conference on Machine Learning}, pages 471--479. PMLR, 2013.

\bibitem{kissas2022learning}
Georgios Kissas, Jacob~H Seidman, Leonardo~Ferreira Guilhoto, Victor~M Preciado, George~J Pappas, and Paris Perdikaris.
\newblock Learning operators with coupled attention.
\newblock {\em The Journal of Machine Learning Research}, 23(215):1--63, 2022.

\bibitem{korba2018structured}
Anna Korba, Alexandre Garcia, and Florence d'Alch{\'e} Buc.
\newblock A structured prediction approach for label ranking.
\newblock {\em Advances in Neural Information Processing Systems}, 31, 2018.

\bibitem{kovachki2021neural}
Nikola Kovachki, Zongyi Li, Burigede Liu, Kamyar Azizzadenesheli, Kaushik Bhattacharya, Andrew Stuart, and Anima Anandkumar.
\newblock Neural operator: Learning maps between function spaces with applications to {PDEs}.
\newblock {\em Journal of Machine Learning Research}, 24(89):1--97, 2023.

\bibitem{li2020fourier}
Zongyi Li, Nikola~Borislavov Kovachki, Kamyar Azizzadenesheli, Burigede Liu, Kaushik Bhattacharya, Andrew~M. Stuart, and Anima Anandkumar.
\newblock Fourier neural operator for parametric {Partial Differential Equations}.
\newblock In {\em International Conference on Learning Representations}, 2021.

\bibitem{lin2021operator}
Chensen Lin, Zhen Li, Lu~Lu, Shengze Cai, Martin Maxey, and George~Em Karniadakis.
\newblock Operator learning for predicting multiscale bubble growth dynamics.
\newblock {\em The Journal of Chemical Physics}, 154(10):104118, 2021.

\bibitem{MR3714260}
Junhong Lin and Lorenzo Rosasco.
\newblock Optimal rates for multi-pass stochastic gradient methods.
\newblock {\em The Journal of Machine Learning Research}, 18(1):3375--3421, 2017.

\bibitem{lu2021learning}
Lu~Lu, Pengzhan Jin, Guofei Pang, Zhongqiang Zhang, and George~Em Karniadakis.
\newblock Learning nonlinear operators via {DeepONet} based on the universal approximation theorem of operators.
\newblock {\em Nature Machine Intelligence}, 3(3):218--229, 2021.

\bibitem{lu2022comprehensive}
Lu~Lu, Xuhui Meng, Shengze Cai, Zhiping Mao, Somdatta Goswami, Zhongqiang Zhang, and George~Em Karniadakis.
\newblock A comprehensive and fair comparison of two neural operators (with practical extensions) based on fair data.
\newblock {\em Computer Methods in Applied Mechanics and Engineering}, 393:114778, 2022.

\bibitem{lutkepohl2013vector}
Helmut L{\"u}tkepohl.
\newblock Vector autoregressive models.
\newblock In {\em Handbook of Research Methods and Applications in Empirical Macroeconomics}, pages 139--164. Edward Elgar Publishing, 2013.

\bibitem{malfait2003historical}
Nicole Malfait and James~O Ramsay.
\newblock The historical functional linear model.
\newblock {\em Canadian Journal of Statistics}, 31(2):115--128, 2003.

\bibitem{micchelli2005learning}
Charles~A Micchelli and Massimiliano Pontil.
\newblock On learning vector-valued functions.
\newblock {\em Neural computation}, 17(1):177--204, 2005.

\bibitem{mollenhauer2022learning}
Mattes Mollenhauer, Nicole M{\"u}cke, and TJ~Sullivan.
\newblock Learning linear operators: Infinite-dimensional regression as a well-behaved non-compact inverse problem.
\newblock {\em arXiv preprint arXiv:2211.08875}, 2022.

\bibitem{nelsen2021random}
Nicholas~H Nelsen and Andrew~M Stuart.
\newblock The random feature model for input-output maps between banach spaces.
\newblock {\em SIAM Journal on Scientific Computing}, 43(5):A3212--A3243, 2021.

\bibitem{owhadi2020ideas}
Houman Owhadi.
\newblock Do ideas have shape? {Idea} registration as the continuous limit of artificial neural networks.
\newblock {\em Physica D: Nonlinear Phenomena}, 444:133592, 2023.

\bibitem{persson2024randomized}
David Persson, Nicolas Boull{\'e}, and Daniel Kressner.
\newblock Randomized {N}ystr{\"o}m approximation of non-negative self-adjoint operators.
\newblock {\em SIAM Journal on Mathematics of Data Science}, 7(2):670--698, 2025.

\bibitem{pillaud2018statistical}
Loucas Pillaud-Vivien, Alessandro Rudi, and Francis Bach.
\newblock Statistical optimality of stochastic gradient descent on hard learning problems through multiple passes.
\newblock {\em Advances in Neural Information Processing Systems}, 31, 2018.

\bibitem{reiss2020nonasymptotic}
Markus Reiss and Martin Wahl.
\newblock Nonasymptotic upper bounds for the reconstruction error of pca.
\newblock {\em The Annals of Statistics}, 48(2):1098--1123, 2020.

\bibitem{schafer2024sparse}
Florian Sch{\"a}fer and Houman Owhadi.
\newblock Sparse recovery of elliptic solvers from matrix-vector products.
\newblock {\em SIAM Journal on Scientific Computing}, 46(2):A998--A1025, 2024.

\bibitem{smale2006online}
Steve Smale and Yuan Yao.
\newblock Online learning algorithms.
\newblock {\em Foundations of computational mathematics}, 6:145--170, 2006.

\bibitem{szabo2016learning}
Zolt{\'a}n Szab{\'o}, Bharath~K Sriperumbudur, Barnab{\'a}s P{\'o}czos, and Arthur Gretton.
\newblock Learning theory for distribution regression.
\newblock {\em The Journal of Machine Learning Research}, 17(1):5272--5311, 2016.

\bibitem{tarres2014online}
Pierre Tarres and Yuan Yao.
\newblock Online learning as stochastic approximation of regularization paths: Optimality and almost-sure convergence.
\newblock {\em IEEE Transactions on Information Theory}, 60(9):5716--5735, 2014.

\bibitem{tsybakov2004introduction}
Alexandre~B. Tsybakov.
\newblock {\em Introduction to Nonparametric Estimation}.
\newblock Springer Series in Statistics. Springer, New York, NY, 2009.

\bibitem{wainwright2019high}
Martin~J Wainwright.
\newblock {\em High-Dimensional Statistics: A Non-Asymptotic Viewpoint}, volume~48.
\newblock Cambridge university press, 2019.

\bibitem{wang2023learning}
Dixi Wang and Cheng Yu.
\newblock Learning operators for identifying weak solutions to the navier-stokes equations.
\newblock {\em arXiv preprint arXiv:2306.10685}, 2023.

\bibitem{wang2022improved}
Sifan Wang, Hanwen Wang, and Paris Perdikaris.
\newblock Improved architectures and training algorithms for deep operator networks.
\newblock {\em Journal of Scientific Computing}, 92(2):35, 2022.

\bibitem{weston2002kernel}
Jason Weston, Olivier Chapelle, Vladimir Vapnik, Andr{\'e} Elisseeff, and Bernhard Sch{\"o}lkopf.
\newblock Kernel dependency estimation.
\newblock {\em Advances in Neural Information Processing Systems}, 15, 2002.

\bibitem{yang2025learning}
Jia-Qi Yang and Lei Shi.
\newblock Learning operators by regularized stochastic gradient descent with operator-valued kernels.
\newblock {\em arXiv preprint arXiv:2504.18184}, 2025.

\bibitem{ying2008online}
Yiming Ying and Massimiliano Pontil.
\newblock Online gradient descent learning algorithms.
\newblock {\em Foundations of Computational Mathematics}, 8(5):561--596, 2008.

\bibitem{yosida2012functional}
K{\"o}saku Yosida.
\newblock {\em Functional Analysis}.
\newblock Springer Science \& Business Media, 2012.

\bibitem{yuan2010reproducing}
Ming Yuan and T~Tony Cai.
\newblock A reproducing kernel {Hilbert} space approach to functional linear regression.
\newblock {\em The Annals of Statistics}, 38(6):3412--3444, 2010.

\end{thebibliography}

\end{document}